\documentclass{article}

\usepackage[T1]{fontenc}
\usepackage{iclr2027_conference,times}

\usepackage{microtype}
\usepackage{needspace}
\usepackage{float} %
\usepackage{placeins} %
\usepackage{graphicx}
\usepackage{subcaption}
\usepackage{booktabs}
\usepackage{multirow}
\usepackage{longtable}
\usepackage{amsmath}
\usepackage{amssymb}
\usepackage{mathtools}
\usepackage{xcolor}

\usepackage{tabularx}
\usepackage{array}
\usepackage{hyperref}
\usepackage{xurl}
\hypersetup{hidelinks,pdfauthor={Jiacheng Li, Yaping Li, Congsheng Xu, Qingchen Liu, Haonan Wang, Zhijie Deng, Jiahu Qin},pdftitle={ManipScope: A Diagnostic Benchmark for Instruction Following and Generalization in Robotic Manipulation}}

\newcolumntype{C}{>{\centering\arraybackslash}X}
\newcommand{\Benchmark}{\textsc{ManipScope}}
\newcommand{\NumBaseTasks}{50}

\title{ManipScope: A Diagnostic Benchmark for\\
Instruction Following and Generalization\\
in Robotic Manipulation}

\author{\parbox{0.96\textwidth}{\centering\normalfont
Jiacheng Li\textsuperscript{1,2}\quad Yaping Li\textsuperscript{2,5}\quad
Congsheng Xu\textsuperscript{2,3}\quad Qingchen Liu\textsuperscript{1}\\[3pt]
Haonan Wang\textsuperscript{2}\quad Zhijie Deng\textsuperscript{4,*}\quad
Jiahu Qin\textsuperscript{1,*}\\[7pt]
{\small\textsuperscript{1}University of Science and Technology of China\\
\textsuperscript{2}Sharpa Robotics\\
\textsuperscript{3}Shanghai Jiao Tong University\\
\textsuperscript{4}SJTU-Sharpa Joint Lab for Embodied AI, SJTU\\
\textsuperscript{5}The Chinese University of Hong Kong}\\[5pt]
{\footnotesize\textsuperscript{*}Co-corresponding authors.\\
\texttt{zhijied@sjtu.edu.cn}\quad\texttt{jhqin@ustc.edu.cn}}}}

\iclrfinalcopy

\begin{document}

\maketitle
\fancyhead{}
\renewcommand{\headrulewidth}{0pt}

\begin{abstract}
Completing a manipulation task does not establish that a robot follows the requested goal rather than a familiar routine. \Benchmark{} is an evaluation-only {diagnostic} benchmark for robotic instruction following under changed task requirements. {Built on RoboTwin~2.0's simulation resources, it specifies 48 held-out scenes across six task families with explicit source-task links; policies are trained on unmodified RoboTwin~2.0 demonstrations, and no demonstrations of the scenes are provided, so success requires reusing skills under a changed request rather than fitting supplied examples. Beyond task success, \Benchmark{} provides a chain of diagnostics in which each step tests one explanation of a failure. For twelve policy checkpoints, 84.6\% of failures in 13,200 episodes on eleven selection scenes displace a distractor, and gripper records show that failed episodes usually grasp a non-target. When the target is reassigned among the candidates on identical initial states, ten checkpoints follow every instruction on 3.9\% of 1,000 states, below the 16\% of an instruction-blind chooser, while a GPT-6 agent run as positive control reaches 88\%. The failure recurs on unmodified training tasks and is not explained by missing manipulation skills, so success in these scenes largely reflects instruction-independent preferences; two checkpoints trained on the same data nevertheless follow color instructions. In a feasibility test, letting an agent select the target and the policy execute partly restores instruction following. Case studies with future-video and internal-intervention probes further show that forecasts and execution can agree on the wrong target and that action changes need not transfer the goal. Success,} consistency, and sensitivity must each be checked against the requested goal. Resources and separate policy and agent leaderboards: {\href{https://embodied-ai-research.github.io/ManipScope_webpage/}{\textbf{Web}}} and {\href{https://github.com/USTC-AIR-Lab/ManiScope}{\textbf{Code}}}.

\end{abstract}

\begin{figure}[t]
  \captionsetup{font=small,skip=4pt}
  \centering
  \includegraphics[width=\textwidth]{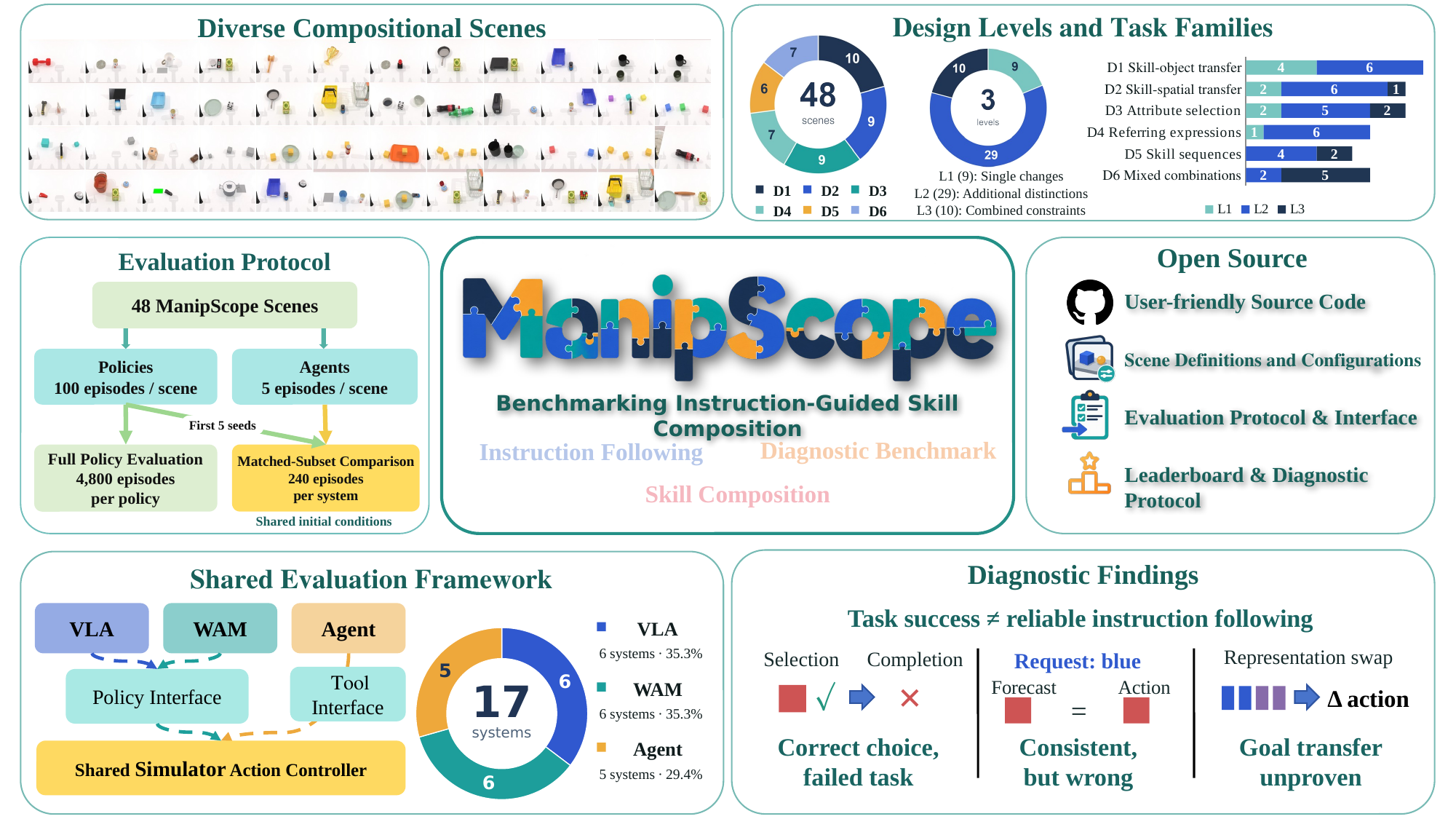}
  \caption{{Overview of \Benchmark{}. Forty-eight scenes span six task families and three design levels. Policies and tool-using agents are evaluated through their respective interfaces. Diagnostics distinguish selection from completion, prediction--action agreement from instruction correctness, and action sensitivity from goal-directed transfer. Open resources support evaluation and separate policy and agent leaderboards.}}
  \label{fig:overview}
\end{figure}

\begingroup
\setlength{\parskip}{4.5pt plus 0.5pt minus 0.5pt}
\setlength{\textfloatsep}{12pt plus 2pt minus 2pt}
\setlength{\floatsep}{10pt plus 2pt minus 2pt}
\setlength{\intextsep}{10pt plus 2pt minus 2pt}
\captionsetup{skip=6pt}
\makeatletter
\def\section{\@startsection{section}{1}{\z@}{-1.8ex plus -0.4ex minus -0.2ex}{1.0ex plus 0.2ex}{\large\sc\raggedright}}
\def\subsection{\@startsection{subsection}{2}{\z@}{-1.5ex plus -0.4ex minus -0.2ex}{0.6ex plus 0.2ex}{\normalsize\sc\raggedright}}
\makeatother
\Needspace{4\baselineskip}
\section{Introduction}
\label{sec:introduction}

Language-conditioned robot policies aim to turn verbal goals into physical behavior. RT-2 and OpenVLA illustrate how vision--language pretraining supports robotic control~\citep{zitkovich2023rt2,kim2025openvla}. When the requested object, relation, or combination changes, does the policy reuse its skills according to that request, or continue a familiar scene-dependent routine?

High success on familiar tasks cannot resolve this question alone. LIBERO-CF reports that evaluated VLAs can follow visual shortcuts and repeat learned behaviors despite alternative instructions in plausible layouts~\citep{fang2026liberocf}. InstructMove identifies a related ambiguity: visually salient or uniquely feasible targets can permit success without grounding the instruction~\citep{zhao2026instructmove}. When training on benchmark tasks is permitted, aggregate scores may not distinguish task-specific fitting from semantic generalization. Evaluation must therefore change the requirements that a familiar scene or routine would otherwise predict.

Such changes should remain interpretable. If a new task introduces an unavailable manipulation skill, failure may reflect that missing skill; if it reproduces demonstrated behavior, success need not establish generalization. This motivates a deliberate design choice{: policies are trained on the unmodified RoboTwin~2.0 demonstrations, and the evaluation scenes are held out. Each scene is linked to its source tasks and released without demonstrations of its own.} The links make intended reuse inspectable; actual novelty and prerequisite mastery still require checkpoint-specific evidence.

Task design also needs measurements that distinguish choice from completion. A robot may select the wrong object or choose correctly but fail to finish. {We therefore record which object a policy commits to, whether that commitment follows an instruction reassigned on an identical initial state, and whether a failure arises at selection or after it.} For world-model-based policies that expose predicted video alongside actions~\citep{bi2025motus}, we can additionally check whether a forecast depicts the requested target and whether execution agrees with it. Agreement can nevertheless preserve the same wrong choice. A useful diagnosis must distinguish correct goal following from both successful execution in isolation and mere sensitivity of outputs or internal signals.

We propose \Benchmark{}, an evaluation-only {diagnostic} benchmark for robotic instruction following under changed requirements (Figure~\ref{fig:overview}). Using RoboTwin~2.0's simulation resources~\citep{chen2025robotwin2}, it specifies 48 scenes in six task families with explicit source-task links. Controlled tests examine responses to changed instructions and layouts, with future-video and internal-intervention evidence where access permits. Separate policy and agent leaderboards retain their respective protocols.\footnote{Project website: {\url{https://embodied-ai-research.github.io/ManipScope_webpage/}}. Code repository: {\url{https://github.com/USTC-AIR-Lab/ManiScope}}.}

\begin{itemize}
\item We introduce \Benchmark{}, an evaluation-only {diagnostic} benchmark with 48 {held-out} scenes across six task families. {Policies train on unmodified RoboTwin~2.0 demonstrations; we link each scene to its source tasks and release no demonstrations of the scenes themselves}, to test skill reuse under changed requests.

\item We pair attribute and target-selection metrics with three diagnostic layers: behavior, future video, and internal interventions. {The diagnostics form a chain in which each step removes one alternative explanation of a failure (Table~\ref{tab:diagnostic-chain}): which object a failed episode moves, whether the choice follows an instruction reassigned on an identical initial state, and whether the scene change or a missing skill explains it.} Our case studies expose wrong-target prediction--action agreement and action sensitivity without established goal-directed transfer where diagnostic access permits.

\item We release scene definitions, success rules, evaluation and diagnostic code, and separate policy and agent leaderboards, with reference results for twelve policy checkpoints and {five agent models}.
\end{itemize}

\section{Related Work}
\label{sec:related}

\paragraph{Task generalization benchmarks.}
Our evaluation-only design relates first to benchmarks that define how test requirements differ from training. CompoSuite combines robots, objects, obstacles, and objectives to evaluate unseen task combinations~\citep{mendez2022composuite}. VIMA uses multimodal prompts and separates several levels of generalization, including new combinations of familiar objects and textures~\citep{jiang2023vima}. LIBERO studies knowledge transfer through task suites with different demands~\citep{liu2023libero}. These works motivate explicit training--evaluation relationships. RoboTwin~2.0 provides the simulation and training foundation for this design: automated data generation, \NumBaseTasks{} bimanual tasks, and domain randomization~\citep{chen2025robotwin2}.

\paragraph{Instruction following and partial progress.}
Beyond task design, instruction-following evaluation must distinguish satisfying a requirement from merely completing a manipulation. ATOM-Bench evaluates atomic skills and held-out compositions in real-world manipulation, including partial progress~\citep{wu2026atom}. Qi et al.\ use instruction-factor preferences to guide data collection, with competing-factor tests that include jointly unsatisfiable instructions~\citep{qi2026scale}. Here the distinction between jointly unsatisfiable instructions and valid candidate targets matters: our attribute measurements concern the latter. Controlled tests make this distinction more direct. LIBERO-CF varies feasible instructions in shared layouts and separates target-contact grounding from task success~\citep{fang2026liberocf}.

{\paragraph{Instruction probes in LIBERO.}
LIBERO-PRO and LIBERO-Plus show that behavior persists under corrupted, blank, or goal-replaced instructions and attribute this to memorized training behavior~\citep{zhou2025liberopro,fei2026liberoplus}. LIBERO-CF scores contact with the instructed object under alternative feasible instructions, shows that policies default to the training task of the scene, and proposes a mitigation~\citep{fang2026liberocf}. Following LIBERO-CF, we trace the default within a task: on three RoboTwin training tasks the object displaced first under every instruction is the one the demonstrations handle first, and in a scene with one fixed instruction the success rate records whether that object is the target (Appendix~\ref{app:libero-relation}, Table~\ref{tab:libero-comparison}).}

\paragraph{Cross-system evaluation and model-access diagnostics.}
\citet{zhang2026wamrobustness} compare VLA and world-action models under visual and language perturbations in LIBERO-Plus and RoboTwin~2.0-Plus. World-action models that jointly generate video and actions~\citep{bi2025motus} permit checks of target agreement across outputs; where internal access permits, representation replacement provides a separate intervention. The latter tests a specified computational connection and requires controls on both the replacement and the measured effect~\citep{zhang2024patching}. Model diversity broadens coverage; the substantive comparison remains the behavior of the evaluated systems under the specified tasks and access-dependent probes.

These studies motivate going beyond aggregate success: completion alone does not distinguish correct target choice from partial satisfaction, and output agreement or sensitivity does not establish correct goal following. \Benchmark{} connects {a held-out, source-linked task suite to per-episode commitment records} and controlled tests under paired instructions and layouts, with future-video and internal-intervention evidence where access permits. The contribution connects source-linked task evaluation with separate measures of choice, completion, forecast correctness, and intervention direction; the model-access probes are optional case-study protocols.

\section[Benchmark Design]{Benchmark Design}
\label{sec:benchmark}

\Benchmark{} tests whether existing manipulation skills can satisfy changed instruction requirements. We connect each evaluation scene to relevant RoboTwin tasks, specify what changes, and fix the resulting tests for comparison.

\subsection[Design Rationale and Scene Construction]{Design Rationale and Scene Construction}

RoboTwin~2.0 provides simulation resources, object assets, and existing robot training tasks~\citep{chen2025robotwin2}. Building on these resources connects new requirements to available manipulation skills: a scene can change the object being handled, its required spatial relation, the description used to select it, or the order of familiar actions. {The training set is RoboTwin~2.0's own: the evaluated checkpoints are trained on unmodified RoboTwin demonstrations (Appendix~\ref{app:checkpoints}). The 48 \Benchmark{} scenes are held out, and we release no demonstrations of them. In the RoboTwin block tasks behind our selection scenes, each task has one fixed goal that its instruction templates paraphrase, so the instruction never has to distinguish between objects; our scenes are built so that it must (Appendix~\ref{app:what-separates}).} The aim is to distinguish applying the requested rule from reproducing a familiar solution.

To construct a scene, we select relevant task components, specify the changed requirements, and instantiate object roles, asset variants, and initial-position regions. Instruction templates state the requested target and goal; physical success conditions specify the required object state, spatial relation, or action order. For example, placing a tea box in a basket and then pressing a bell combines two actions in a required order. Appendix~\ref{app:scene-construction} gives construction examples; Figure~\ref{fig:formal-scene-atlas} shows the complete scene overview.

The source-task mapping records reused components and changed requirements; Appendix~\ref{app:scene-construction} provides the detailed links.

\paragraph{Remark (Scope of generalization claims).}
Source-task links document intended skill reuse and changed requirements; they do not by themselves establish what a checkpoint has learned or previously encountered. Claims of compositional generalization additionally require evidence of component familiarity and the absence of the tested combination from a specified training set.

{\paragraph{Remark (Why held-out scenes, and what they leave open).}
A training scene lets a policy succeed from memory, and an unrelated scene can fail it for reasons other than language. \Benchmark{} takes the middle ground: trained objects and skills, with a new composition and request. A failure can still come from a missing skill, the changed scene, or a choice that ignores the instruction, and a success from a preference that coincides with the single rule of a scene. The diagnostic chain of Section~\ref{sec:diagnostic-method} separates these cases (Table~\ref{tab:diagnostic-chain}).}

\subsection[Task Families and Levels]{Task Families and Levels}

We group the scenes by the requirements changed or combined (D1--D6; Figure~\ref{fig:overview}). D1, skill--object transfer, applies a manipulation skill to a changed object. D2, skill--spatial transfer, requires a specified spatial relation in the evaluation configuration. D3, attribute selection, identifies targets by properties such as color, size, or category. D4, referring expressions, identifies targets through relations, order, or negation. D5, skill sequences, requires actions to satisfy successive goals in the requested order. D6, mixed combinations, combines multiple such requirements. These families organize overlapping task demands; selecting a spatially described object, for example, requires both resolving the reference and manipulating the object.

Levels summarize the designed combination of requirements. L1 comprises scenes with a single replacement, few distractors, and one action segment. L2 comprises scenes with an additional distinction or a two-action task. L3 comprises scenes with combined constraints or dependent actions. We note that these are manual design annotations, not a calibrated difficulty scale: changes in object geometry and execution demands can also affect performance. Appendix~\ref{app:inventory} gives the detailed assignment rules and family--level counts.

The resulting suite specifies which requirements a robot must satisfy and how they relate to existing tasks. The next section defines what task completion and observed choices reveal, then introduces controlled changes to test whether target choices follow the instructions.

\section[Evaluation Metrics and Diagnostic Protocol]{Evaluation Metrics and Diagnostic Protocol}
\label{sec:measurements}

The scene design specifies what is requested. We now define how recorded behavior becomes evidence. A single correct choice does not establish that language determined it.

\subsection{Task Success and Selection Metrics}
\label{sec:selection-metrics}
\label{sec:behavior-method}

\paragraph{Task success rate (SR).}
Let $y_i$ be the binary success label for episode $i$, determined by the requested target, physical goal, reference constraints, and any required action order. Success is recorded when these conditions are satisfied during execution. SR is the fraction of evaluation episodes with $y_i=1$, counting failures and timeouts in the denominator. Appendices~\ref{app:native-success} and \ref{app:ordered-success} summarize the physical success criteria.

For Overall, we average task success rates within each family, then weight the six families equally (Appendix~\ref{app:aggregate-results}). This gives each family the same contribution despite unequal task counts; Pooled instead weights episodes equally.
Pooled success is the fraction of all evaluated episodes that succeed. Family--level rates average tasks in the corresponding subset; global level rates average all tasks at that level.

\paragraph{Selection metrics.}
SR alone does not reveal which object a policy initially selects: an unsuccessful episode may involve a wrong target or failure after a correct choice. We therefore measure the target choice expressed through interaction and, for attribute-selection tasks, which requested properties that choice satisfies. These measurements provide behavioral evidence of target choice rather than a direct readout of latent intention.

\textit{Target selection.} For tasks with a designated target and an observable selection event, let $c_i=1$ when the object selected at that predefined event is the designated target, and zero otherwise. Target-selection correctness is the fraction of episodes with $c_i=1$ among those with complete event records. Ambiguous choices and trajectories without a qualifying event receive zero, but are distinguished from observed wrong-object choices. Missing records are unmeasurable; Appendix~\ref{app:attribute-events} specifies coverage, conditional correctness, and their denominators.

In the formal-task attribute study, selection is measured at {\emph{first contact},} the first loaded gripper--object or hammer--object interaction {of at least 0.1\,N, which also registers incidental touches. We therefore also report \emph{commitment}, the first candidate displaced by at least 5\,cm; it ignores such touches but is undefined for tasks completed without moving a candidate.} Event thresholds, sampling, and sensitivity checks appear in Appendices~\ref{app:attribute-events} and \ref{app:attribute-thresholds}.

\textit{Attribute correctness.} For attribute-selection tasks, we score the requested properties of the selected object. A task is eligible only when its instruction requests the tested attribute and its candidates permit that choice to be scored. The per-scene rate averages correctness over complete event records (Appendix~\ref{app:attribute-events}).
Eligibility is fixed before behavior is observed. Ambiguous or absent selections receive zero; unrequested attributes are not applicable. We report coverage and eligible denominators, weighting eligible tasks equally when aggregating each attribute (Appendix~\ref{app:attribute-task-coverage}).

This refinement retains partial matches that exact-target correctness would miss. Selecting a large red object when a small red object is requested earns color credit, but neither size nor exact-target credit. Color-versus-size comparisons depend on the eligible task designs and are not inferred from aggregate success rates.

Selection--completion outcomes are analyzed jointly, with native success scored independently of the selection event (Appendix~\ref{app:selection-completion}). {For pick-type selection scenes, formal logs give each episode's maximum target and distractor displacements; a failed episode is labeled \emph{distractor only}, \emph{target only}, \emph{both}, or \emph{neither} by which exceeded 5\,cm. The order is not logged, so \emph{both} stays unresolved.}

\subsection[Controlled Language-Following Diagnostics]{Controlled Language-Following Diagnostics}
\label{sec:diagnostic-method}

Controlled contrasts test whether target choices follow changed instructions by varying instructions and layouts independently. Future-video and internal-intervention probes require access to the corresponding outputs or representations. These diagnostic layers concern access to evidence, not a shared serial architecture; unavailable layers are not scored as failures (Figure~\ref{fig:diagnostic-protocols}; Appendix~\ref{app:language-diagnostics}).

{\paragraph{Target reassignment and its controls.}
For a scene with $k$ candidates, each initial state is replayed under $k$ instructions, each designating a different candidate, and under an empty instruction. All-$k$ correctness, the fraction of states on which the commitment follows every instruction, cannot be met by a fixed preference. An \emph{instruction-blind chooser} gives its chance reference; a \emph{positive control}, a tool-using GPT-6 agent on the same scenes, shows that the requests are solvable and gives an upper reference. Two controls test the other explanations: reassignment on unmodified training tasks removes the scene change, and the linked source tasks test the prerequisite manipulation skills (Table~\ref{tab:diagnostic-chain}; Appendix~\ref{app:diagnostic-scope}).}

{\paragraph{Model-access case studies.}
The chain above is behavioral and applies to every policy. Three further readouts are case studies on two checkpoints (Appendix~\ref{app:language-diagnostics}). The \emph{paired study} crosses two target instructions with two layouts in which the red and blue objects are exchanged: instruction-paired correctness requires both instructions to be followed in one layout, and layout-paired correctness requires the requested color in both layouts. For systems that expose predicted images, the \emph{future-video readout} compares the predicted approach target with the instruction and with the later contact target; a forecast and execution can agree on the wrong object. Where representations can be accessed, an \emph{internal intervention} replaces them with those of an alternative-instruction run and measures the change in predicted actions and in donor-goal preference, without closed-loop execution.}

Together, these diagnostics distinguish correct target choice from agreement between prediction and action, and from action sensitivity alone. Behavior provides a common starting point across systems; available forecasts and interventions help identify whether to examine target choice, future prediction, or goal-directed action.

\section{Experiments and Analysis}
\label{sec:evaluation}

We organize the evaluation around three questions that connect benchmark performance to evidence of instruction following:
\begin{enumerate}
\item[\textbf{RQ1.}] How well do current systems complete tasks under changed requirements?
\item[\textbf{RQ2.}] What does task success conceal about target selection and satisfaction of the requested properties?
\item[\textbf{RQ3.}] Do choices correctly track changed requests, and what additional evidence do forecasts and internal interventions provide?
\end{enumerate}

\subsection{Experimental Setup and Reporting}
\label{sec:setup}

We evaluate twelve RoboTwin-trained policy checkpoints, identified with their sources in Table~\ref{tab:main}. Each runs 100 episodes per scene without adaptation on \Benchmark{}. Training backgrounds and control interfaces differ, so comparisons concern the evaluated systems rather than isolated architecture effects. Appendices~\ref{app:checkpoints} and \ref{app:runtime-modes} give checkpoint and inference settings. The evaluated paths bypass WLA\textquotesingle s World Expert and subtask generation, skip future-video generation in FastWAM, and disable chain-of-thought (CoT) and autoregressive action generation in G0.5. {Two rows do not measure instruction following and are excluded from pooled statements about it: X-VLA is conditioned on task-name strings, not instructions, and $\pi_0$ almost never completes its RoboTwin source tasks (Appendix~\ref{app:checkpoints}).}

{Five agents run five episodes per scene at high reasoning (Appendix~\ref{app:agents}). GPT-6.2 Astra uses Responses (4,096 output tokens); Gemini 4 Argon uses \texttt{generateContent} (65,536), since 8,192-token trials exhausted the cap on reasoning before a tool action. For these models and Opus 5.5, the 3,300\,s budget excludes model waiting to avoid latency-driven truncation; all retain the 120-call limit.}

{Scene definitions, instruction pools, and initial states are fixed independently of outcomes and shared across systems} (Appendix~\ref{app:initial-states}). Policies receive three RGB views with model-dependent state and history inputs; agents receive images, measured state, and tool feedback. {GPT-6 Astra and Gemini 3.8 Flash} differ from policy instruction wording in 84 of 240 matched episodes and receive complete instructions on two sequence tasks where policies use stage-dependent instructions; {GPT-6.2 Astra, Opus 5.5, and Gemini 4 Argon use policy wording} (Appendix~\ref{app:matched5-extraction}).

Table~\ref{tab:main} reports the full 4,800-episode policy evaluation; Table~\ref{tab:agent-main} summarizes agent success and timing on the shared first-five subset of 240 episodes per model. Table~\ref{tab:matched5} gives the complete cross-system comparison.

A separate instrumented study measures selection and completion on nine attribute tasks. Five policy checkpoints, {GPT-6 Astra, and Gemini 3.8 Flash} each run five fixed trials per task, yielding 315 episodes (Appendix~\ref{app:attribute-events}). Instructions, initial object states, candidate properties, and the event definition are matched across systems; agent budgets are specified in Appendix~\ref{app:attribute-provenance}. These episodes are analyzed separately from the full-suite and matched-subset results.
\subsection{Task Performance and Selection Outcomes}
\label{sec:attribute-results}
\begin{table}[!t]
\centering\small
\caption{Task success (\%) for the evaluated checkpoints and inference paths by family and level, with 100 episodes per task. Overall equally averages six family means; Pooled uses all 4,800 episodes (Equation~\ref{eq:overall}). Bold marks maxima across policies within each family--level subset or aggregate, including ties, without implying significance. Dashes denote empty subsets; global summaries appear in Appendix~\ref{app:aggregate-results}.}
\label{tab:main}
\setlength{\tabcolsep}{2.5pt}
\renewcommand{\arraystretch}{1.00}
\begin{tabularx}{\textwidth}{@{}Xcrrrrrrrr@{}}
\toprule
Policy & Level & Overall$\uparrow$ & D1$\uparrow$ & D2$\uparrow$ & D3$\uparrow$ & D4$\uparrow$ & D5$\uparrow$ & D6$\uparrow$ & Pooled$\uparrow$ \\
\midrule
\multirow{3}{=}{\raggedright DM0.5\textsuperscript{$\ast$}\\{\scriptsize\citep{dexmal2026dm05}}} & L1 & \multirow{3}{*}{34.14\%} & 57.75\% & \textbf{70.00\%} & 73.50\% & 0.00\% & -- & -- & \multirow{3}{*}{36.21\%} \\
 & L2 &  & 43.17\% & 26.83\% & 71.60\% & 15.33\% & 49.50\% & 41.50\% &  \\
 & L3 &  & -- & 16.00\% & 17.00\% & -- & 0.50\% & 3.60\% &  \\
\specialrule{0.25pt}{2pt}{2pt}
\multirow{3}{=}{\raggedright $\pi_{0.5}$\textsuperscript{$\ast$}\\{\scriptsize\citep{pi2025pi05}}} & L1 & \multirow{3}{*}{31.36\%} & 38.00\% & 43.50\% & 50.00\% & 0.00\% & -- & -- & \multirow{3}{*}{32.65\%} \\
 & L2 &  & 52.00\% & 14.17\% & 56.00\% & 21.50\% & 39.25\% & 39.00\% &  \\
 & L3 &  & -- & 42.00\% & 24.00\% & -- & 17.50\% & 12.40\% &  \\
\specialrule{0.25pt}{2pt}{2pt}
\multirow{3}{=}{\raggedright StarVLA\textsuperscript{$\dagger$}\\{\scriptsize\citep{community2026starvla}}} & L1 & \multirow{3}{*}{27.90\%} & 54.00\% & 59.00\% & 25.50\% & \textbf{9.00\%} & -- & -- & \multirow{3}{*}{29.38\%} \\
 & L2 &  & 44.17\% & 11.00\% & 50.60\% & 13.67\% & 54.00\% & 12.00\% &  \\
 & L3 &  & -- & 36.00\% & 30.00\% & -- & 0.00\% & 2.80\% &  \\
\specialrule{0.25pt}{2pt}{2pt}
\multirow{3}{=}{\raggedright FastWAM\textsuperscript{$\dagger$}\\{\scriptsize\citep{yuan2026fastwam}}} & L1 & \multirow{3}{*}{24.18\%} & 48.25\% & 42.00\% & 70.00\% & 0.00\% & -- & -- & \multirow{3}{*}{26.56\%} \\
 & L2 &  & 43.67\% & 11.67\% & 51.40\% & 11.83\% & 24.50\% & 12.00\% &  \\
 & L3 &  & -- & 14.00\% & 30.50\% & -- & 0.00\% & 0.20\% &  \\
\specialrule{0.25pt}{2pt}{2pt}
\multirow{3}{=}{\raggedright $\pi_0$\textsuperscript{$\ddagger${,\P}}\\{\scriptsize\citep{black2025pi0}}} & L1 & \multirow{3}{*}{12.75\%} & 24.50\% & 5.00\% & 39.00\% & 0.00\% & -- & -- & \multirow{3}{*}{13.33\%} \\
 & L2 &  & 24.50\% & 4.83\% & 17.80\% & 10.50\% & 27.25\% & 0.50\% &  \\
 & L3 &  & -- & 0.00\% & 8.00\% & -- & 0.00\% & 0.00\% &  \\
\specialrule{0.25pt}{2pt}{2pt}
\multirow{3}{=}{\raggedright AHA-WAM\textsuperscript{$\dagger$}\\{\scriptsize\citep{cai2026ahawam}}} & L1 & \multirow{3}{*}{28.63\%} & 60.25\% & 56.50\% & 72.00\% & 0.00\% & -- & -- & \multirow{3}{*}{31.27\%} \\
 & L2 &  & 43.33\% & 12.33\% & 69.60\% & 14.83\% & 28.25\% & 17.50\% &  \\
 & L3 &  & -- & 17.00\% & 30.00\% & -- & 3.00\% & 0.20\% &  \\
\specialrule{0.25pt}{2pt}{2pt}
\multirow{3}{=}{\raggedright X-VLA\textsuperscript{$\ddagger${,\S}}\\{\scriptsize\citep{zheng2025xvla}}} & L1 & \multirow{3}{*}{22.23\%} & 42.75\% & 19.50\% & 66.50\% & 0.00\% & -- & -- & \multirow{3}{*}{22.42\%} \\
 & L2 &  & 25.33\% & 9.33\% & 25.80\% & 10.33\% & 66.75\% & 0.00\% &  \\
 & L3 &  & -- & 11.00\% & 21.50\% & -- & 4.00\% & 1.00\% &  \\
\specialrule{0.25pt}{2pt}{2pt}
\multirow{3}{=}{\raggedright WLA\textsuperscript{$\dagger$}\\{\scriptsize\citep{yang2026wla}}} & L1 & \multirow{3}{*}{25.35\%} & 42.75\% & 46.50\% & 47.50\% & 0.00\% & -- & -- & \multirow{3}{*}{26.71\%} \\
 & L2 &  & 40.00\% & 9.33\% & 48.20\% & 16.17\% & 45.75\% & 15.00\% &  \\
 & L3 &  & -- & 39.00\% & 18.50\% & -- & 0.00\% & 0.00\% &  \\
\specialrule{0.25pt}{2pt}{2pt}
\multirow{3}{=}{\raggedright GalaxeaVLA (G0.5)\textsuperscript{$\ast$}\\{\scriptsize\citep{liu2026g05}}} & L1 & \multirow{3}{*}{\textbf{40.40\%}} & 67.00\% & 63.50\% & 75.50\% & 0.00\% & -- & -- & \multirow{3}{*}{\textbf{42.88\%}} \\
 & L2 &  & 51.50\% & \textbf{41.83\%} & \textbf{78.60\%} & 17.50\% & 48.00\% & 19.50\% &  \\
 & L3 &  & -- & 19.00\% & 31.50\% & -- & 8.50\% & \textbf{24.80\%} &  \\
\specialrule{0.25pt}{2pt}{2pt}
\multirow{3}{=}{\raggedright X-WAM\textsuperscript{$\ast$}\\{\scriptsize\citep{guo2026xwam}}} & L1 & \multirow{3}{*}{37.29\%} & \textbf{86.00\%} & 55.50\% & 74.00\% & 0.00\% & -- & -- & \multirow{3}{*}{40.29\%} \\
 & L2 &  & \textbf{73.17\%} & 11.17\% & 64.20\% & 23.67\% & 42.25\% & 4.50\% &  \\
 & L3 &  & -- & 34.00\% & 36.00\% & -- & 18.50\% & 8.20\% &  \\
\specialrule{0.25pt}{2pt}{2pt}
\multirow{3}{=}{\raggedright Motus\textsuperscript{$\ast$}\\{\scriptsize\citep{bi2025motus}}} & L1 & \multirow{3}{*}{35.03\%} & 50.75\% & 62.00\% & \textbf{84.50\%} & 0.00\% & -- & -- & \multirow{3}{*}{37.00\%} \\
 & L2 &  & 51.33\% & 35.83\% & 53.00\% & 15.67\% & {\textbf{59.00\%}} & 10.00\% &  \\
 & L3 &  & -- & 17.00\% & 39.00\% & -- & 6.00\% & 7.00\% &  \\
\specialrule{0.25pt}{2pt}{2pt}
\multirow{3}{=}{\raggedright LingBot-VA\textsuperscript{$\ast$}\\{\scriptsize\citep{li2026lingbotva}}} & L1 & \multirow{3}{*}{38.29\%} & 60.50\% & 56.50\% & 72.00\% & 6.00\% & -- & -- & \multirow{3}{*}{39.92\%} \\
 & L2 &  & 52.67\% & 16.00\% & 59.40\% & \textbf{28.50\%} & 45.75\% & \textbf{42.00\%} &  \\
 & L3 &  & -- & \textbf{53.00\%} & \textbf{53.50\%} & -- & \textbf{27.50\%} & 9.80\% &  \\
\bottomrule
\end{tabularx}
\par\smallskip
\begin{minipage}{\textwidth}\footnotesize
Level subsets contain different tasks and do not isolate a single difficulty factor. Training provenance: \textsuperscript{$\ast$}robot/latent-action pretraining followed by RoboTwin adaptation; \textsuperscript{$\dagger$}pretrained backbones followed by RoboTwin action training, with no additional robot-action pretraining reported; \textsuperscript{$\ddagger$}checkpoint lineage unverified. {\textsuperscript{\S}Conditioned on task-name strings; \textsuperscript{\P}near-zero success on RoboTwin source tasks: these rows do not measure instruction following and are not marked as maxima (Section~\ref{sec:setup}).} Details: Appendix~\ref{app:checkpoints}; task counts: Appendix~\ref{app:inventory}.
\end{minipage}
\end{table}

For RQ1, the full-suite results show uneven competence across task requirements. G0.5 attains the highest observed Overall score, 40.40\% (Table~\ref{tab:main}), but does not lead every family. Referring expressions (D4) and mixed combinations (D6) remain difficult across the evaluated policies, even when performance is substantially stronger in other families (Appendix~\ref{app:aggregate-results}). The family breakdown therefore matters beyond ranking: it identifies requirements under which an otherwise competitive system remains unreliable. These scores describe performance on the changed requirements; without matched source-task evaluations and checkpoint-specific training coverage, they do not measure a loss from mastered components to unseen combinations. {Nor, for the selection families, do they measure attribute or reference resolution: there, all-$k$ correctness (Table~\ref{tab:reassignment}) is the instruction-following score, and success in the other families measures skill transfer.}

\begin{table}[!htbp]\centering\small
\caption{Agent success and runtime on the shared first-five-seed subset (240 episodes per model, including failures). Overall averages the six task families equally; Pooled averages episodes. Bold marks the highest success and lowest time, including ties. $^{*}$Instruction-delivery differences: Section~\ref{sec:setup}. Full system results and timing definitions: Table~\ref{tab:matched5-complete} and Appendix~\ref{app:timing}.}\label{tab:agent-main}
\setlength{\tabcolsep}{4pt}
\begin{tabularx}{\textwidth}{@{}Xrrrrr@{}}\toprule
Agent & Overall$\uparrow$ & Pooled$\uparrow$ & \shortstack{Calls\\(/ep.)} & \shortstack{Model$\downarrow$\\(s/call)} & \shortstack{Episode wall$\downarrow$\\(s/ep.)}\\\midrule
GPT-6 Astra$^{*}$ & \textbf{76.35\%} & \textbf{77.08\%} & 24.9 & 13.248 & 730.4 \\
{GPT-6.2 Astra} & {75.65\%} & {76.25\%} & {15.8} & {\textbf{12.429}} & {\textbf{284.5}} \\
{Opus 5.5} & {62.76\%} & {62.08\%} & {30.2} & {21.160} & {835.3} \\
{Gemini 4 Argon} & {29.48\%} & {29.58\%} & {15.3} & {125.558} & {2826.0} \\
Gemini 3.8 Flash$^{*}$ & 24.35\% & 24.17\% & 35.2 & 46.260 & 2002.2 \\
\bottomrule\end{tabularx}
\end{table}

On the shared subset of five episodes per scene, GPT-6 Astra attains 76.35\% Overall and the best policy, G0.5, 42.30\%, under different interfaces (Appendices~\ref{app:matched-sensitivity},~\ref{app:timing}). Among agents, GPT-6 Astra and GPT-6.2 Astra attain the highest Overall scores, while GPT-6.2 Astra has the shortest mean episode time (Table~\ref{tab:agent-main}). These comparisons reflect the observation, instruction, and budget settings in Section~\ref{sec:setup}.

RQ2 asks what these completion scores leave unresolved. {Table~\ref{tab:diagnostic-chain} lists the diagnostic steps. \textbf{Where failures arise.} Of the 9,587 failures among the 13,200 formal episodes on the eleven pick-type selection scenes, 84.6\% displace a distractor, 3.0\% the target alone, and 12.4\% nothing (Table~\ref{tab:failure-decomposition}); without X-VLA and $\pi_0$, a distractor is displaced in 88.3\%. The profile separates scenes that success rates do not: both below 10\% success, \nolinkurl{ref_block_on_plate} has 71.8\% of failures displacing only a distractor and \nolinkurl{ref_block_behind_can} 60.1\% displacing nothing (Appendix~\ref{app:failure-decomposition}).}

\begin{table}[t]
\centering\footnotesize\color{black}
\setlength{\tabcolsep}{4pt}
\renewcommand{\arraystretch}{1.08}
\caption{{The diagnostic chain. Each step tests one alternative explanation of a failure.}}
\label{tab:diagnostic-chain}
\begin{tabularx}{\textwidth}{@{}p{0.255\textwidth}p{0.325\textwidth}X@{}}
\toprule
Question & Measurement & Finding \\
\midrule
Where does a failure arise? & Object displaced in failed episodes; contact class from the gripper state & A distractor is displaced in 84.6\% of failures; a non-target is grasped in 83\% of failed reruns \\
Does the choice follow the instruction? & Target reassignment on frozen held-out states (Table~\ref{tab:reassignment}) & Every instruction is followed on 3.9\% of states; instruction-blind chooser 16\%, positive control 88\% \\
Is the scene change the cause? & The same reassignment on unmodified training tasks & No: five of six checkpoints displace the requested blue block first in 0/100 runs \\
Is a missing manipulation skill the cause? & Linked source tasks under training instructions & No: nine of ten checkpoints complete at least nine of the ten tasks \\
Can the selection be supplied? & An agent selects the target and the policy executes & For one checkpoint, every instruction is followed on 7/10, 6/10, and 3/10 states (three scenes), from 0/10 \\
\bottomrule
\end{tabularx}
\end{table}

In the separate attribute study, 50/315 episodes (15.9\%) select the correct target first but do not complete the task, with such outcomes observed among both policies and agents (Figure~\ref{fig:attributes}; Table~\ref{tab:selection-completion-counts}). Failure thus need not begin with a wrong-target choice. Conversely, even an incorrect target can satisfy part of the request: on the conjunctive color--category task, 14/35 episodes have a first interaction that matches only one requested property (Table~\ref{tab:attribute-binding}). Completion alone does not distinguish these partial matches from other initial choices, or identify failures after a correct choice, although these distinctions pose different questions for diagnosis.

{\textbf{Is a recorded selection a choice?} First contact cannot tell choosing a wrong object from touching one in passing: in 15 of the 21 non-target/success episodes the touched object moved less than 5\,cm. In 2,400 reruns with gripper-state records, 83\% of failed episodes grasp a non-target and 5.8\% of successful ones do, so a wrong selection in a failed episode is usually a wrong grasp. On the six pick-type tasks, 47/210 episodes fail after a target first contact and 39/210 after a target commitment (Appendix~\ref{app:displacement-rescoring}).}

\subsection[Results of the Language-Following Diagnostics]{{Results of the} Language-Following Diagnostics}
\label{sec:language-results}
A remaining ambiguity motivates RQ3: a correct choice under one instruction could still reflect a preference for a familiar object or location. We therefore test whether target choices change to match the revised request.

{\textbf{Target reassignment.} Target reassignment covers ten pick-type selection scenes and twelve checkpoints (Table~\ref{tab:reassignment}, Figure~\ref{fig:probe-overview}). Ten checkpoints (Section~\ref{sec:setup}) follow every instruction on 39 of 1,000 initial states (3.9\%), below the 16\% chance level of the instruction-blind chooser, and displace the same object under every instruction on 68\% (chooser: 33\%): the choices are systematic, not random. The positive control, a GPT-6 agent, follows every instruction on 44 of 50 states (88\%): the requests are solvable, and a system that follows them reaches this level. LingBot-VA and $\pi_{0.5}$, trained on the same reported demonstrations, follow color (10/10 and 6/10 states of the three-color scene). Formal success in such scenes therefore records whether the preferred object happens to be the target, not attribute or reference resolution: 77.0\% in the scene that requests the smaller block, which is preferred, and 1.2\% in the scene that requests the block on the plate, which is not (Table~\ref{tab:failure-decomposition}).}

\begin{figure}[t]
 \centering
 \includegraphics[width=\textwidth]{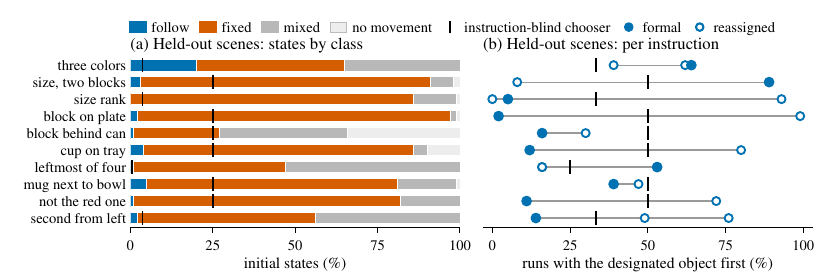}
\caption{{Target reassignment on the ten held-out selection scenes, ten checkpoints pooled (100 initial states per scene). (a) Initial states by class; ticks mark the \emph{follow} share of an instruction-blind chooser. (b) Runs in which the designated object is displaced first, one marker per instruction; ticks mark $1/k$.}}
 \label{fig:probe-overview}
\end{figure}

{\textbf{Controls.} Two controls argue against the scene change and against missing manipulation skills. On unmodified RoboTwin training tasks, five of six verified checkpoints never displace the blue block first when asked to (0/100 runs; $\pi_{0.5}$: 15/20), and on a size task eight displace the requested largest block first in 1/160 runs. In all three tasks the preferred object is the one the scripted demonstrations handle first, as predicted before the size data (Figure~\ref{fig:probe-training}, Appendix~\ref{app:reassignment-controls}). Under training instructions, nine of the ten checkpoints complete at least nine of the ten linked source tasks in 70\% or more of episodes, and without the six checkpoint--scene cells with a linked task below 30\%, every instruction is followed on 3.1\% of states. On LIBERO-Object, two LIBERO checkpoints follow a reassigned instruction between the two trained target positions (99--100\% of runs) and rarely elsewhere (49/400, 0/400; Appendix~\ref{app:libero-relation}).}

\begin{figure}[t]
\centering
\includegraphics[width=\textwidth]{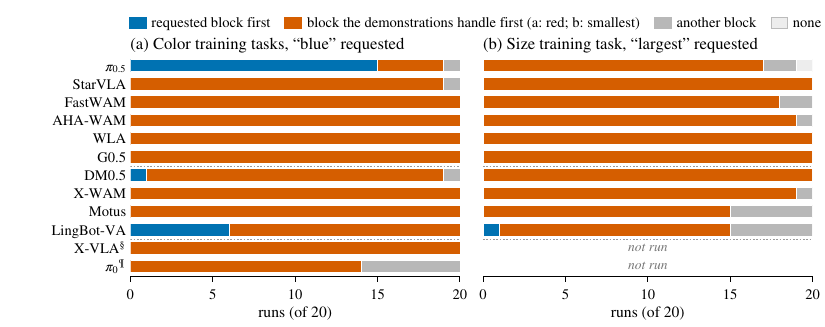}
\caption{{Target reassignment on unmodified RoboTwin training tasks: the block displaced first (gray: another block or none). (a) Under ``Pick up the blue block.'' on the two color training tasks. (b) Under ``Pick up the largest block.'' on the size training task. Row markers and groups as in Table~\ref{tab:training-scene}.}}
\label{fig:probe-training}
\end{figure}

\begin{table}[t]
\centering\footnotesize\color{black}
\setlength{\tabcolsep}{4.5pt}
\caption{{Target reassignment on ten pick-type selection scenes: share (\%) of initial states in each class (ten states per scene; 95\% Wilson interval for \emph{follow}). \emph{Instruction-blind chooser}: expectation for a uniform choice among the $k$ candidates that ignores the instruction. \emph{Positive control}: the GPT-6 agent on five initial states per scene, as in the agent evaluation; it shows that the requests are solvable and gives an upper reference. Excluded from pooled statements: $^{\S}$conditioned on task-name strings; $^{\P}$near-zero success on the source tasks (Appendix~\ref{app:checkpoints}).}}
\label{tab:reassignment}
\begin{tabular}{@{}lrrrrr@{}}
\toprule
System & States & Follow & Fixed & No move & Mixed \\
\midrule
LingBot-VA & 100 & 17 (11--26) & 50 & 2 & 31 \\
StarVLA & 100 & 9 (5--16) & 52 & 8 & 31 \\
$\pi_{0.5}$ & 100 & 8 (4--15) & 56 & 2 & 34 \\
Motus & 100 & 2 (1--7) & 72 & 4 & 22 \\
X-WAM & 100 & 1 (0--5) & 87 & 1 & 11 \\
AHA-WAM & 100 & 1 (0--5) & 71 & 6 & 22 \\
G0.5 & 100 & 1 (0--5) & 64 & 6 & 29 \\
WLA & 100 & 0 (0--4) & 90 & 5 & 5 \\
FastWAM & 100 & 0 (0--4) & 72 & 10 & 18 \\
DM0.5 & 100 & 0 (0--4) & 65 & 5 & 30 \\
\midrule
Ten checkpoints pooled & 1000 & 4 (3--5) & 68 & 5 & 23 \\
\midrule
X-VLA$^{\S}$ & 100 & 0 (0--4) & 69 & 9 & 22 \\
$\pi_0$$^{\P}$ & 100 & 0 (0--4) & 44 & 17 & 39 \\
\midrule
Positive control (GPT-6 agent) & 50 & 88 (76--94) & 2 & 0 & 10 \\
Instruction-blind chooser & & 16 & 33 & & \\
\bottomrule
\end{tabular}
\end{table}

{\textbf{Model-access case studies.}} We instantiate the diagnostic protocol with $\pi_{0.5}$ as a VLA example and Motus as a video-generating world-action example. These checkpoint-specific case studies illustrate the protocol rather than establish class-wide behavior.

{In the paired study, Motus succeeds in 40.0\% of individual trials, yet follows both instructions in only 1/40 same-layout pairs; $\pi_{0.5}$ does so in 29/40 (Figure~\ref{fig:language-diagnostics}, Appendix~\ref{app:language-paired}): repeating successful behavior is insufficient if the requested target can change.} {The same mismatch is visible in the predicted future: all 77 readable Motus forecasts (of 80) agree with the later contact target, 46 of them on the wrong object (Figure~\ref{fig:language-diagnostics}; Appendix~\ref{app:language-future}). Of twelve checkpoints, only LingBot-VA's first motion fully heads for the named color (Figure~\ref{fig:first-decision}).} {Internal intervention on Motus's future representations changes predicted actions without establishing goal-directed transfer: the mean donor-goal preference change is 0.560\,mm (95\% interval $[-0.019,1.299]$\,mm), and only two readable contrasts predict different targets (Appendix~\ref{app:language-internal}).}

{\textbf{From diagnosis to remedy.} The chain locates the failure at target selection, and supplying the selection recovers part of it. When a vision--language agent selects the target and DM0.5 executes, every instruction is followed on 7/10, 6/10, and 3/10 states of three scenes (0/10 for DM0.5 alone; Appendix~\ref{app:cause-hypotheses}).}

Together, these case studies identify a recurring diagnostic pitfall: successful episodes, consistent forecasts, and responsive actions each leave correct language following unresolved. Evidence must be tied to the requested goal at each observed stage. This is the contribution of the diagnostic analysis: narrowing what to investigate next, rather than inferring a complete failure mechanism from one score or signal (Appendix~\ref{app:diagnostic-interpretation}).

\section[Limitations]{Limitations}
\label{sec:limitations}

{A wrong-object commitment does not distinguish misperceiving the candidates from disregarding the instruction. Two scene-specific success predicates accept a non-target grasped with the target (Appendix~\ref{app:displacement-rescoring}). The demonstration-order account (Appendix~\ref{app:cause-hypotheses}) is observational, without retraining; the remedies are tested on one checkpoint and are illustrative. The findings concern checkpoints trained on RoboTwin~2.0, whose block tasks never require the instruction to choose among objects.}

Our internal interventions are evaluated in open loop: we measure changes in predicted actions and relative target-distance preference under commanded kinematics without executing the intervened policy. These readouts do not establish improved target selection or task completion.

The benchmark and diagnostic studies are conducted entirely in simulation. Their findings have not been validated on physical robots, where perception noise, contact dynamics, and hardware constraints may affect both task performance and diagnostic measurements. Extending the tasks and protocols to real robots is therefore needed to assess their applicability beyond simulation.

\Needspace{4\baselineskip}
\section{Conclusion}
\label{sec:conclusion}

\Benchmark{} tests robotic instruction following under changed requirements through {held-out scenes with explicit source-component links, for policies trained on unmodified RoboTwin~2.0 demonstrations}. Joint selection and success measurements distinguish partial instruction satisfaction, correct choice, and completion{, and a chain of controls indicates where failures arise and what does not explain them}. Controlled diagnostics then test whether choices track changed requests, with future-video and internal-intervention evidence where access permits. Our studies show why successful execution, prediction--action agreement, and action sensitivity each leave aspects of goal following unresolved. Together, the benchmark and diagnostics support more informative system comparisons and targeted investigation of language-following failures.

\clearpage
\endgroup
\newpage
\section*{Reproducibility Statement}

The project website ({\url{https://embodied-ai-research.github.io/ManipScope_webpage/}}) provides task documentation and separate policy and agent leaderboards. The code repository is {\url{https://github.com/USTC-AIR-Lab/ManiScope}}. Appendices~\ref{app:code-design}, \ref{app:implementation}, and~\ref{app:results} specify task construction and success criteria, evaluation configurations, and per-task results. Appendices~\ref{app:attribute-selection} and~\ref{app:language-diagnostics} describe the selection metrics and controlled diagnostic procedures. {Appendices~\ref{app:diagnostic-scope} and~\ref{app:libero-relation} document target reassignment, its controls, the remedy tests, and the LIBERO-Object experiment.}

\section*{Ethics Statement}

All experiments are conducted in simulation and do not establish real-world safety or reliability. Use and redistribution of inherited simulation assets and model resources remain subject to their original licenses and terms.

\section*{AI Use Statement}

OpenAI Codex assisted with code review and manuscript polishing. {AI coding assistants also assisted with writing code for the experiments and analyses.} {GPT-6 Astra, GPT-6.2 Astra, Opus 5.5, Gemini 3.8 Flash, and Gemini 4 Argon were evaluated as tool-using agents for robotic manipulation in the experiments.} The authors are responsible for the experimental design, verification of results, and all claims and content in the paper.

\bibliography{references}
\bibliographystyle{iclr2027_conference}

\clearpage
\appendix
\renewcommand{\topfraction}{0.92}\renewcommand{\bottomfraction}{0.85}\renewcommand{\textfraction}{0.06}\renewcommand{\floatpagefraction}{0.7}
\setcounter{topnumber}{4}\setcounter{bottomnumber}{3}\setcounter{totalnumber}{6}
\noindent\textbf{Appendix guide.}
Appendix~\ref{app:code-design} gives task definitions and success criteria; Appendix~\ref{app:implementation} specifies evaluated systems and timing. Appendix~\ref{app:aggregate-results} collects aggregate task scores, followed by formal-task attribute analyses in Appendix~\ref{app:attribute-selection} and controlled diagnostics in Appendix~\ref{app:language-diagnostics}. The complete per-task result tables are collected last in Appendix~\ref{app:results}. {The revision keeps this numbering and adds the validation of the selection event and the failure decomposition (Appendices~\ref{app:displacement-rescoring},~\ref{app:failure-decomposition}), target reassignment with its controls (Appendix~\ref{app:diagnostic-scope}), and the relation to LIBERO-based probes (Appendix~\ref{app:libero-relation}).}

\section{Task Inventory, Construction, and Success Criteria}
\label{app:code-design}

This appendix collects the task inventory, construction examples, source-task links, initial-state procedure, and physical success criteria used by the benchmark.

\subsection{Scene Inventory and Design Levels}
\label{app:inventory}

Table~\ref{tab:inventory} lists the 48 evaluation scenes by task family and level. D1--D6 denote skill--object transfer, skill--spatial transfer, attribute selection, referring expressions, skill sequences, and mixed combinations.

\begin{table}[h]
\centering\small
\caption{Scene counts by task family and level.}
\label{tab:inventory}
\begin{tabularx}{\textwidth}{lXrrrr}
\toprule
ID & Family & L1 & L2 & L3 & Total \\
\midrule
D1 & Skill--object transfer & 4 & 6 & 0 & 10 \\
D2 & Skill--spatial transfer & 2 & 6 & 1 & 9 \\
D3 & Attribute selection & 2 & 5 & 2 & 9 \\
D4 & Referring expressions & 1 & 6 & 0 & 7 \\
D5 & Skill sequences & 0 & 4 & 2 & 6 \\
D6 & Mixed combinations & 0 & 2 & 5 & 7 \\
\midrule
& Total & 9 & 29 & 10 & 48 \\
\bottomrule
\end{tabularx}
\end{table}

L1 denotes single changes with few distractors and one action segment; L2 introduces additional distinctions or two-action tasks; L3 combines multiple constraints or dependent actions. The manual assignments comprise 9/29/10 scenes at L1/L2/L3. These annotations summarize designed requirements, not calibrated difficulty: object geometry and execution demands can also affect performance.

The scene atlas (Figure~\ref{fig:formal-scene-atlas}) shows the task configurations. Figure~\ref{fig:task-dependencies} shows the scene identifiers and their source-task links. Execution limits are specified in Appendix~\ref{app:implementation}.

\subsection{Scene Construction and Source-Task Links}
\label{app:scene-construction}

Each scene specifies object roles, assets or geometry, sampling regions, spatial relations, and a goal predicate. Task-dependent slot permutations and relative placements preserve the relation used to identify the requested target.

\begin{figure}[p]
\centering
\includegraphics[width=\textwidth,height=0.83\textheight,keepaspectratio]{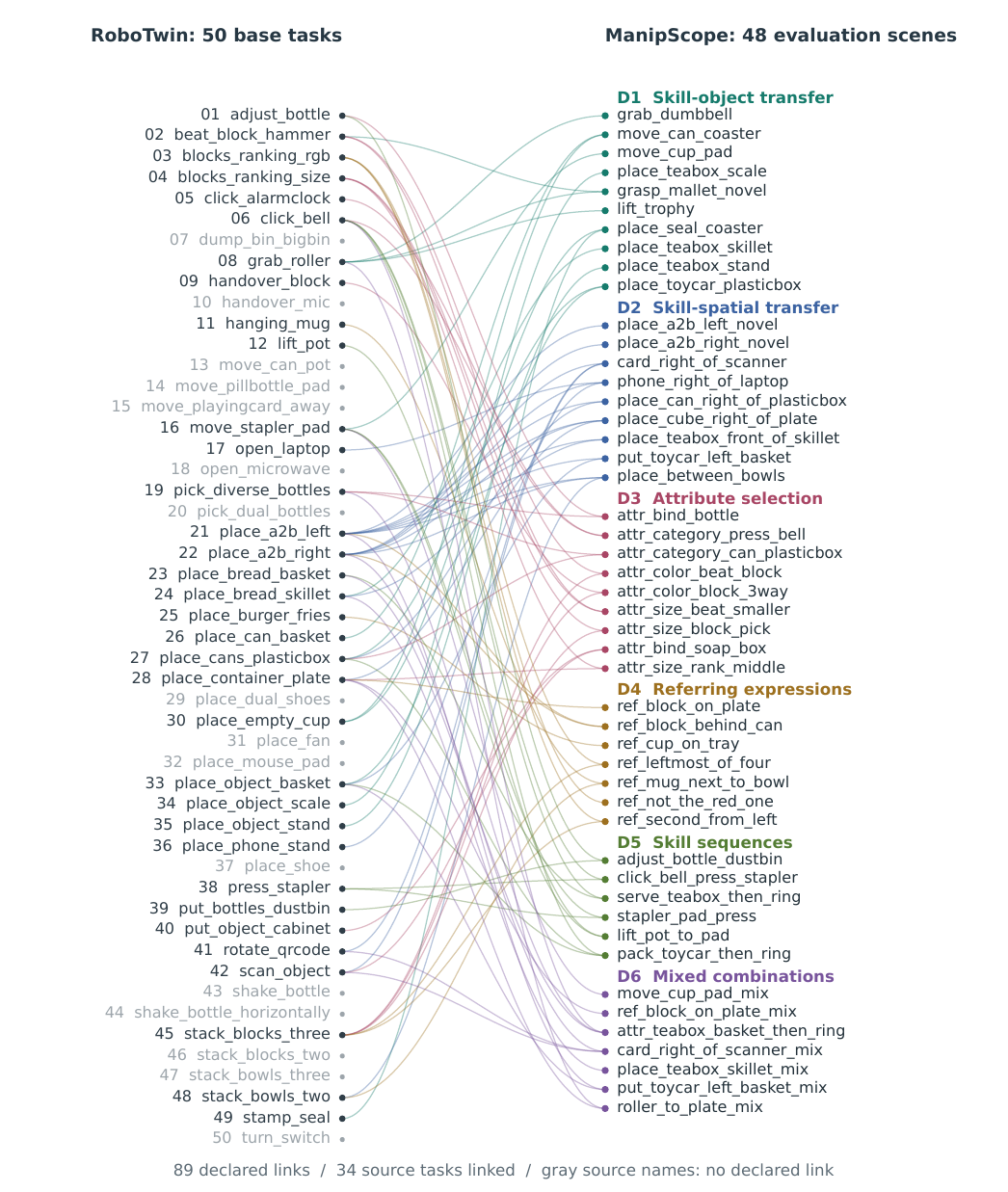}
\caption{Declared task-component links: 48 \Benchmark{} scenes connect to 34 RoboTwin source tasks through 89 links. Edge colors identify task families; gray source names have no declared link. Links describe declared component reuse, not verified training exposure.}
\label{fig:task-dependencies}
\end{figure}

For scene $c$, the source-task set $P_c\subseteq\mathcal{B}$ records its links to RoboTwin base tasks $\mathcal{B}$, together with reused components and changed requirements. The 48 scenes link to 34 of the 50 RoboTwin task identifiers through reused objects, skills, and relations (Figure~\ref{fig:task-dependencies}). This is design metadata, not an execution plan supplied to the system. The links alone do not establish checkpoint-specific training exposure or mastery of the reused components.

Six concrete examples illustrate the construction:
\begin{enumerate}
\item D1, \texttt{place\_teabox\_scale}: reuse the scale-placement task and substitute a tea box as the manipulated object; success uses the shared supported-on predicate.
\item D2, \texttt{card\_right\_of\_scanner}: combine card/scanner handling with the right-of relation. Its goal requires a rightward separation between 5 and 22\,cm, with lateral and height tolerances and released grippers.
\item D3, \texttt{attr\_color\_block\_3way}: place three equal-size red, green and blue blocks in permuted slots and request the red block in the formal scene. Lifting requires 6\,cm clearance together with gripper contact and selection guards. The two-instruction diagnostic study explicitly changes the requested target; it is a separate protocol. {The formal scene always requests red, so a red-block preference suffices for formal success there, as a fixed preference can in any scene with one requested rule; target reassignment (Appendix~\ref{app:reassignment}) removes this confound. Not every scene fixes the requested value: the bottle scene requests the brown or the green bottle at random in each episode, and the soap/box scene draws one of several color--category triples.}
\item D4, \texttt{ref\_block\_on\_plate}: use identical-color blocks and identify the target by its initial placement on a plate. The target's initial pose is specified relative to the plate; the goal is lifting it with 6\,cm clearance.
\item D5, \texttt{serve\_teabox\_then\_ring}: require placing the tea box in the basket, then pressing the bell. The native sequence state records progress and the final predicate retains the placement requirement.
\item D6, \texttt{attr\_teabox\_basket\_then\_ring}: add a distractor object to the placement-and-bell sequence. The scene combines target discrimination, placement and ordered action requirements.
\end{enumerate}

\begin{figure}[p]
\centering
\includegraphics[width=\textwidth,height=0.86\textheight,keepaspectratio]{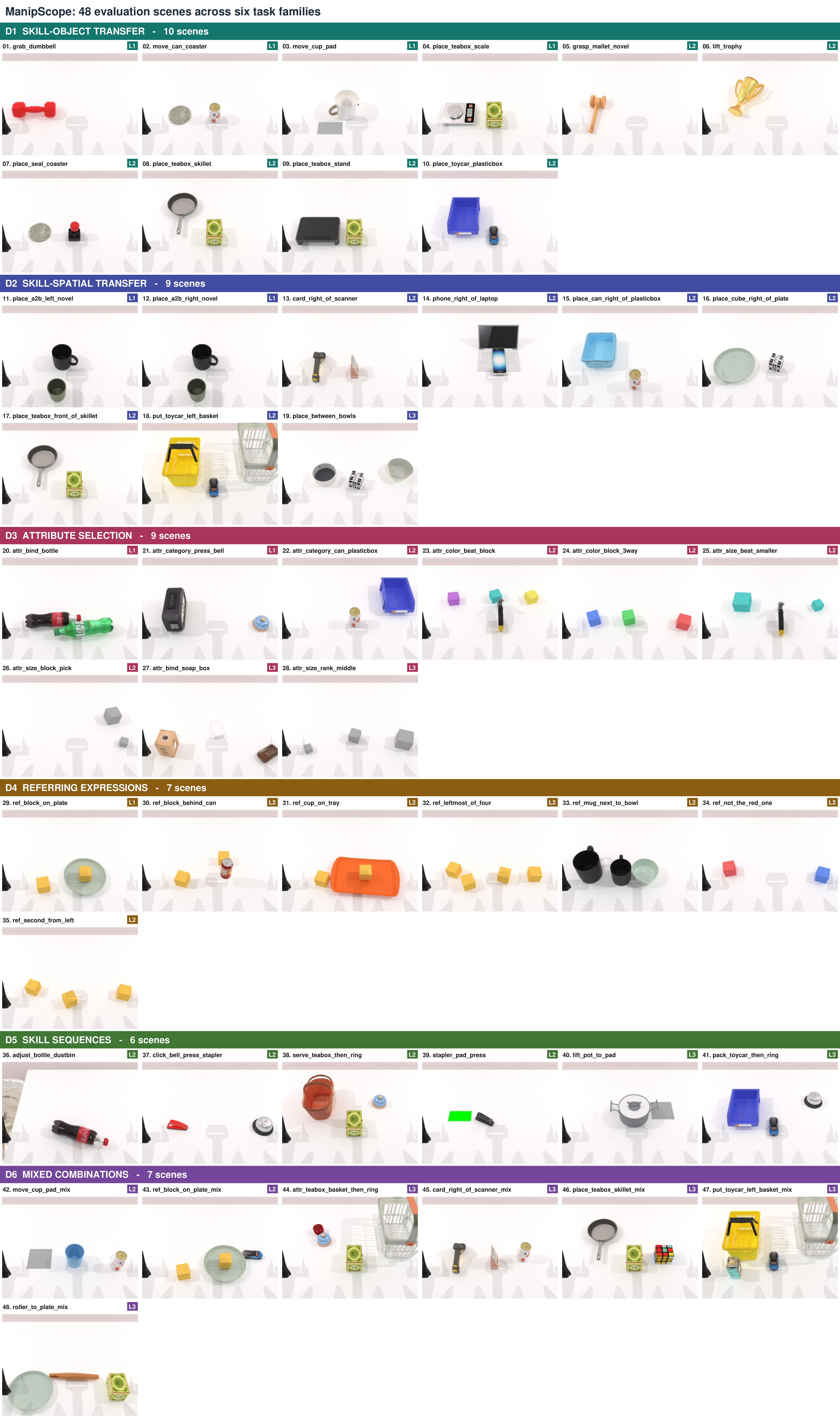}
\caption{All 48 evaluation scenes, grouped into D1--D6, with task identifiers and assigned levels.}
\label{fig:formal-scene-atlas}
\end{figure}

Figure~\ref{fig:formal-scene-atlas} shows the 48 evaluation scenes, grouped by task family; Figure~\ref{fig:task-dependencies} then shows their declared source-task links.

\subsection{Initial States and Instructions}
\label{app:initial-states}

A scripted expert checks task feasibility before initial states are fixed for evaluation. These checks support evaluation construction; expert demonstrations for training on the evaluation scenes are deliberately not part of the release. {The evaluated checkpoints are trained on the unmodified RoboTwin~2.0 demonstrations (Appendix~\ref{app:checkpoints}); the scenes are held out from that training.}

The evaluated system starts from a reset scene with an instruction sampled from the task\textquotesingle s configured template pool.

The organizers retain the corresponding seeds privately to protect public-evaluation fairness. Evaluation reuses the organizer-managed private seed lists. Trial counts, environment settings, and the matched subset are specified in Appendix~\ref{app:implementation}.

\subsection{Native Success Predicates}
\label{app:native-success}

For shared scene predicates, a movement goal requires target displacement above 5\,cm and greater than distractor displacement. The first object crossing the threshold must be the target; simultaneous crossings are resolved by greatest displacement. Tools and goal references are excluded from the distractor set. Reference objects must remain within 5\,cm of their initial positions.

Press and beat goals use location/contact conditions for the target and reject distractor displacement of 5\,cm or more.

Lifting checks object clearance and gripper contact. Placement checks relative location, gripper release, support/container contact, and low object speed (below 0.08\,m/s linear and 1\,rad/s angular). Containment also checks container orientation and geometry. Relative placement uses signed offsets and scene-specific tolerances. Pressing uses gripper proximity to the designated location; hammer goals require tool--target contact.

Task-specific scenes define their own selection guards. Bottle selection uses initial displacement; soap/box selection uses lifting. {Both are more permissive than the shared predicate and can score a success when a non-target is grasped along with the target (Appendix~\ref{app:displacement-rescoring}).} The evaluator records success when the task predicate is first satisfied.

\subsection{Success Conditions for Ordered Tasks}
\label{app:ordered-success}

\begin{table}[t]
\centering\small
\caption{Success conditions for ordered tasks. Distances are in meters; shared sequence scenes also apply target-selection and reference-stability guards.}
\label{tab:implemented-stages}
\begin{tabularx}{\textwidth}{@{}p{0.26\textwidth}XX@{}}
\toprule Task & First condition & Later/final condition \\
\midrule
\nolinkurl{adjust_bottle_dustbin} & Bottle functional point $x<-0.15$, $z>0.9$; upward long-axis cosine $>0.85$. & Recorded stage 1, then bottle origin inside the specified bin XY box and $0.2<z<0.7$; bin condition holds at success. \\[4pt]
\nolinkurl{click_bell_press_stapler} & Bell alignment and gripper proximity to its press point. & Recorded stage 1, then proximity to the stapler press point; stage 2 is latched. \\[4pt]
\nolinkurl{stapler_pad_press} & Stapler within $(0.055,0.035,0.02)$ of pad origin; both grippers open. & Recorded stage 1, then stapler press-point proximity while still at the pad; stage 2 is latched. \\[4pt]
\nolinkurl{serve_teabox_then_ring} & Supported placement of tea box in basket. & Later bell press; placement retained at success. \\[4pt]
\nolinkurl{pack_toycar_then_ring} & Supported placement of toy car in box. & Later bell press; placement retained at success. \\[4pt]
\nolinkurl{attr_teabox_basket_then_ring} & Select tea box amid a distractor; supported placement in basket. & Later bell press; placement retained at success (D6). \\[4pt]
\nolinkurl{lift_pot_to_pad} & Maximum pot-origin rise exceeds $0.06$; stored as a historical condition. & Final XY offset from pad below $(0.09,0.09)$, pot above pad, grippers open. Uses a stored maximum height. \\[4pt]
\bottomrule
\end{tabularx}
\end{table}

Ordered tasks advance through their predicates sequentially. Final success also requires the remaining placement obligations, target-selection guards, and reference stability. Table~\ref{tab:implemented-stages} lists task-specific conditions.

\section{Evaluation Configuration}
\label{app:implementation}

Evaluation uses the ALOHA--AgileX embodiment with head and two wrist cameras. Background, lighting, table-height, and camera-distance randomization are disabled; object placement and assets follow the scene definition. Budgets are 600 native action calls for \nolinkurl{move_cup_pad}, \nolinkurl{place_a2b_left_novel}, \nolinkurl{attr_bind_bottle}, \nolinkurl{attr_bind_soap_box}, \nolinkurl{adjust_bottle_dustbin}, \nolinkurl{click_bell_press_stapler}, and \nolinkurl{stapler_pad_press}; the other 41 scenes use 1,000 native action calls. Each call submits one target to the simulator controller, which can execute multiple physical steps; these are not model-inference calls. The evaluator counter is also subject to the agent termination behavior described below.

\subsection{Checkpoints and Training Resources}
\label{app:checkpoints}

Table~\ref{tab:model-sources} identifies the evaluated checkpoints and their implementations.

\begingroup
\small\setlength{\tabcolsep}{3pt}
\begin{longtable}{p{0.18\textwidth}p{0.47\textwidth}p{0.25\textwidth}}
\caption{Weights and implementations for the twelve evaluated policy checkpoints.}\label{tab:model-sources}\\
\toprule
Model & Weights (Hugging Face) & Code (GitHub) \\
\midrule
\endfirsthead
\multicolumn{3}{l}{Table \thetable{} continued: model sources.}\\
\toprule
Model & Weights (Hugging Face) & Code (GitHub) \\
\midrule
\endhead
\midrule\multicolumn{3}{r}{Continued on next page.}\\
\endfoot
\bottomrule\endlastfoot
\multicolumn{3}{l}{\textit{Evaluated: twelve checkpoints}}\\
DM0.5 & \href{https://huggingface.co/Dexmal/DM05-robotwin2}{\nolinkurl{Dexmal/DM05-robotwin2}} & \href{https://github.com/dexmal/opendm}{\nolinkurl{dexmal/opendm}} \\
$\pi_{0.5}$ & \href{https://huggingface.co/motus-robotics/pi0.5_robotwin2}{\nolinkurl{motus-robotics/pi0.5_robotwin2}} & \href{https://github.com/Physical-Intelligence/openpi}{\nolinkurl{Physical-Intelligence/openpi}} \\
StarVLA & \href{https://huggingface.co/StarVLA/Qwen3-VL-OFT-RoboTwin2-All}{\nolinkurl{StarVLA/Qwen3-VL-OFT-RoboTwin2-All}} & \href{https://github.com/starVLA/starVLA}{\nolinkurl{starVLA/starVLA}} \\
FastWAM & \href{https://huggingface.co/yuanty/fastwam}{\nolinkurl{yuanty/fastwam}} & \href{https://github.com/yuantianyuan01/FastWAM}{\nolinkurl{yuantianyuan01/FastWAM}} \\
$\pi_0$ & \href{https://huggingface.co/yuffish/robotwin-pi0_aloha-agilex_clean-50}{\nolinkurl{yuffish/robotwin-pi0_aloha-agilex_clean-50}} & \href{https://github.com/Physical-Intelligence/openpi}{\nolinkurl{Physical-Intelligence/openpi}} \\
AHA-WAM & \href{https://huggingface.co/SereneC/AHA-WAM-RoboTwin2.0}{\nolinkurl{SereneC/AHA-WAM-RoboTwin2.0}} & \href{https://github.com/serene-sivy/AHA-WAM}{\nolinkurl{serene-sivy/AHA-WAM}} \\
X-VLA & \href{https://huggingface.co/2toINF/X-VLA-RoboTwin2}{\nolinkurl{2toINF/X-VLA-RoboTwin2}} & \href{https://github.com/2toINF/X-VLA}{\nolinkurl{2toINF/X-VLA}} \\
WLA & \href{https://huggingface.co/SJTU-DENG-Lab/wla_robotwin_all_image_action}{\nolinkurl{SJTU-DENG-Lab/wla_robotwin_all_image_action}} & \href{https://github.com/SJTU-DENG-Lab/WLA}{\nolinkurl{SJTU-DENG-Lab/WLA}} \\
GalaxeaVLA (G0.5) & \href{https://huggingface.co/OpenGalaxea/G05}{\nolinkurl{OpenGalaxea/G05}} & \href{https://github.com/OpenGalaxea/GalaxeaVLA}{\nolinkurl{OpenGalaxea/GalaxeaVLA}} \\
X-WAM & \href{https://huggingface.co/sharinka0715/X-WAM-checkpoints}{\nolinkurl{sharinka0715/X-WAM-checkpoints}} & \href{https://github.com/sharinka0715/X-WAM}{\nolinkurl{sharinka0715/X-WAM}} \\
Motus & \href{https://huggingface.co/motus-robotics/Motus_robotwin2}{\nolinkurl{motus-robotics/Motus_robotwin2}} & \href{https://github.com/thu-ml/Motus}{\nolinkurl{thu-ml/Motus}} \\
LingBot-VA & \href{https://huggingface.co/robbyant/lingbot-va-posttrain-robotwin}{\nolinkurl{robbyant/lingbot-va-posttrain-robotwin}} & \href{https://github.com/Robbyant/lingbot-va}{\nolinkurl{Robbyant/lingbot-va}} \\
\end{longtable}
\endgroup

FastWAM uses \nolinkurl{robotwin_uncond_3cam_384.pt}; AHA-WAM uses \nolinkurl{robotwin_ahawam.pt} with its normalization statistics; X-WAM uses \nolinkurl{robotwin_sft}; and G0.5 uses its RoboTwin release with the corresponding action tokenizer and processor.

StarVLA denotes Qwen3-VL-OFT. In Table~\ref{tab:main}, \textsuperscript{$\ast$} denotes robot/latent-action pretraining followed by RoboTwin adaptation; \textsuperscript{$\dagger$} denotes RoboTwin action-policy training from pretrained vision/language/video backbones, with no additional robot-action pretraining reported; \textsuperscript{$\ddagger$} denotes unverified checkpoint lineage. The $\ast$ group does not imply that pretraining excludes RoboTwin.

Training provenance distinguishes backbone pretraining, robot or latent-action pretraining, and RoboTwin adaptation. DM0.5, G0.5, Motus, X-WAM, and LingBot-VA report robot or latent-action pretraining; the evaluated $\pi_{0.5}$ adapts PI\textquotesingle s base checkpoint. StarVLA, FastWAM, AHA-WAM, and WLA report RoboTwin action training from pretrained backbones. Parent robot checkpoints remain unspecified for $\pi_0$ and X-VLA.

Model releases report 50 clean and 500 randomized demonstrations per RoboTwin task for $\pi_{0.5}$, StarVLA, WLA, G0.5, FastWAM, AHA-WAM, Motus, X-WAM, and LingBot-VA. Full quotas are unspecified for DM0.5, $\pi_0$, and X-VLA. X-WAM\textquotesingle s pretraining mixture includes RoboTwin~\citep{guo2026xwam}.

{\paragraph{Two checkpoints excluded from instruction-following statements.}
\emph{X-VLA} is conditioned on task names. Its released RoboTwin evaluation passes the task name with underscores replaced by spaces in place of an instruction~\citep{zheng2025xvla}. With RoboTwin's natural-language training instructions the checkpoint completes \nolinkurl{blocks_ranking_rgb} and \nolinkurl{stack_blocks_three} in 9/30 and 0/30 episodes of our evaluator (published: 79\% and 22\%); with the task-name strings it completes them in 28/30 and 16/30. Every \Benchmark{} instruction is therefore out-of-distribution text for it. The release describes the checkpoint as trained on 50 demonstrations per task. \emph{$\pi_0$} is a community release without a model card or published results. Under training instructions it completes 2 of 140 episodes on the ten RoboTwin source tasks linked to the reassignment scenes, so its \Benchmark{} scores reflect missing prerequisites (rule (v), Appendix~\ref{app:reassignment}). Both checkpoints stay in the tables, marked $^{\S}$ and $^{\P}$, and are excluded from pooled instruction-following statements.}

{\paragraph{Checkpoints with the same reported training data.}
Nine checkpoints report the same RoboTwin training data, 50 clean and 500 randomized demonstrations per task (listed above); DM0.5 and the two excluded checkpoints do not. Restricted to the nine, every reassigned instruction is followed on 39 of 900 initial states (4.3\%) and the same object is displaced under every instruction on 68\%; a distractor is displaced in 88.4\% of their 6,963 failed formal episodes on the eleven selection scenes; and all six checkpoints verified on the color training tasks belong to the nine. The differences among the nine, including the color following of LingBot-VA and $\pi_{0.5}$, arise under the same reported demonstrations.}

\subsection{Policy Interfaces and Inference Settings}
\label{app:runtime-modes}
\label{app:control-interfaces}

Policies receive three RGB views with model-dependent state and history inputs. Table~\ref{tab:control-interfaces} gives state and action interfaces. Joint vectors contain six arm joints and one gripper value per side. End-effector (EEF) vectors use $[\mathrm{xyz},\mathrm{rot6D},\mathrm{grip}]\times2$ or $[\mathrm{xyz},\mathrm{quat},\mathrm{grip}]\times2$. Model preprocessing and simulator action conversion are specified separately.

\begingroup
\small\setlength{\tabcolsep}{3pt}
\begin{longtable}{>{\raggedright\arraybackslash}p{0.25\textwidth}>{\raggedright\arraybackslash}p{0.08\textwidth}>{\raggedright\arraybackslash}p{0.26\textwidth}>{\raggedright\arraybackslash}p{0.32\textwidth}}
\caption{Control interfaces. State dimensions are per step; simulator actions are specified after adapter conversion.}\label{tab:control-interfaces}\\
\toprule
Policy & Control & Proprioceptive state & Simulator action \\
\midrule
\endfirsthead
\multicolumn{4}{l}{Table \thetable{} continued: control interfaces.}\\
\toprule
Policy & Control & Proprioceptive state & Simulator action \\
\midrule
\endhead
\midrule\multicolumn{4}{r}{Continued on next page.}\\
\endfoot
\bottomrule\endlastfoot
DM0.5, $\pi_{0.5}$, $\pi_0$, FastWAM, AHA-WAM, G0.5, Motus & Joint & 14-D joint positions and gripper values & Chunks of 14-D absolute joint/gripper targets \\
\addlinespace
StarVLA & Joint & No proprioceptive input to the model; bridge forwards RGB and instruction & Chunks of 14-D absolute joint/gripper targets \\
\addlinespace
X-VLA & EEF & 20-D: xyz + rot6D + grip, two arms & 16-D: xyz + quaternion + grip, two arms \\
\addlinespace
WLA, X-WAM & EEF & 16-D: xyz + quaternion + grip, two arms & 16-D: xyz + quaternion + grip, two arms \\
\addlinespace
LingBot-VA & EEF & Adapter transports 14-D joint state; model-side conditioning unverified & 16-D EEF targets recovered from 16-D actions relative to the initial pose \\
\end{longtable}
\endgroup

Joint targets use trajectory time parameterization; EEF targets use cuRobo planning~\citep{curobo_report23}. LingBot-VA\textquotesingle s relative EEF actions are converted to absolute targets using the initial pose.

Table~\ref{tab:runtime-modes} summarizes checkpoint and adapter configurations. Prediction and execution lengths count action targets per request. These configurations have not been independently pinned to every evaluated episode.

\begingroup
\small\setlength{\tabcolsep}{3pt}
\begin{longtable}{>{\raggedright\arraybackslash}p{0.14\textwidth}>{\raggedright\arraybackslash}p{0.18\textwidth}>{\raggedright\arraybackslash}p{0.13\textwidth}>{\raggedright\arraybackslash}p{0.46\textwidth}}
\caption{Checkpoint and adapter inference settings. Prediction/execution lengths count action targets per request. FM: flow matching. Configuration scope is defined in Appendix~\ref{app:runtime-modes}.}\label{tab:runtime-modes}\\
\toprule
Policy & Action generation & Prediction / execution & Inference path and context \\
\midrule\endfirsthead
\multicolumn{4}{l}{Table \thetable{} continued: checkpoint and inference configuration.}\\
\toprule
Policy & Action generation & Prediction / execution & Inference path and context \\
\midrule\endhead
\midrule\multicolumn{4}{r}{Continued on next page.}\\\endfoot
\bottomrule\endlastfoot
DM0.5 & Flow matching & 50 / 50 & No explicit future-image sampling. \\
\addlinespace
$\pi_{0.5}$ & Flow matching & 32 / 32 & Current observation and discretized state. \\
\addlinespace
$\pi_0$ & Flow matching & 50 / 50 & Current observation; parent checkpoint and training quota unverified. \\
\addlinespace
StarVLA & MLP / $\ell_1$ regression & 50 / 50 & RGB and instruction only at the model boundary; continuous regression from token hidden states. \\
\addlinespace
X-VLA & Flow matching & 30 / 30 & EEF state and actions; motion planning follows action decoding. \\
\addlinespace
WLA & Flow matching & 32 / 32 & Historical head image; action sampling bypasses World Expert and subtask-text generation. \\
\addlinespace
G0.5 & Flow matching & 32 / 8 & Continuous actions enabled; autoregressive actions and CoT disabled. \\
\addlinespace
FastWAM & Action denoising & 32 / 24 & Video co-training; evaluated uncond path skips future-video generation. \\
\addlinespace
AHA-WAM & Video/action FM & 16 / 16 & Video prefill every two action chunks; total configured horizon 64. \\
\addlinespace
Motus & Joint video/action FM & 16 / 16 & Inference returns predicted frames and actions; full action chunk executed. \\
\addlinespace
X-WAM & Asynchronous denoising & 32 / 32 & Action-completion early stop enabled; depth branch disabled. \\
\addlinespace
LingBot-VA & Causal video/action FM & Variable & Cross-chunk KV updates using execution observations; lengths depend on returned frame/action dimensions. \\
\end{longtable}
\endgroup

The following details complement the interface and inference tables. AHA-WAM, FastWAM, and Motus arrange the head view above the two wrist views, yielding a composite model input of $384\times320$ pixels (height $\times$ width). AHA-WAM and FastWAM resize the views before concatenation; Motus concatenates them and then resizes with padding as needed. WLA resizes images to 224 and applies JPEG encoding. The evaluated $\pi_0$ uses OpenPI JAX; $\pi_{0.5}$ uses OpenPI inference with its RoboTwin-adapted checkpoint. X-VLA uses a 0.9B RoboTwin checkpoint with Florence-2-large, and StarVLA uses a Qwen3-VL-4B backbone. X-WAM converts incremental EEF predictions into absolute targets. FastWAM's serving interface accepts the instruction, but model-internal text conditioning remains unverified; LingBot-VA's model-internal state conditioning is likewise unverified.

\subsection{Agent Observations, Tools, and Budgets}
\label{app:agents}

The website separates learned action-generating policies from agents that use an explicit language-model tool/controller interface.

Table~\ref{tab:agent-config} specifies API identifiers, reasoning settings, token caps, and request handling. {All five agents use provider-specific high reasoning settings: \nolinkurl{reasoning.effort=high} for GPT-6 Astra and GPT-6.2 Astra, \nolinkurl{thinkingLevel=HIGH} for Gemini, and adaptive thinking with high effort for Opus 5.5.} Service identifiers do not pin immutable provider weight versions.

\begin{table}[!htbp]\centering\footnotesize
\caption{Agent request settings.}\label{tab:agent-config}
\setlength{\tabcolsep}{3pt}
\begin{tabularx}{\textwidth}{@{}Xlccc@{}}\toprule
Model & API identifier & Interface & Reasoning & Output cap \\\midrule
GPT-6 Astra & \nolinkurl{gpt-6-astra} & Responses & high & 4,096 \\
{GPT-6.2 Astra} & {\nolinkurl{gpt-6.2-astra}} & {Responses} & {high} & {4,096} \\
{Opus 5.5} & {\nolinkurl{claude-opus-5-5}} & {Messages} & {adaptive, high} & {16,000} \\
{Gemini 4 Argon} & {\nolinkurl{gemini-4-argon}} & {\nolinkurl{generateContent}} & {HIGH} & {65,536} \\
Gemini 3.8 Flash & \nolinkurl{gemini-3.8-flash} & \nolinkurl{generateContent} & HIGH & 8,192 \\
\bottomrule\end{tabularx}
\par\smallskip\raggedright\noindent GPT-6 Astra uses required tools with parallel calls disabled, encrypted reasoning content with \texttt{store=false}, and an episode-specific cache key. Gemini 3.8 Flash uses function-calling mode \texttt{ANY}, provider implicit caching, and \nolinkurl{thoughtSignature}; error fallback removes the signature, then uses stateless context.
\end{table}

{All five agents} receive three RGB views, measured EEF and joint/gripper state, and contact/tracking feedback. The tools are \nolinkurl{move_eef} and \nolinkurl{finish}. The controller retains text/tool history within an episode and uses only the current camera images. Budgets are 120 model calls, a soft 3,300\,s wall limit checked between requests {(excluding model waiting time for GPT-6.2 Astra, Opus 5.5, and Gemini 4 Argon)}, and the task\textquotesingle s action-call limit.

\subsection{Agent Control and Termination}
\label{app:agent-termination}

The shared prompt specifies world-frame absolute EEF targets, scalar-first quaternions, a 0.12\,m EEF-to-gripper offset, and measured versus commanded gripper aperture.

The EEF controller limits transit, approach, and contact commands to 45\,cm/90$^\circ$, 12\,cm/30$^\circ$, and 3\,cm/10$^\circ$, respectively. Gripper changes occur at a fixed pose; recovery allows up to 3\,cm translation. Accepted commands are interpolated, with early stopping on unexpected contact or tracking error above 2.5\,cm/10$^\circ$. Rejected commands return feedback without advancing simulation.

Requests may be retried after API errors. The finish channel includes both model-requested stopping and controller-triggered stopping after repeated request failure. It ends an attempt and does not necessarily claim successful completion; task success is determined by the environment.

All 185 GPT-6 Astra and 58 Gemini 3.8 Flash successes terminate on environment success.

\subsection{Matched Evaluation Subset}
\label{app:matched5-extraction}

Each scene has 100 distinct saved seeds for the full policy evaluation. The cross-system subset uses the first five entries of each scene\textquotesingle s fixed seed list, yielding 240 episodes per system and 45/145/50 episodes at L1/L2/L3. Success and timing use the same episodes. Table~\ref{tab:matched5-complete} retains all eleven policies with recorded timing and {all five agents}{.}

Initial instruction strings match within the policy group and between {GPT-6 Astra and Gemini 3.8 Flash}. {GPT-6.2 Astra, Opus 5.5, and Gemini 4 Argon use policy wording.} {GPT-6 Astra and Gemini 3.8 Flash} differ from policy wording in 84 of 240 episodes. On \texttt{adjust\_bottle\_dustbin} and \texttt{stapler\_pad\_press}, {these two agents} receive the full compound instruction from the start, while policies receive stage-dependent instructions updated after the first stage succeeds.

\paragraph{Sensitivity to instruction-string differences.}
\label{app:matched-sensitivity}
We rescore the existing matched records without rerunning any system. For {GPT-6 Astra and Gemini 3.8 Flash}, exact initial instruction strings agree with policies in 156 episodes spanning 40 tasks. All ten episodes from the two sequence tasks belong to the 84 mismatches, so excluding both mismatches and sequence tasks still retains 156 episodes. Excluding only the sequence tasks retains 230 episodes across 46 tasks. Table~\ref{tab:matched-sensitivity} reports Pooled success with explicit denominators; these restricted samples have different task mixtures and do not replace the original Overall score. Text matching does not equalize observations, feedback, controllers, budgets, or hardware.

\begin{table}[htbp]
\centering\small
\caption{Pooled success on descriptive subsets of the existing 240 matched episodes. Identical and different refer to initial instruction strings across the policy and agent groups. Sequence denotes the two tasks with stage-dependent policy instructions. All sequence episodes are in the different-text subset. Each entry gives successes/episodes and a percentage; no interface equivalence is implied.}
\label{tab:matched-sensitivity}

\begin{tabular}{lrrrr}
\toprule
System & Identical text & Different text & Non-sequence & Sequence \\
\midrule
G0.5 & 57/156 (36.54\%) & 50/84 (59.52\%) & 98/230 (42.61\%) & 9/10 (90.00\%) \\
Motus & 46/156 (29.49\%) & 42/84 (50.00\%) & 78/230 (33.91\%) & 10/10 (100.00\%) \\
$\pi_{0.5}$ & 49/156 (31.41\%) & 31/84 (36.90\%) & 72/230 (31.30\%) & 8/10 (80.00\%) \\
StarVLA & 35/156 (22.44\%) & 32/84 (38.10\%) & 58/230 (25.22\%) & 9/10 (90.00\%) \\
FastWAM & 38/156 (24.36\%) & 25/84 (29.76\%) & 58/230 (25.22\%) & 5/10 (50.00\%) \\
AHA-WAM & 47/156 (30.13\%) & 36/84 (42.86\%) & 77/230 (33.48\%) & 6/10 (60.00\%) \\
WLA & 35/156 (22.44\%) & 26/84 (30.95\%) & 53/230 (23.04\%) & 8/10 (80.00\%) \\
X-VLA & 29/156 (18.59\%) & 31/84 (36.90\%) & 52/230 (22.61\%) & 8/10 (80.00\%) \\
$\pi_0$ & 19/156 (12.18\%) & 13/84 (15.48\%) & 27/230 (11.74\%) & 5/10 (50.00\%) \\
X-WAM & 61/156 (39.10\%) & 36/84 (42.86\%) & 91/230 (39.57\%) & 6/10 (60.00\%) \\
LingBot-VA & 62/156 (39.74\%) & 36/84 (42.86\%) & 89/230 (38.70\%) & 9/10 (90.00\%) \\
\midrule
GPT-6 & 121/156 (77.56\%) & 64/84 (76.19\%) & 180/230 (78.26\%) & 5/10 (50.00\%) \\
Gemini 3.8 Flash & 45/156 (28.85\%) & 13/84 (15.48\%) & 58/230 (25.22\%) & 0/10 (0.00\%) \\
\bottomrule
\end{tabular}
\end{table}

\subsection{Timing Definitions and Additional Summaries}
\label{app:timing}
\label{app:reported-latency}

For episode $i$, let $C_i$ be model-call count, $M_i$ total model-side time, $V_i$ total evaluation-call time, and $W_i$ episode wall time. {Table~\ref{tab:matched5-complete} reports} $\sum_iM_i/\sum_iC_i$ and episode means of $C_i$, $M_i$, $V_i$, and $W_i$, including failures and timeouts.

Policy model-side time includes service-side preprocessing; agent model-side time uses the returned API duration, with request wall time as fallback. Evaluation-call time includes inference, action execution, and rendering. Episode wall time additionally includes observation retrieval and loop overhead, excluding reset and final file writing. These boundaries define system-level timing rather than isolated model compute.

The step column retained in Table~\ref{tab:matched5-complete} reports the evaluator\textquotesingle s action counter. Agent finish and budget termination set this counter to the step limit, so agent values do not count actual executed actions.

DM0.5 is omitted from the timing comparison because its timing was not recorded. Policy evaluation uses shared GPU/CPU resources, with warm-up included. Hardware and worker placement are not matched across systems; timing characterizes the deployed evaluation pipelines.

The shared-subset comparison adds a second consideration: higher success can come at greater runtime, but more inference time does not by itself imply higher success. Under different observation, control, and instruction interfaces, GPT-6 attains 76.35\% Overall compared with 42.30\% for G0.5, with greater model-side and episode wall time in the observed deployments. Gemini takes longer than GPT-6 while completing fewer tasks than G0.5, $\pi_{0.5}$, and Motus (Table~\ref{tab:matched5}). Restricting the comparison to the 156 episodes with identical initial instruction strings gives 121/156 successes (77.56\%) for GPT-6 and 57/156 (36.54\%) for G0.5 (Appendix~\ref{app:matched-sensitivity}). This descriptive check retains the other interface and deployment differences and does not isolate an architecture effect. Taken together, the results suggest judging progress by reliable coverage of task requirements alongside aggregate success and runtime.

\begin{table}[!htbp]
\centering\footnotesize
\caption{Complete success and timing comparison on the first five seeds per scene (240 episodes per system, including failures). Overall weights families equally; Pooled weights episodes equally. Bold marks groupwise maxima for success and minima for time, including ties. $^{*}$Instruction-delivery differences: Appendix~\ref{app:matched5-extraction}. $^{\dagger}$Agent counters are not executed-step counts; timing and counter definitions: Appendix~\ref{app:timing}.}\label{tab:matched5-complete}\label{tab:matched5}
\setlength{\tabcolsep}{3pt}

\begin{tabularx}{\textwidth}{Xrrrrrrrr}
\toprule
System & Overall$\uparrow$ & Pooled$\uparrow$ & \shortstack{Calls\\(/ep.)} & \shortstack{Model$\downarrow$\\(s/call)} & \shortstack{Model$\downarrow$\\(s/ep.)} & \shortstack{Eval wall$\downarrow$\\(s/ep.)} & \shortstack{Episode\\wall (s/ep.)$\downarrow$} & \shortstack{Step\\counter}\\
\midrule
\multicolumn{9}{l}{\textit{Policies}}\\
GalaxeaVLA (G0.5) & \textbf{42.30\%} & \textbf{44.58\%} & 80.9 & 1.313 & 106.2 & 448.8 & 487.2 & 609 \\
LingBot-VA & 39.31\% & 40.83\% & 25.5 & 6.365 & 162.1 & 604.0 & 615.3 & 635 \\
X-WAM & 37.33\% & 40.42\% & 25.2 & 1.620 & 40.8 & 506.6 & 519.7 & 636 \\
Motus & 35.14\% & 36.67\% & 45.2 & 6.578 & 297.1 & 661.3 & 683.7 & 650 \\
AHA-WAM & 32.03\% & 34.58\% & 46.6 & 1.155 & 53.8 & 516.8 & 545.9 & 675 \\
$\pi_{0.5}$ & 31.92\% & 33.33\% & 26.3 & \textbf{0.226} & 6.0 & \textbf{346.2} & \textbf{359.6} & 699 \\
StarVLA & 26.37\% & 27.92\% & 18.2 & 0.285 & 5.2 & 421.3 & 433.1 & 740 \\
FastWAM & 24.23\% & 26.25\% & 34.1 & 0.928 & 31.6 & 489.4 & 510.8 & 737 \\
WLA & 24.18\% & 25.42\% & 27.0 & 0.884 & 23.9 & 971.5 & 986.3 & 749 \\
X-VLA & 24.28\% & 25.00\% & 29.6 & 0.701 & 20.7 & 958.4 & 980.4 & 782 \\
$\pi_0$ & 12.76\% & 13.33\% & 19.4 & 0.229 & \textbf{4.4} & 476.0 & 488.0 & 886 \\
\midrule
\multicolumn{9}{l}{\textit{Agents}}\\
GPT-6 Astra$^{*}$ & \textbf{76.35\%} & \textbf{77.08\%} & 24.9 & 13.248 & 329.4 & 710.3 & 730.4 & 942$^{\dagger}$ \\
{GPT-6.2 Astra} & {75.65\%} & {76.25\%} & {15.8} & {\textbf{12.429}} & {\textbf{196.2}} & {\textbf{275.7}} & {\textbf{284.5}} & {942$^{\dagger}$} \\
{Opus 5.5} & {62.76\%} & {62.08\%} & {30.2} & {21.160} & {639.3} & {821.5} & {835.3} & {942$^{\dagger}$} \\
{Gemini 4 Argon} & {29.48\%} & {29.58\%} & {15.3} & {125.558} & {1921.6} & {2816.4} & {2826.0} & {942$^{\dagger}$} \\
Gemini 3.8 Flash$^{*}$ & 24.35\% & 24.17\% & 35.2 & 46.260 & 1630.3 & 1983.9 & 2002.2 & 942$^{\dagger}$ \\
\bottomrule
\end{tabularx}

\end{table}

\FloatBarrier
\section{Global Level and Complete-Family Summaries}
\label{app:aggregate-results}

Let $s_c$ be the success fraction for task $c$, and let $\mathcal{C}_d$ contain the tasks in family $d$. We average tasks within each family, then weight the six families equally:
\begin{equation}
 S_d=\frac{1}{|\mathcal{C}_d|}\sum_{c\in\mathcal{C}_d}s_c,
 \qquad S_{\mathrm{overall}}=\frac{1}{6}\sum_{d=1}^{6}S_d.
 \label{eq:overall}
\end{equation}

Global level rates average tasks within each level, and complete-family means average tasks within each family. {Table~\ref{tab:global-levels} reports} these complementary summaries; Overall and Pooled appear in Table~\ref{tab:main}.

\begin{table}[!htbp]
\centering\footnotesize
\caption{Global level success (\%) for twelve policies, with 100 episodes per task. L1/L2/L3 average their 9/29/10 tasks equally. Bold marks column maxima, including ties. Overall and Pooled appear in Table~\ref{tab:main}; per-task intervals appear in Appendix~\ref{app:results}. Training conditions differ across checkpoints. Mean task success (\%) within each family. D1--D6 contain 10/9/9/7/6/7 equally weighted tasks; Overall equally weights the six unrounded family means. These are completion rates, not attribute-selection rates.}
\label{tab:global-levels}\label{tab:families}
\setlength{\tabcolsep}{3pt}
\resizebox{\textwidth}{!}{%
\begin{tabular}{@{}lrrr@{\hspace{10pt}}rrrrrr@{}}
\toprule
Policy & L1$\uparrow$ & L2$\uparrow$ & L3$\uparrow$ & D1$\uparrow$ & D2$\uparrow$ & D3$\uparrow$ & D4$\uparrow$ & D5$\uparrow$ & D6$\uparrow$ \\
\midrule
DM0.5 & 57.56\% & 39.69\% & 6.90\% & 49.00\% & 35.22\% & 59.89\% & 13.14\% & 33.17\% & 14.43\% \\
$\pi_{0.5}$ & 37.67\% & 35.90\% & 18.70\% & 46.40\% & 23.78\% & 47.56\% & 18.43\% & 32.00\% & 20.00\% \\
StarVLA (Qwen3-VL-OFT) & 43.78\% & 31.24\% & 11.00\% & 48.10\% & 24.44\% & 40.44\% & 13.00\% & 36.00\% & 5.43\% \\
FastWAM & 46.33\% & 26.97\% & 7.60\% & 45.50\% & 18.67\% & 50.89\% & 10.14\% & 16.33\% & 3.57\% \\
$\pi_0$ & 20.67\% & 15.10\% & 1.60\% & 24.50\% & 4.33\% & 20.33\% & 9.00\% & 18.17\% & 0.14\% \\
AHA-WAM & 55.33\% & 31.69\% & 8.40\% & 50.10\% & 22.67\% & 61.33\% & 12.71\% & 19.83\% & 5.14\% \\
X-VLA & 38.11\% & 22.97\% & 6.70\% & 32.30\% & 11.78\% & 33.89\% & 8.86\% & \textbf{45.83\%} & 0.71\% \\
WLA & 39.89\% & 29.21\% & 7.60\% & 41.10\% & 20.89\% & 41.44\% & 13.86\% & 30.50\% & 4.29\% \\
GalaxeaVLA (G0.5) & 60.67\% & \textbf{44.45\%} & 22.30\% & 57.70\% & \textbf{44.11\%} & \textbf{67.44\%} & 15.00\% & 34.83\% & \textbf{23.29\%} \\
X-WAM & \textbf{67.00\%} & 39.55\% & 18.40\% & \textbf{78.30\%} & 23.56\% & 60.11\% & 20.29\% & 34.33\% & 7.14\% \\
Motus & 55.11\% & 39.24\% & 14.20\% & 51.10\% & 39.56\% & 56.89\% & 13.43\% & 41.33\% & 7.86\% \\
LingBot-VA & 56.11\% & 39.55\% & \textbf{26.40\%} & 55.80\% & 29.11\% & 60.89\% & \textbf{25.29\%} & 39.67\% & 19.00\% \\
\bottomrule
\end{tabular}}
\end{table}

All twelve checkpoints show decreasing global success from L1 to L3. X-WAM leads L1 (67.00\%), G0.5 leads L2 (44.45\%), and LingBot-VA leads L3 (26.40\%).

\FloatBarrier
\section{Attribute Selection and Completion on Formal Tasks}
\label{app:attribute-selection}
\suppressfloats[t]

\subsection{Evaluated Systems and Event Definitions}
\label{app:attribute-events}

The study evaluates $\pi_{0.5}$, Motus, G0.5, X-VLA, StarVLA, GPT-6, and Gemini 3.8 Flash on nine tasks with five fixed seeds each: 45 episodes per system and 315 overall. Instructions, initial poses, object properties, required attributes, and effectors are matched across systems.

Force is the sum of contact-point impulse magnitudes divided by the simulation timestep. Selection is measured at the first physical step with a loaded interaction of at least 0.1\,N, using gripper--candidate interactions for seven tasks and hammer--candidate interactions for two. A unique candidate identifies the selected object; simultaneous candidates are ambiguous. These measurements sample every physical step, whereas the separate paired study uses 40\,ms telemetry (Appendix~\ref{app:language-paired}). Table~\ref{tab:attribute-coverage} reports 252 unique events, 63 no-event outcomes, and no ambiguous or unavailable measurements.

Attribute correctness compares the selected object's annotated properties with those requested by the instruction. Eligibility is fixed before observing behavior: the instruction must request the attribute, and candidate annotations must permit it to be scored. Unrequested attributes are not applicable. A selected object can receive credit for one requested attribute and not another.

For a required attribute $a$, let $z_{i,a}=1$ when the selected object's annotated value satisfies the instruction, and zero otherwise. For example, $a$ may denote size, color, or shape; a single object can have multiple attributes, each scored when required by the instruction. With $\mathcal{E}_{c,a}$ denoting the episodes of eligible scene $c$ with complete event records, the per-scene rate, defined when $|\mathcal{E}_{c,a}|>0$, is
\begin{equation}
 S_{c,a}=\frac{1}{|\mathcal{E}_{c,a}|}\sum_{i\in\mathcal{E}_{c,a}}z_{i,a}.
 \label{eq:attribute}
\end{equation}

For selection and attribute rates, complete event records form the denominator. Ambiguous and no-event outcomes receive zero credit; missing records are reported as unmeasurable rather than treated as wrong choices. Unique-event coverage is the fraction of complete records containing one qualifying selected object. Correctness conditional on a unique event uses only that subset; it is undefined if no unique event exists. Attribute rates are averaged equally over eligible tasks, each of which contributes five trials here.

\paragraph{Selection and completion.}
\label{app:selection-completion}
An initial selection error may be corrected, while a correct choice may still end in failure. We therefore report the fraction of episodes with $c_i=1$ and $y_i=0$, and completion conditional on correct selection. Here $c_i$ indicates correct first-event selection and $y_i$ indicates native task success. Completion conditional on correct selection is $\sum_i c_i y_i/\sum_i c_i$ over complete records and is undefined when $\sum_i c_i=0$. It describes a model-specific subset, not overall performance. The correct-selection/failure fraction uses all complete records as its denominator. Native task success $y_i$ is evaluated independently of first-event correctness $c_i$; their joint outcomes distinguish selection from completion without assigning a cause to later failure. Selection and property-level measurements help distinguish cases requiring closer examination of target choice from those requiring examination of what happens after a correct choice.

\begin{table}[htbp]
\centering
\small
\setlength{\tabcolsep}{3.5pt}
\caption{Selection-event coverage, target correctness, and task success across nine tasks, with 45 episodes per system. Correct selection denotes a correct first qualifying selection; Correct/fail counts such selections without completion.}
\label{tab:attribute-coverage}
\begin{tabular}{lrrrrr}
\toprule
System & Unique event & Correct selection & Native success & Correct/fail ($n$) & No event \\
\midrule
$\pi_{0.5}$ & 37/45 (82.2\%) & 25/45 (55.6\%) & 27/45 (60.0\%) & 5 & 8 \\
Motus & 39/45 (86.7\%) & 22/45 (48.9\%) & 26/45 (57.8\%) & 2 & 6 \\
G0.5 & 32/45 (71.1\%) & 20/45 (44.4\%) & 32/45 (71.1\%) & 3 & 13 \\
X-VLA & 39/45 (86.7\%) & 20/45 (44.4\%) & 17/45 (37.8\%) & 12 & 6 \\
StarVLA & 37/45 (82.2\%) & 19/45 (42.2\%) & 17/45 (37.8\%) & 8 & 8 \\
\midrule
GPT-6 & 37/45 (82.2\%) & 35/45 (77.8\%) & 31/45 (68.9\%) & 10 & 8 \\
Gemini 3.8 Flash & 31/45 (68.9\%) & 19/45 (42.2\%) & 16/45 (35.6\%) & 10 & 14 \\
\bottomrule
\end{tabular}
\end{table}

\begin{table}[htbp]
\centering\small
\caption{First-selection and completion counts across the 315 instrumented episodes (seven systems, nine tasks, five trials). The 0.1\,N event rule yields 252 unique selections and 63 no-event episodes. Native success is scored independently; correct selection followed by failure accounts for 50/315 episodes. {Added columns: the same episodes with the selection event taken as commitment (first candidate displaced by at least 5\,cm), and both events on the six pick-type tasks (210 episodes; without bell-press and the two hammer tasks, whose predicates require neither event). S/F: native success/failure.}}
\label{tab:selection-completion-counts}\label{tab:displacement-rescoring}

\setlength{\tabcolsep}{4.5pt}
\begin{tabular}{@{}lrrr@{\hspace{10pt}}rr@{\hspace{10pt}}rrrr@{}}
\toprule
 & \multicolumn{3}{c}{All nine tasks} & \multicolumn{2}{c}{{All nine tasks}} & \multicolumn{4}{c}{{Six pick-type tasks}} \\
 & \multicolumn{3}{c}{First contact (0.1\,N)} & \multicolumn{2}{c}{{Commitment}} & \multicolumn{2}{c}{{First contact}} & \multicolumn{2}{c}{{Commitment}} \\
\cmidrule(lr){2-4}\cmidrule(lr){5-6}\cmidrule(lr){7-8}\cmidrule(l){9-10}
First selection & Success & Failure & Total & {S} & {F} & {S} & {F} & {S} & {F} \\
\midrule
Correct target & 110 & 50 & 160 & {99} & {39} & {89} & {47} & {99} & {39} \\
Wrong unique target & 21 & 71 & 92 & {6} & {52} & {16} & {54} & {6} & {51} \\
No event & 35 & 28 & 63 & {61} & {58} & {0} & {4} & {0} & {15} \\
\midrule
Total & 166 & 149 & 315 & {166} & {149} & {105} & {105} & {105} & {105} \\
\bottomrule
\end{tabular}
\end{table}

\subsection{Task Coverage and Attribute Results}
\label{app:attribute-task-coverage}

\begin{figure}[htbp]
\centering
\includegraphics[width=\textwidth]{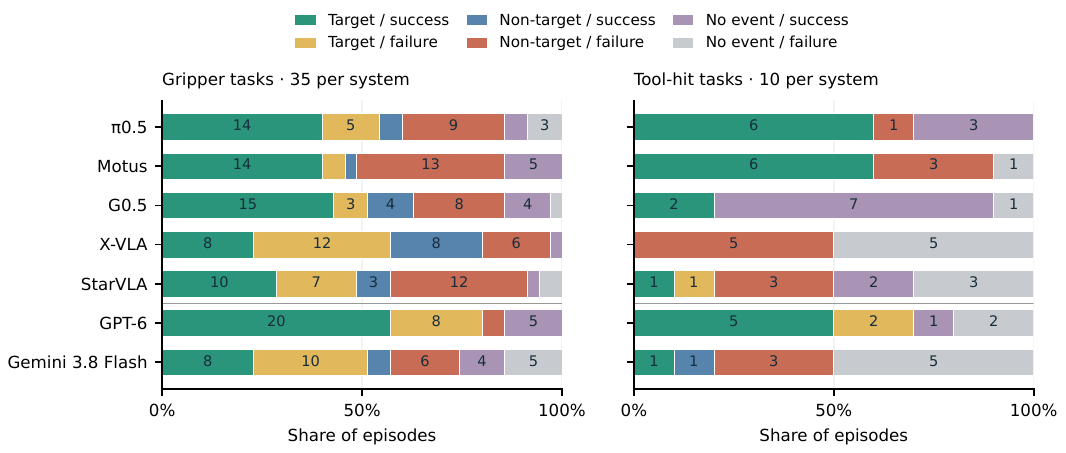}\\[-1pt]
{\small{(a)}}\\[4pt]
\includegraphics[width=\textwidth]{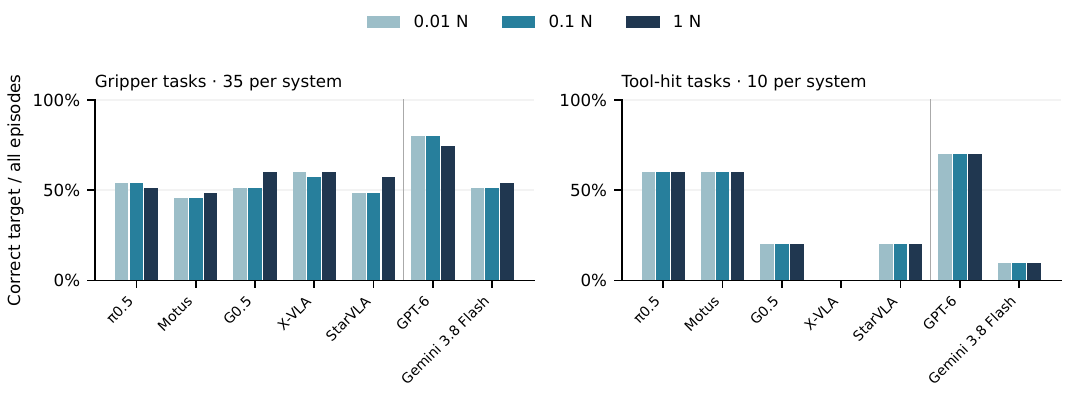}\\[-1pt]
{\small{(b)}}
\caption{{(a)} First loaded interaction ($\geq0.1$\,N) and native task success for five policies and two agents: 35 gripper and 10 tool-hit trials per system. Stacks show episode shares; labels show counts. Target/non-target refers to the first qualifying interaction; no event means none met the threshold. Success is scored independently. {(b)} First-interaction target correctness at 0.01, 0.1, and 1\,N. Rates use all 35 gripper or 10 tool-hit episodes per system, including no-event trials; the same episodes are rescored at each threshold.}
\label{fig:attributes}\label{fig:attribute-thresholds}
\end{figure}

\begin{figure}[htbp]
\centering
\includegraphics[width=\textwidth]{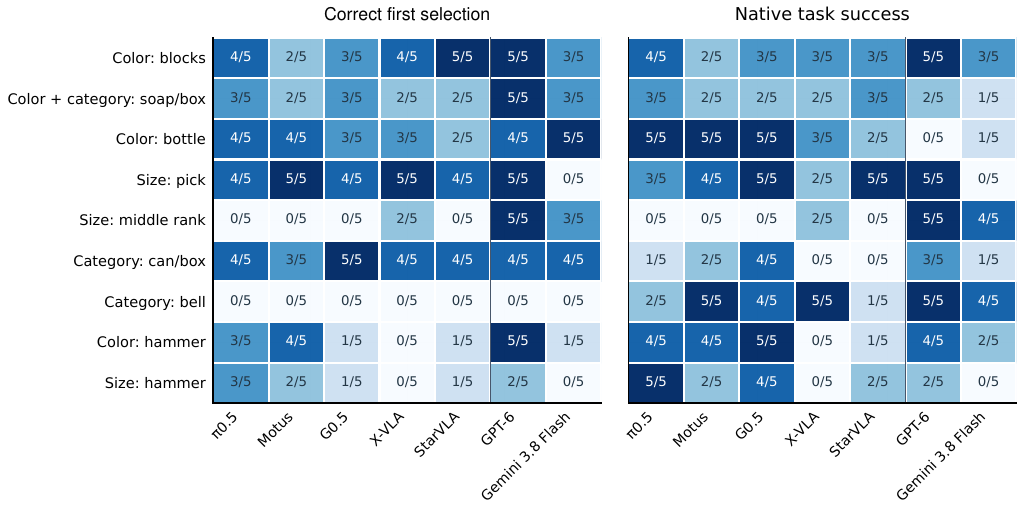}
\caption{Correct first selection at 0.1\,N (left) and native task success (right). Each cell shows the count out of five trials for one system--task pair; darker shading indicates higher rates. No-event trials remain in the denominator.}
\label{fig:attribute-task-overview}
\end{figure}

The nine task identifiers are \nolinkurl{attr_color_block_3way}, \nolinkurl{attr_bind_soap_box}, \nolinkurl{attr_bind_bottle}, \nolinkurl{attr_size_block_pick}, \nolinkurl{attr_size_rank_middle}, \nolinkurl{attr_category_can_plasticbox}, \nolinkurl{attr_category_press_bell}, \nolinkurl{attr_color_beat_block}, and \nolinkurl{attr_size_beat_smaller}. The final two use the hammer effector; the remaining seven use the gripper. Figure~\ref{fig:attribute-task-overview} reports target-event and native-success counts for every cell, and Table~\ref{tab:attribute-task-coverage} reports the corresponding unique-event coverage. {Figure~\ref{fig:attributes}(a) appeared in Section~\ref{sec:attribute-results} of the submitted version and is moved here unchanged; panel (b) is the submitted threshold figure of Appendix~\ref{app:attribute-thresholds}.}

\begin{table}[t]
\centering
\small
\setlength{\tabcolsep}{3.5pt}
\caption{Unique selection events at 0.1 N, with five episodes per cell.}
\label{tab:attribute-task-coverage}
\begin{tabular}{lrrrrrrr}
\toprule
Task & $\pi_{0.5}$ & Motus & G0.5 & X-VLA & StarVLA & GPT-6 & Gemini \\
\midrule
Color: blocks & 5/5 & 5/5 & 5/5 & 5/5 & 5/5 & 5/5 & 5/5 \\
Color + category: soap/box & 5/5 & 5/5 & 5/5 & 5/5 & 5/5 & 5/5 & 4/5 \\
Color: bottle & 5/5 & 5/5 & 5/5 & 5/5 & 5/5 & 5/5 & 5/5 \\
Size: pick & 5/5 & 5/5 & 5/5 & 5/5 & 5/5 & 5/5 & 2/5 \\
Size: middle rank & 5/5 & 5/5 & 5/5 & 5/5 & 5/5 & 5/5 & 5/5 \\
Category: can/box & 5/5 & 5/5 & 5/5 & 5/5 & 5/5 & 5/5 & 5/5 \\
Category: bell & 0/5 & 0/5 & 0/5 & 4/5 & 2/5 & 0/5 & 0/5 \\
Color: hammer & 4/5 & 5/5 & 1/5 & 3/5 & 4/5 & 5/5 & 4/5 \\
Size: hammer & 3/5 & 4/5 & 1/5 & 2/5 & 1/5 & 2/5 & 1/5 \\
\bottomrule
\end{tabular}
\end{table}

Color is evaluated on the three-block, bottle, soap/box, and color-hammer tasks; size on block-pick, middle-rank, and size-hammer tasks; and category on soap/box, can/plastic-box, and bell tasks. The soap/box task contributes to both color and category rates (Table~\ref{tab:attribute-properties}).

\begin{table}[t]
\centering
\small
\setlength{\tabcolsep}{3.5pt}
\caption{Attribute correctness and unique-event conditional rates. Color, size, and category use four, three, and three tasks respectively, with five episodes per task.}
\label{tab:attribute-properties}
\begin{tabular}{llrrr}
\toprule
System & Attribute & Correct / eligible & Unique / eligible & Correct / unique \\
\midrule
$\pi_{0.5}$ & Color & 15/20 (75.0\%) & 19/20 (95.0\%) & 15/19 (78.9\%) \\
 & Size & 7/15 (46.7\%) & 13/15 (86.7\%) & 7/13 (53.8\%) \\
 & Category & 8/15 (53.3\%) & 10/15 (66.7\%) & 8/10 (80.0\%) \\
Motus & Color & 13/20 (65.0\%) & 20/20 (100.0\%) & 13/20 (65.0\%) \\
 & Size & 7/15 (46.7\%) & 14/15 (93.3\%) & 7/14 (50.0\%) \\
 & Category & 7/15 (46.7\%) & 10/15 (66.7\%) & 7/10 (70.0\%) \\
G0.5 & Color & 10/20 (50.0\%) & 16/20 (80.0\%) & 10/16 (62.5\%) \\
 & Size & 5/15 (33.3\%) & 11/15 (73.3\%) & 5/11 (45.5\%) \\
 & Category & 10/15 (66.7\%) & 10/15 (66.7\%) & 10/10 (100.0\%) \\
X-VLA & Color & 11/20 (55.0\%) & 18/20 (90.0\%) & 11/18 (61.1\%) \\
 & Size & 7/15 (46.7\%) & 12/15 (80.0\%) & 7/12 (58.3\%) \\
 & Category & 7/15 (46.7\%) & 14/15 (93.3\%) & 7/14 (50.0\%) \\
StarVLA & Color & 11/20 (55.0\%) & 19/20 (95.0\%) & 11/19 (57.9\%) \\
 & Size & 5/15 (33.3\%) & 11/15 (73.3\%) & 5/11 (45.5\%) \\
 & Category & 8/15 (53.3\%) & 12/15 (80.0\%) & 8/12 (66.7\%) \\
\midrule
GPT-6 & Color & 19/20 (95.0\%) & 20/20 (100.0\%) & 19/20 (95.0\%) \\
 & Size & 12/15 (80.0\%) & 12/15 (80.0\%) & 12/12 (100.0\%) \\
 & Category & 9/15 (60.0\%) & 10/15 (66.7\%) & 9/10 (90.0\%) \\
Gemini 3.8 Flash & Color & 13/20 (65.0\%) & 18/20 (90.0\%) & 13/18 (72.2\%) \\
 & Size & 3/15 (20.0\%) & 8/15 (53.3\%) & 3/8 (37.5\%) \\
 & Category & 7/15 (46.7\%) & 9/15 (60.0\%) & 7/9 (77.8\%) \\
\bottomrule
\end{tabular}
\end{table}

\begin{table}[t]
\centering
\small
\setlength{\tabcolsep}{3.5pt}
\caption{Joint color--category selection on the soap/box task, with five episodes per system.}
\label{tab:attribute-binding}
\begin{tabular}{lrrrr}
\toprule
System & Both / exact target & Color only & Category only & No event \\
\midrule
$\pi_{0.5}$ & 3/5 (60.0\%) & 1/5 (20.0\%) & 1/5 (20.0\%) & 0/5 (0.0\%) \\
Motus & 2/5 (40.0\%) & 1/5 (20.0\%) & 2/5 (40.0\%) & 0/5 (0.0\%) \\
G0.5 & 3/5 (60.0\%) & 0/5 (0.0\%) & 2/5 (40.0\%) & 0/5 (0.0\%) \\
X-VLA & 2/5 (40.0\%) & 2/5 (40.0\%) & 1/5 (20.0\%) & 0/5 (0.0\%) \\
StarVLA & 2/5 (40.0\%) & 1/5 (20.0\%) & 2/5 (40.0\%) & 0/5 (0.0\%) \\
\midrule
GPT-6 & 5/5 (100.0\%) & 0/5 (0.0\%) & 0/5 (0.0\%) & 0/5 (0.0\%) \\
Gemini 3.8 Flash & 3/5 (60.0\%) & 1/5 (20.0\%) & 0/5 (0.0\%) & 1/5 (20.0\%) \\
\bottomrule
\end{tabular}
\end{table}

Table~\ref{tab:attribute-binding} shows that, on the soap/box task, 20/35 episodes select the exact target, six match color only, and eight match category only; one Gemini episode has no qualifying event. In the Motus example, ``Raise the brown coffee box'' leads to first loaded interaction with brown soap, satisfying color but not category; the task is unsuccessful.

\subsection{Threshold Sensitivity and Measurement Boundaries}
\label{app:attribute-thresholds}

{The following paragraph appeared in Section~\ref{sec:attribute-results} of the submitted version and is moved here unchanged; the main text keeps a one-sentence summary.}

This distinction is behavioral, not a direct decomposition into language and motor errors. The first event cannot identify the cause of subsequent failure, and successful completion can follow an initial mistake or occur without a loaded selection event. The bell task illustrates the latter because its native success predicate is proximity-based. We therefore retain event coverage alongside correctness and score native success independently (Table~\ref{tab:attribute-coverage}; Appendices~\ref{app:attribute-task-coverage} and \ref{app:attribute-thresholds}). Raising the event threshold from 0.1 to 1\,N changes gripper-task correct-target rates by at most 3/35 (8.6 percentage points) per system; tool-hit counts are unchanged. These measurements answer RQ2 by retaining which requested properties the initial choice satisfies, even when the task fails.

Figure~\ref{fig:attribute-thresholds}{(b)} compares the prespecified 0.01, 0.1, and 1\,N thresholds by rescoring the same trajectories. At 0.1 versus 1\,N, gripper-task correct-target counts change from 19 to 18 for $\pi_{0.5}$, 16 to 17 for Motus, 18 to 21 for G0.5, 20 to 21 for X-VLA, 17 to 20 for StarVLA, 28 to 26 for GPT-6, and 18 to 19 for Gemini, each out of 35 episodes. Tool-hit counts remain unchanged across the three thresholds.

Threshold sensitivity can arise from incidental contact: one successful G0.5 bottle trial first contacts a distractor at approximately 0.30\,N before interacting with the target above 1\,N.

Press and beat predicates can register task success through geometric proximity or contact-pair conditions without reaching the loaded-event threshold. This accounts for successful bell and hammer trials with no loaded target interaction.

\subsection{Agent Evaluation Protocol}
\label{app:attribute-provenance}

Agents use the shared RGB/state observations and EEF tools described in Appendix~\ref{app:agents}, with limits of 120 model calls, 3,300\,s, and the task\textquotesingle s action budget. GPT-6 ends on success in 31 trials, finish in eight, the step limit in five, and the call limit in one. Gemini ends on success in 16, finish in 27, and the wall limit in two. Budget termination counts as task failure.

Infrastructure-interrupted attempts are restarted with the same task and seed; completed task failures are retained. Evaluation uses one completed valid episode per system--task--seed combination.

\begin{figure}[!htbp]\centering
\includegraphics[width=\textwidth]{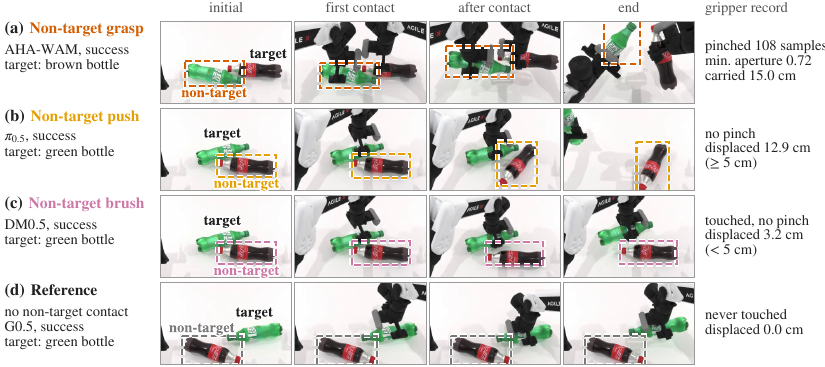}
\caption{{Non-target contacts in successful episodes, classified from the gripper state; the dashed box marks the non-target bottle. (a) Grasp: the other gripper pinches the non-target and carries it. (b) Push: no pinch; the non-target is displaced by at least 5\,cm. (c) Brush: touched and displaced by less than 5\,cm. (d) Reference: the non-target is never touched. Only in (a) is the non-target picked up.}}
\label{fig:keyframes-w3}
\end{figure}

{\subsection{Validating the Selection Event: Commitment and Contact Classes}
\label{app:displacement-rescoring}

First contact registers any loaded touch. Two additions test whether a recorded selection is a choice of object, and a third states where the event does not apply.

\paragraph{Commitment.}
The added columns of Table~\ref{tab:selection-completion-counts} rescore the 315 instrumented episodes with commitment, the first candidate displaced by at least 5\,cm, in place of first contact. Of the 21 episodes with a non-target first contact and native success, the touched object moved less than 5\,cm in 15.

\paragraph{Non-target contacts from the gripper state.}
To separate a wrong choice from an incidental touch, we reran the ten checkpoints of the pooled statements on twelve pick-type scenes (the eleven of Table~\ref{tab:failure-decomposition} and \nolinkurl{attr_bind_soap_box}) under the formal instructions, on the first 20 formal episodes of each scene, and recorded gripper state and contacts at every step (2,400 episodes). Each non-target takes its most severe class. \emph{Grasp}: both fingers of one gripper touch it, with the fingers closed by at least one tenth of the full aperture, for three or more consecutive samples, and the object moves at least 2\,cm with the gripper. \emph{Push}: no grasp, and a displacement of at least 5\,cm. \emph{Brush}: touched and displaced by less than 5\,cm. \emph{None}: not touched. An episode takes the most severe class over its non-targets; reference objects named in an instruction (the can, plate, tray, and bowl) are not counted and are never grasped. Table~\ref{tab:nontarget-contact} gives the classes per checkpoint and Figure~\ref{fig:keyframes-w3} an example of each.

Failures are mostly wrong choices: a non-target is grasped in 1,380 of the 1,655 failed episodes (83\%; 72--91\% per checkpoint). In successful episodes a non-target contact is usually a brush: of the 745 successes, 43 (5.8\%) include a non-target grasp, 17 a push, 248 a brush, and 437 no non-target contact. Counted over episodes with a non-target contact, 1,380 of 1,530 failed episodes (90.2\%) are grasps, 58 pushes, and 92 brushes; of 308 successful episodes, 248 (80.5\%) are brushes. A grasp requires every condition at once, so the grasp share is a lower bound on wrong choices; it stays between 58\% and 63\% of all episodes under alternative thresholds. The soap/box and bottle scenes use scene-specific predicates that accept a non-target grasped along with the target (Appendix~\ref{app:native-success}): the former rejects a non-target only once it has risen 10\,cm, and the latter requires only that the target's displacement exceed the distractor's. In the formal evaluation of the bottle scene, 119 of the 574 successful episodes also displace the distractor by at least 5\,cm; without them its success rate is 37.9\% instead of 47.8\%.}

\begin{table}[!htbp]
\centering\small\color{black}
\setlength{\tabcolsep}{5pt}
\caption{{Non-target contacts by class, from gripper-state records: ten checkpoints on twelve pick-type scenes, the first 20 formal episodes of each (240 episodes per checkpoint). Class shares are in \% of all episodes; an episode takes its most severe class. The last two columns give the episodes with a non-target grasp among failed and among successful episodes.}}
\label{tab:nontarget-contact}
\begin{tabular}{@{}lrrrrrrr@{}}
\toprule
 & & \multicolumn{4}{c}{Class (\%)} & \multicolumn{2}{c}{Non-target grasp} \\
\cmidrule(lr){3-6}\cmidrule(l){7-8}
Checkpoint & Successes & Grasp & Push & Brush & None & In failures & In successes \\
\midrule
DM0.5 & 70 & 61 & 2 & 16 & 21 & 142/170 (84\%) & 4/70 \\
$\pi_{0.5}$ & 65 & 60 & 2 & 16 & 21 & 138/175 (79\%) & 6/65 \\
StarVLA & 76 & 52 & 7 & 18 & 23 & 120/164 (73\%) & 5/76 \\
FastWAM & 62 & 54 & 5 & 14 & 27 & 129/178 (72\%) & 0/62 \\
AHA-WAM & 73 & 62 & 5 & 12 & 20 & 142/167 (85\%) & 8/73 \\
WLA & 64 & 67 & 4 & 14 & 15 & 156/176 (89\%) & 4/64 \\
G0.5 & 77 & 58 & 2 & 17 & 22 & 138/163 (85\%) & 2/77 \\
X-WAM & 87 & 59 & 1 & 15 & 25 & 136/153 (89\%) & 5/87 \\
Motus & 71 & 66 & 0 & 11 & 22 & 152/169 (90\%) & 6/71 \\
LingBot-VA & 100 & 54 & 2 & 8 & 36 & 127/140 (91\%) & 3/100 \\
\midrule
All ten & 745 & 59 & 3 & 14 & 23 & 1380/1655 (83\%) & 43/745 \\
\bottomrule
\end{tabular}
\end{table}

\begin{figure}[!htbp]\centering
\includegraphics[width=\textwidth]{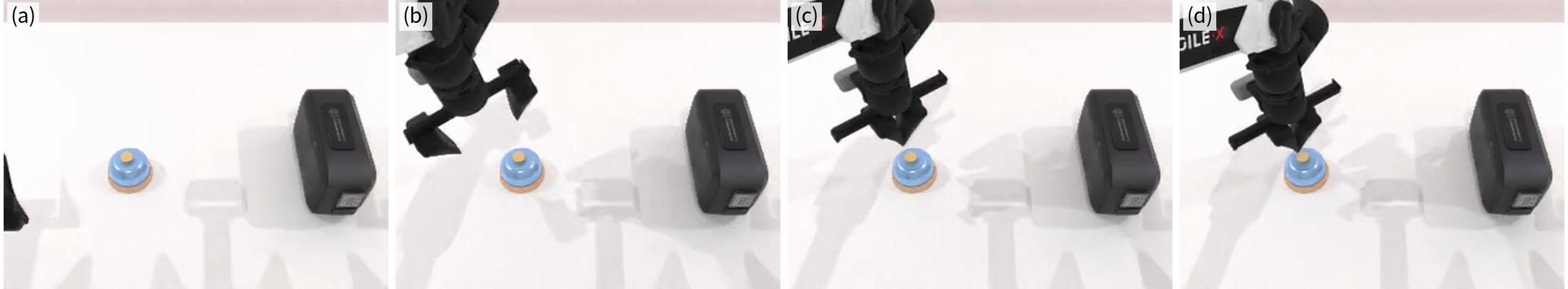}
\caption{{A successful bell-press episode without a selection event (Motus, seed 100001; ``Please press the bell.''). (a) Initial state: bell on the left, alarm clock on the right. (b) The left arm approaches the bell. (c) The gripper is above the bell. (d) The gripper presses the bell and the native predicate reports success. The peak contact force recorded on the bell and on the alarm clock is 0\,N, so the episode has no first-contact event; it is counted as \emph{no event}, not as a wrong selection.}}
\label{fig:keyframes-bell}
\end{figure}

{\paragraph{Tasks without a selection event.}
The bell-press and hammer tasks are completed without displacing a candidate (no commitment event in 104 of their 105 episodes), so commitment is read on the six pick-type tasks: 47 episodes fail after a target first contact and 39 after a target commitment. No loaded event is recorded on the bell in any of the 35 bell-press episodes, 26 of which succeed (Figure~\ref{fig:keyframes-bell}). Without this task, which contributes 0/35, pooled category correctness at first contact is 56/70 (80.0\%) instead of 56/105 (53.3\%); color without the color-hammer task is 77/105 (73.3\%, from 65.7\%), and size 37/70 (52.9\%, from 43.8\%).}

{\subsection{Failure Decomposition of the Formal Evaluation}
\label{app:failure-decomposition}

\paragraph{Scene sets used by the analyses.}
\emph{Pick-type} denotes a scene or task that is completed by grasping and moving one of several candidates. The instrumented attribute study uses nine attribute tasks, six of them pick-type (Appendix~\ref{app:attribute-events}). Target reassignment uses ten pick-type selection scenes: three attribute-selection scenes and the seven referring-expression scenes (Appendix~\ref{app:reassignment}). The failure decomposition uses these ten and the bottle scene (eleven), and the gripper-state reruns add the soap/box scene (twelve; Appendix~\ref{app:displacement-rescoring}).

Table~\ref{tab:failure-decomposition} uses the per-episode displacement record of the formal evaluation on the eleven pick-type selection scenes (four attribute-selection, seven referring-expression): the maximum displacement of the target and of the most displaced distractor. A failed episode is assigned by which of the two exceeded 5\,cm; the record neither orders the two displacements nor separates a grasp from a push. Table~\ref{tab:failure-decomposition-policy} gives the decomposition per checkpoint. Without X-VLA and $\pi_0$ (Appendix~\ref{app:checkpoints}), the ten remaining checkpoints have 7,757 failed episodes: 37.2\% displace only a distractor, 51.1\% both, 2.7\% only the target, and 9.0\% neither (a distractor in 88.3\%).}

\begin{table}[t]
\centering\footnotesize\color{black}
\setlength{\tabcolsep}{4pt}
\caption{{Failure decomposition of the formal evaluation on eleven pick-type selection scenes (twelve checkpoints, 1,200 episodes per scene): share of failed episodes in which only a distractor, both a distractor and the target, only the target, or neither was displaced by at least 5\,cm.}}
\label{tab:failure-decomposition}
\begin{tabular}{@{}llrrrrrr@{}}
\toprule
 & & & & \multicolumn{4}{c}{Share of failed episodes (\%)} \\
\cmidrule(l){5-8}
Family & Scene & SR (\%) & Failures & Distractor only & Both & Target only & Neither \\
\midrule
D3 & \nolinkurl{attr_color_block_3way} & 70.2 & 358 & 41.6 & 42.2 & 9.8 & 6.4 \\
D3 & \nolinkurl{attr_size_block_pick} & 77.0 & 276 & 13.8 & 18.5 & 23.6 & 44.2 \\
D3 & \nolinkurl{attr_size_rank_middle} & 5.1 & 1139 & 46.3 & 43.3 & 1.4 & 9.0 \\
D3 & \nolinkurl{attr_bind_bottle} & 47.8 & 626 & 17.1 & 81.8 & 0.6 & 0.5 \\
D4 & \nolinkurl{ref_block_on_plate} & 1.2 & 1185 & 71.8 & 22.2 & 0.6 & 5.4 \\
D4 & \nolinkurl{ref_block_behind_can} & 9.9 & 1081 & 25.9 & 4.6 & 9.3 & 60.1 \\
D4 & \nolinkurl{ref_cup_on_tray} & 2.8 & 1167 & 64.7 & 25.1 & 2.8 & 7.4 \\
D4 & \nolinkurl{ref_leftmost_of_four} & 33.7 & 796 & 21.0 & 71.5 & 1.0 & 6.5 \\
D4 & \nolinkurl{ref_mug_next_to_bowl} & 30.2 & 837 & 15.2 & 81.5 & 1.1 & 2.3 \\
D4 & \nolinkurl{ref_not_the_red_one} & 10.1 & 1079 & 12.7 & 85.1 & 0.2 & 2.0 \\
D4 & \nolinkurl{ref_second_from_left} & 13.1 & 1043 & 52.3 & 42.4 & 1.0 & 4.3 \\
\midrule
\multicolumn{2}{@{}l}{All eleven scenes} & 27.4 & 9587 & 38.4 & 46.1 & 3.0 & 12.4 \\
\bottomrule
\end{tabular}
\end{table}

\section{Paired Behavioral and Model-Access Diagnostics}
\label{app:language-diagnostics}
\suppressfloats[t]

The controlled study crosses target instructions and object layouts on the three-color block task for $\pi_{0.5}$ and Motus, with future-video and internal-intervention analyses for Motus. For a policy that also predicts video, we then ask whether the predicted target follows the request, and whether altering future representations changes the relative target-distance preference of predicted actions.

\begin{figure}[htbp]
\centering
\includegraphics[width=\textwidth]{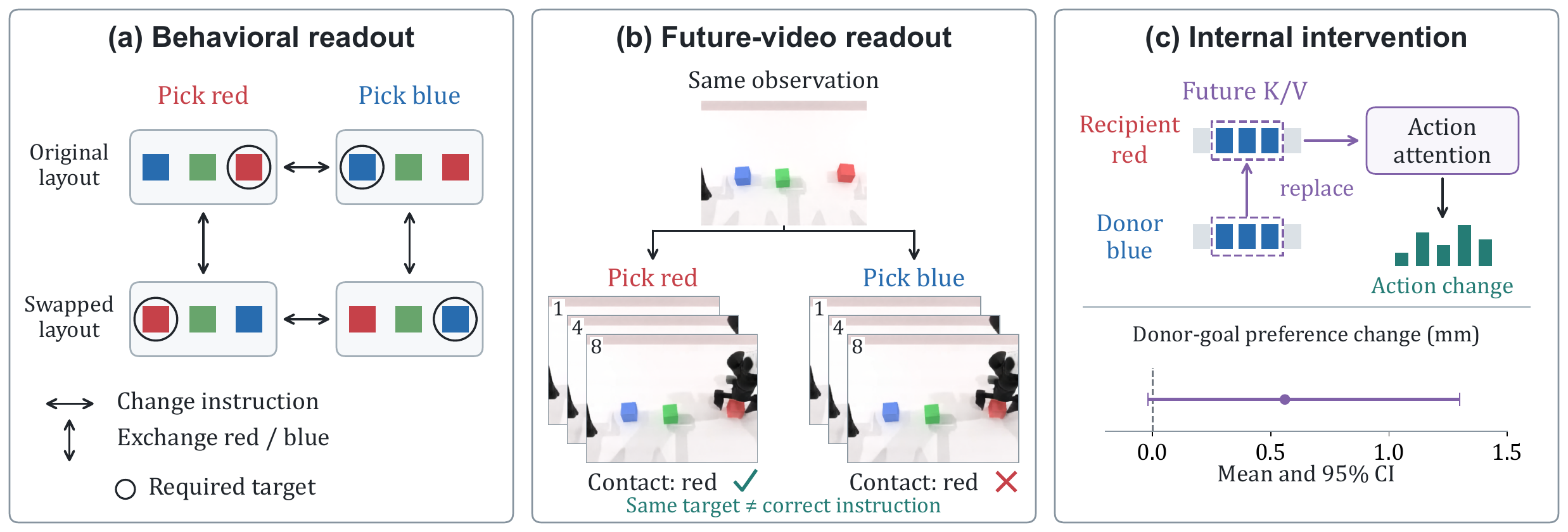}
\caption{Three diagnostic layers. Left: paired instructions and layouts, with rings marking requested targets. Center: frames 1, 4, and 8 (back to front) from each Motus forecast; final frames and later contacts indicate red under both requests. Right: alternate-instruction future key/value (K/V) representations replace the corresponding entries at the action-attention interface, with other inputs fixed. The plot shows mean donor-goal preference change and its 95\% configuration-bootstrap interval, measured from commanded kinematics; blocks and action bars are schematic. Motus results come from {Appendix~\ref{app:language-internal}}.}
\label{fig:diagnostic-protocols}
\end{figure}

\subsection{Paired Conditions and Observable Events}
\label{app:language-paired}

{The following four paragraphs appeared in Sections~\ref{sec:diagnostic-method} and~\ref{sec:language-results} of the submitted version and are moved here unchanged apart from cross-references; the main text keeps a summary.}

In the paired behavioral study, each initial scene configuration has two valid target instructions and two layouts, with the red and blue objects physically exchanged. The paired and formal-task studies use different event definitions and denominators ({below and} Appendix~\ref{app:attribute-events}).

Instruction-paired correctness requires both target instructions to be followed in the same view; always selecting one object cannot pass. Layout-paired correctness requires the requested color to be selected across both layouts. These are paired correctness rates, not merely measures of action change. A correct trial has a unique first recorded loaded gripper contact with the requested object; no qualifying contact or an ambiguous contact receives zero credit. Uncertainty estimation keeps all four conditions of an initial configuration together {(below)}.

The controlled study crosses 20 initial configurations of one three-color block task with two instructions and two layouts for each checkpoint, yielding 160 trajectories. The 95\% intervals use 20,000 bootstrap resamples of the 20 configurations {(below)}.

\begin{figure}[!htbp]
 \centering
 \includegraphics[width=\textwidth]{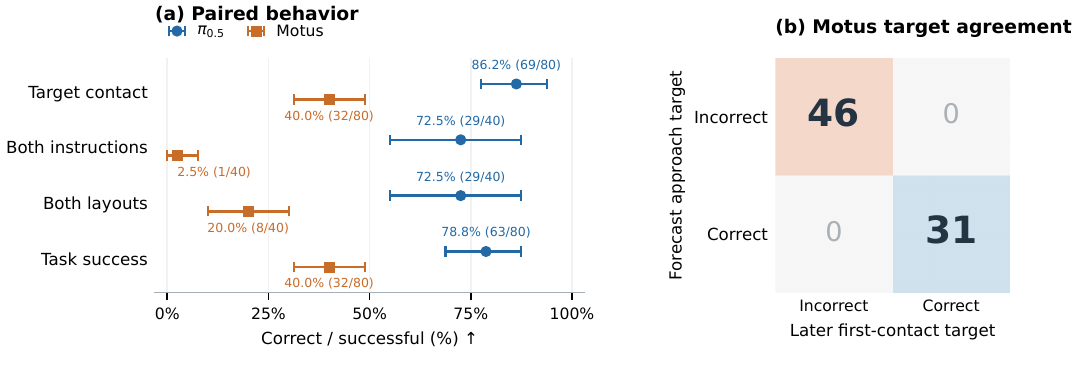}
\caption{Checkpoint-specific color-selection diagnostics under a separate 256-native-action protocol. (a) Contact and success rates use 80 trajectories per checkpoint; paired correctness uses 40 pairs. Error bars show 95\% bootstrap intervals over 20 initial configurations. (b) Motus forecast and later first-contact targets are jointly wrong in 46 cases and jointly correct in 31; three forecasts are unresolved.}
 \label{fig:language-diagnostics}
\end{figure}

The paired behavioral results separate isolated success from reliable responses to alternative requests. Motus succeeds in 40.0\% of individual trials, yet selects the requested target under both instructions in only 1/40 same-layout pairs; $\pi_{0.5}$ does so in 29/40 pairs (Figure~\ref{fig:language-diagnostics}). Motus contacts the same color under both instructions in 38/40 pairs, compared with 9/40 for $\pi_{0.5}$; Table~\ref{tab:language-color-layout} separates target colors and layouts. Layout-paired results are reported in Table~\ref{tab:language-paired}. The practical implication is to test alternative valid requests in the same scene: repeating successful behavior is insufficient if the requested target can change.

Twenty initial configurations of one three-color block task each define four conditions: original or red--blue-exchanged object poses, crossed with ``Pick up the red block.'' or ``Pick up the blue block.'' The green block and initial robot state remain fixed. The evaluator\textquotesingle s target matches the instruction and is hidden from the policy. Both policies receive three RGB views, the instruction, and 14-dimensional commanded joint state. Initial configurations are included independently of policy outcome. Figure~\ref{fig:diagnostic-protocols} illustrates the paired design and an example forecast contrast.

Action chunks contain 32 commands for $\pi_{0.5}$ and 16 for Motus. Runs terminate at task success or 256 native action commands, not 256 model-inference decisions. Inference noise uses seed $17+d$ at decision $d$, matching the first decision across conditions.

Loaded contact is the summed gripper--candidate contact-impulse magnitude divided by the 4\,ms physical timestep, thresholded at 0.1\,N. The first qualifying event in 40\,ms telemetry determines the selected object. Ambiguous and no-contact cases receive zero correctness credit. Target-correct counts remain unchanged at 0.01 and 1\,N.

Native success checks lift clearance and target selection. An auxiliary approach event requires tool center point (TCP) distance below 0.12\,m, a 0.02\,m margin over alternatives, 0.03\,m approach from initialization, and 0.20\,s persistence. Stable holding requires both fingers of one hand in loaded contact, 0.02\,m object rise, and 0.20\,s persistence.

In the paired behavioral study, each initial configuration $s$ has two valid target instructions $t\in\{r,b\}$ and two layouts $\ell\in\{0,1\}$, with the red and blue objects physically exchanged. Let $c_{s,\ell,t}$ indicate that the first recorded loaded gripper contact is uniquely with the requested object. No qualifying contact or an ambiguous contact receives zero. Across $S$ initial configurations, alongside contact accuracy and task success, we report
\begin{equation}
 P_{\mathrm{instr}}=\frac{1}{2S}\sum_{s,\ell}c_{s,\ell,r}c_{s,\ell,b},\qquad
 P_{\mathrm{layout}}=\frac{1}{2S}\sum_{s,t}c_{s,0,t}c_{s,1,t}.
 \label{eq:paired-diagnostics}
\end{equation}

\begin{table}[t]
\centering\small
\caption{Controlled color-task results: rates and 95\% configuration-bootstrap intervals in percent, with counts over 80 trajectories or 40 pairs per checkpoint. Bold marks the higher rate, including ties. These trials are separate from the formal benchmark.}
\label{tab:language-paired}
\begin{tabularx}{\textwidth}{@{}Xcc@{}}
\toprule
Measurement $\uparrow$ & $\pi_{0.5}$ & Motus \\
\midrule
Correct target contact & \textbf{86.2\%} [77.5, 93.8] (69/80) & 40.0\% [31.2, 48.8] (32/80) \\
Both instructions correct & \textbf{72.5\%} [55.0, 87.5] (29/40) & 2.5\% [0.0, 7.5] (1/40) \\
Both layouts correct & \textbf{72.5\%} [55.0, 87.5] (29/40) & 20.0\% [10.0, 30.0] (8/40) \\
Task success & \textbf{78.8\%} [68.8, 87.5] (63/80) & 40.0\% [31.2, 48.8] (32/80) \\
\bottomrule
\end{tabularx}
\end{table}

{Figure~\ref{fig:language-diagnostics} appeared in Section~\ref{sec:language-results} of the submitted version and is moved here unchanged.}

Table~\ref{tab:language-paired} reports the paired and per-episode results. Layout-paired correctness requires the requested color to be selected after the objects exchange positions. Paired rates average the two instruction or layout pairs within each initial configuration, then average configurations. Confidence intervals use 20,000 bootstrap samples of configurations, retaining all four conditions, and the 2.5th and 97.5th percentiles. Conditional completion is 63/69 for $\pi_{0.5}$ and 32/32 for Motus.

\begin{table}[htbp]
\centering\small
\caption{Checkpoint-specific color-selection outcomes by layout and requested color, with 20 trials per row. Red/blue/green columns count first-contact colors, followed by target-correct and native-success counts. Swapped exchanges the red and blue poses while retaining the green pose.}
\label{tab:language-color-layout}

\begin{tabular}{lllrrrrr}
\toprule
Checkpoint & Layout & Request & Red & Blue & Green & Correct & Success \\
\midrule
$\pi_{0.5}$ & Original & red & 18 & 0 & 2 & 18/20 & 17/20 \\
$\pi_{0.5}$ & Original & blue & 3 & 17 & 0 & 17/20 & 15/20 \\
$\pi_{0.5}$ & Swapped & red & 19 & 1 & 0 & 19/20 & 17/20 \\
$\pi_{0.5}$ & Swapped & blue & 5 & 15 & 0 & 15/20 & 14/20 \\
\midrule
Motus & Original & red & 13 & 1 & 6 & 13/20 & 13/20 \\
Motus & Original & blue & 13 & 1 & 6 & 1/20 & 1/20 \\
Motus & Swapped & red & 12 & 5 & 3 & 12/20 & 12/20 \\
Motus & Swapped & blue & 10 & 6 & 4 & 6/20 & 6/20 \\
\bottomrule
\end{tabular}
\end{table}

Table~\ref{tab:language-color-layout} shows a descriptive red preference for Motus: 48/80 contacts are red, 13 blue, and 19 green. The same contacted color persists under both instructions in 38/40 pairs, versus 9/40 for $\pi_{0.5}$. Across all four conditions within an initial configuration, Motus is correct in four/three/two/one/zero conditions in 0/1/13/3/3 configurations; the corresponding counts for $\pi_{0.5}$ are 12/5/3/0/0. These patterns do not establish a universal color preference or a causal position bias.

\subsection{Output Sensitivity and Future-Video Readout}
\label{app:language-future}

{The following paragraph appeared in Section~\ref{sec:diagnostic-method} of the submitted version and is moved here unchanged apart from cross-references; the main text keeps a summary.}

For systems that expose predicted images, we compare the visible approach target with the instruction and with the later contact target, retaining unresolved predictions. The readout compares target identity across horizons, not frame-aligned contact accuracy. It separates instruction correctness from forecast--action agreement: a forecast and execution can agree on the wrong object. Agreement alone does not establish that the forecast causally directs action. {This section} specifies the labeling procedure and its limits.

{The following paragraph appeared in Section~\ref{sec:language-results} of the submitted version and is moved here unchanged apart from cross-references; the main text keeps a one-sentence summary.}

The future-video readout asks whether the same mismatch is already visible in the predicted outcome. In the Motus case, all 77 readable first-decision forecasts (out of 80) agree with the later contact target beyond the forecast horizon, including 46 agreements on the wrong object (Figure~\ref{fig:language-diagnostics}). Prediction and execution can therefore be mutually consistent while jointly violating the request. Labels use a task-specific image rule with visual review, without independent blinded annotation. This agreement does not establish that the forecast causes the later choice.

\begin{figure}[t]
 \centering
 \includegraphics[width=0.66\textwidth]{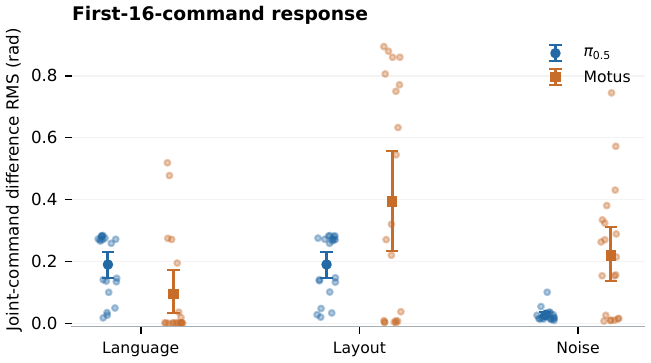}
\caption{First-decision action sensitivity to instruction, layout, and noise changes, measured by joint-command difference RMS over the first 16 commands. Faint points show 20 configuration means; solid markers and error bars show overall means and 95\% configuration-bootstrap intervals. Internal interventions are reported in Table~\ref{tab:language-internal}.}
 \label{fig:language-sensitivity}
\end{figure}

Figure~\ref{fig:language-sensitivity} summarizes first-decision action sensitivity to instruction, layout, and inference-noise changes. To measure first-decision action sensitivity, we change the target instruction while holding the observation, state, and inference noise fixed, or exchange object poses while holding the instruction and inference noise fixed. The noise control changes inference seed 17 to 29. RMS differences between paired joint-command predictions use the first 16 commands and exclude grippers. Language/layout/noise configuration means are 0.191/0.191/0.028\,rad for $\pi_{0.5}$ and 0.096/0.394/0.220\,rad for Motus.

Motus predicts eight video frames and 16 action commands jointly. The future readout uses the last predicted head-camera frame. A task-specific RGB rule identifies the gripper and three block colors, requiring a nearest candidate within 12 pixels and a margin of at least 6 pixels. It is defined using the first initial configuration. Visual review checks all 80 conditions, with disagreements left unresolved; labels lack independent blinded human validation.

The readout identifies 77/80 targets, of which 31 match the instruction; both instructions are correct in 1/40 pairs. All 77 readable targets match later first contact, including 46 wrong selections. First contacts occur after the first action chunk, beyond the first-decision forecast horizon. Three unresolved forecasts remain in the instruction-correctness denominator. This makes forecast-target correctness a necessary companion to agreement when diagnosing these cases: agreement alone would leave the wrong goal unchallenged.

An auxiliary ``Please pick up\ldots'' wording control yields action RMS of 0.005\,rad for $\pi_{0.5}$ and 0.023\,rad for Motus on original-layout inputs. Automated Motus readouts yield 1/40 correct instruction pairs under changed noise and 0/20 under the wording change. These auxiliary labels use the image rule alone. Language-induced final-frame RGB mean absolute error (MAE) is 0.015 on $[0,1]$ intensities.

\begin{figure}[!htbp]
\centering
\includegraphics[width=\textwidth]{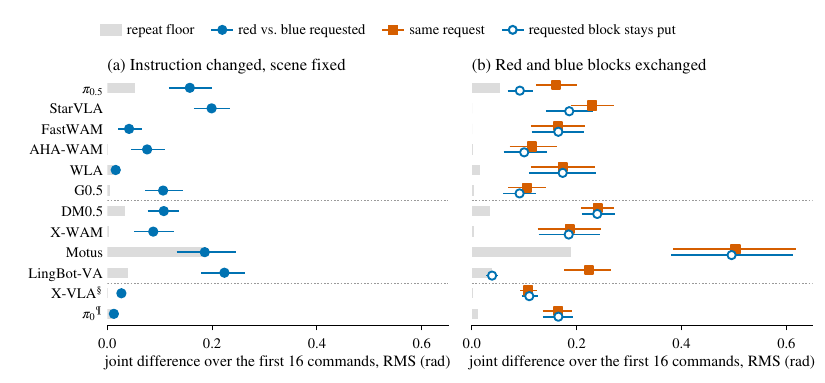}
\caption{{First-decision sensitivity of twelve checkpoints on the three-color scene: RMS joint difference over the first 16 commands between paired episodes from one initial state, with 95\% bootstrap intervals over 20 initial states. Gray bars: two repetitions of one condition. (a) Red against blue requested in one layout. (b) Red and blue blocks exchanged under one request, and the pairing in which the requested block stays in place. Row markers and groups as in Figure~\ref{fig:probe-training}.}}
\label{fig:first-decision}
\end{figure}

{\paragraph{First-decision sensitivity for all checkpoints.}
We repeated the measurement for the twelve policy checkpoints on the 20 frozen initial states of the three-color scene used by target reassignment (Figure~\ref{fig:first-decision}). The conditions cross the instructions ``Pick up the red block.'' and ``Pick up the blue block.'' with the original layout and a layout in which the red and blue blocks exchange poses, with three repetitions each (240 episodes per checkpoint). An episode runs until at least 16 commands are issued. Distances are RMS differences over the first 16 commands and the twelve arm joints between paired episodes from one initial state; checkpoints that command end-effector poses are compared on the reached joint states. Inference noise is not fixed: two repetitions of one condition give the repeat baseline. A further pairing keeps the requested block in place (red requested in the original layout against blue requested in the exchanged layout); a checkpoint that heads for the named color then falls to the repeat baseline.

The color word changes the first motion by a factor of about three or more over the repeat baseline in nine of the twelve checkpoints. For Motus, WLA, and $\pi_0$ the change stays at the baseline (1.0, 1.0, and 1.1 times): the instruction does not reach the first motion. A change is not yet a move toward the named block. When the requested block stays in place, only LingBot-VA falls to its baseline (0.039 against 0.039\,rad), and $\pi_{0.5}$ falls partway (0.092\,rad, between its baseline of 0.053 and its layout difference of 0.160). For the other checkpoints this pairing stays close to the layout difference: the first motion follows the layout, and the color word adds a smaller change that is unrelated to where the named block is. LingBot-VA and $\pi_{0.5}$ are also the two checkpoints that follow color under target reassignment on this scene (10/10 and 6/10 states; Table~\ref{tab:reassignment-scene}); StarVLA, which follows on 3/10 states there, does not show it in its first motion. For Motus the pattern of the submitted measurement recurs: 0.186\,rad for language, 0.503 for layout, and 0.189 for the repeat baseline.}

\subsection{Internal Interventions: Design, Controls, and Results}
\label{app:language-internal}

{The following paragraph appeared in Section~\ref{sec:diagnostic-method} of the submitted version and is moved here unchanged apart from cross-references; the main text keeps a summary.}

Where representations can be accessed, we replace selected representations with those from an alternative-instruction run while holding recipient inputs and sampling noise fixed. We measure action change and donor-goal preference change: the change from baseline in the difference between commanded distances to the recipient and donor targets. Positive values favor the donor relatively, without necessarily reducing donor-target distance. Closed-loop benefit requires executing the intervened policy and is not tested here. {This section} specifies the readouts and controls. These probes concern the tested computational connection, not a global ranking of language and vision.

{The following paragraph appeared in Section~\ref{sec:language-results} of the submitted version and is moved here unchanged apart from cross-references; the main text keeps a summary.}

Internal intervention tests whether future-representation replacement changes predicted actions and donor-goal preference. Replacement produces a mean root-mean-square (RMS) joint-command difference of 0.030\,rad and a mean donor-goal preference change of 0.560\,mm (95\% interval: $[-0.019,1.299]$\,mm). Changing the full instruction produces a 0.096\,rad RMS difference and a 2.529\,mm preference change. The actions are sensitive to replacement, yet useful goal transfer is not established. Only two readable contrasts predict different target objects, both from one initial configuration, and no intervened closed-loop improvement is tested {(below)}. The result supports evaluating the direction of an intervention's effect, rather than treating any action change as evidence of improved goal following; it does not establish that future representations have no goal influence.

For each of 80 Motus conditions, the donor uses the alternative instruction with observation, state, and noise fixed. At all 30 joint-attention layers and ten denoising steps, K/V entries for 120 tokens in the first future temporal group are replaced where action queries read them; only action-expert output rows are updated. Later computation can propagate the intervention. Token groups may contain mixed information after attention, so intervention effects are specific to the replaced representations~\citep{zhang2024patching}.

Controls comprise original-value writeback, the full donor instruction, current-video and understanding replacements, half-strength future replacement, and norm-matched or scrambled perturbations. K and V norms are matched separately at each layer and denoising step. Sham replacements reproduce baseline predictions. The internal analysis is exploratory, including a swapped-layout extension selected after initial inspection.

Forward kinematics maps each predicted joint command $q_k$ to the two TCP positions $p_h(q_k)$, $h\in\{L,R\}$. For target $g$ at its initial position $x_g$, define $D_g(q_k)=\min_h\|p_h(q_k)-x_g\|_2$. Each target independently uses the nearer arm. For intervention $m$ and baseline $0$, donor-goal preference change is
\[
\Delta_{\mathrm{pref}}^{(m)}=\frac{1}{16}\sum_{k=1}^{16}
\Bigl[\bigl(D_{\mathrm{rec}}(q_k^{(m)})-D_{\mathrm{don}}(q_k^{(m)})\bigr)
-\bigl(D_{\mathrm{rec}}(q_k^{(0)})-D_{\mathrm{don}}(q_k^{(0)})\bigr)\Bigr].
\]
Here rec and don denote the recipient and donor instruction targets. Distances are converted to millimeters for reporting. Positive values indicate a relative preference change toward the donor; they do not necessarily mean that donor-target distance decreases, and do not measure executed displacement or success.

\begin{table}[t]
\centering\small
\caption{Motus interventions on 80 first-decision conditions from 20 initial configurations: means and 95\% configuration-bootstrap intervals. Donor-goal preference change measures relative target distances under commanded kinematics, not executed displacement or success. Matched denotes nominal key/value (K/V) perturbation-norm matching.}
\label{tab:language-internal}
\begin{tabularx}{\textwidth}{@{}Xrr@{}}
\toprule
Intervention & \shortstack{Joint-command\\difference RMS (rad)} & \shortstack{Donor-goal preference\\change (mm)} \\
\midrule
Original-value writeback & 0.000 [0.000, 0.000] & 0.000 [0.000, 0.000] \\
Full instruction change & 0.096 [0.033, 0.173] & 2.529 [0.734, 5.339] \\
Future replacement & 0.030 [0.013, 0.051] & 0.560 [-0.019, 1.299] \\
Future, half strength & 0.010 [0.005, 0.015] & 0.041 [-0.190, 0.257] \\
Current replacement & 0.007 [0.003, 0.012] & 0.157 [0.017, 0.347] \\
Understanding replacement & 0.006 [0.003, 0.011] & 0.058 [-0.366, 0.430] \\
Future, matched & 0.030 [0.013, 0.051] & 0.544 [-0.043, 1.292] \\
Current, matched & 0.015 [0.006, 0.025] & 0.357 [0.090, 0.748] \\
Understanding, matched & 0.080 [0.030, 0.142] & 1.330 [-1.560, 4.050] \\
Future, scrambled + matched & 0.003 [0.002, 0.005] & 0.048 [-0.041, 0.162] \\
\bottomrule
\end{tabularx}
\end{table}

Table~\ref{tab:language-internal} reports the intervention results. Near-future replacement produces a mean joint-command RMS difference of 0.030\,rad and a donor-goal preference change of 0.560\,mm (95\% interval: $[-0.019,1.299]$\,mm). Donor forecasts match their instruction in 31/80 conditions. Two recipient--donor contrasts from one initial configuration predict different readable targets; their relative distance-preference change toward the donor forecast target is $-4.432$\,mm.

After BF16 arithmetic, 100.00\%, 96.32\%, 100.00\%, and 99.98\% of applied perturbation norms fall within 10\% of the reference for matched future, current, understanding, and scrambled future replacements, respectively.

\subsection{Diagnostic Interpretation}
\label{app:diagnostic-interpretation}

The combined readouts suggest concrete next checks---whether target choices track alternative requests, whether forecasts already select the wrong target, and whether representation changes favor the intended target in predicted actions.

Joint failure to switch forecasts and actions motivates examination of conditioning and target binding. Correct forecasts with unresponsive actions motivate testing the connection to action generation, while correct selection followed by failure motivates execution analysis. These patterns guide subsequent experiments.

\section{Complete Task Results}
\label{app:results}

Tables~\ref{tab:all-scenes}, \ref{tab:all-scenes-b}, and \ref{tab:remaining-policy-scenes} report per-task policy success and 95\% Wilson intervals; Table~\ref{tab:gpt6-scenes} gives {all five agents}.

\subsection{Policy Task Results}
\label{app:policy-results}

\begingroup
\footnotesize\setlength{\tabcolsep}{3pt}
\begin{longtable}{lcccc}
\caption{Per-task policy success (\%), part 1 of 3: 100 trials per task, with 95\% Wilson intervals in parentheses. Bold marks taskwise maxima across all twelve policies in the three parts, including ties. Intervals describe per-task success, not Overall; task families and execution limits appear in Appendices~\ref{app:inventory} and~\ref{app:implementation}. {Printed with one line per task in the revision; values, in \%, are unchanged.}}\label{tab:all-scenes}\\
\toprule
Scene & DM0.5$\uparrow$ & $\pi_{0.5}$$\uparrow$ & StarVLA$\uparrow$ & FastWAM$\uparrow$ \\
\midrule\endfirsthead
\multicolumn{5}{l}{Table \thetable{} continued: task success (95\% Wilson interval).}\\
\toprule
Scene & DM0.5$\uparrow$ & $\pi_{0.5}$$\uparrow$ & StarVLA$\uparrow$ & FastWAM$\uparrow$ \\
\midrule\endhead
\midrule\multicolumn{5}{r}{Continued on next page.}\\\endfoot
\bottomrule\endlastfoot
{\scriptsize\nolinkurl{grab_dumbbell}} & 76 {\scriptsize(66.8, 83.3)} & 56 {\scriptsize(46.2, 65.3)} & 42 {\scriptsize(32.8, 51.8)} & 52 {\scriptsize(42.3, 61.5)} \\
{\scriptsize\nolinkurl{move_can_coaster}} & 0 {\scriptsize(0.0, 3.7)} & 7 {\scriptsize(3.4, 13.7)} & 45 {\scriptsize(35.6, 54.8)} & 31 {\scriptsize(22.8, 40.6)} \\
{\scriptsize\nolinkurl{move_cup_pad}} & \textbf{99} {\scriptsize(94.6, 99.8)} & 68 {\scriptsize(58.3, 76.3)} & 93 {\scriptsize(86.3, 96.6)} & 67 {\scriptsize(57.3, 75.4)} \\
{\scriptsize\nolinkurl{place_teabox_scale}} & 56 {\scriptsize(46.2, 65.3)} & 21 {\scriptsize(14.2, 30.0)} & 36 {\scriptsize(27.3, 45.8)} & 43 {\scriptsize(33.7, 52.8)} \\
{\scriptsize\nolinkurl{grasp_mallet_novel}} & 90 {\scriptsize(82.6, 94.5)} & 97 {\scriptsize(91.5, 99.0)} & 94 {\scriptsize(87.5, 97.2)} & 94 {\scriptsize(87.5, 97.2)} \\
{\scriptsize\nolinkurl{lift_trophy}} & 54 {\scriptsize(44.3, 63.4)} & 66 {\scriptsize(56.3, 74.5)} & 74 {\scriptsize(64.6, 81.6)} & 49 {\scriptsize(39.4, 58.7)} \\
{\scriptsize\nolinkurl{place_seal_coaster}} & 3 {\scriptsize(1.0, 8.5)} & 34 {\scriptsize(25.5, 43.7)} & 42 {\scriptsize(32.8, 51.8)} & 28 {\scriptsize(20.1, 37.5)} \\
{\scriptsize\nolinkurl{place_teabox_skillet}} & 0 {\scriptsize(0.0, 3.7)} & 5 {\scriptsize(2.2, 11.2)} & 13 {\scriptsize(7.8, 21.0)} & 0 {\scriptsize(0.0, 3.7)} \\
{\scriptsize\nolinkurl{place_teabox_stand}} & 57 {\scriptsize(47.2, 66.3)} & 39 {\scriptsize(30.0, 48.8)} & 1 {\scriptsize(0.2, 5.4)} & 45 {\scriptsize(35.6, 54.8)} \\
{\scriptsize\nolinkurl{place_toycar_plasticbox}} & 55 {\scriptsize(45.2, 64.4)} & 71 {\scriptsize(61.5, 79.0)} & 41 {\scriptsize(31.9, 50.8)} & 46 {\scriptsize(36.6, 55.7)} \\
{\scriptsize\nolinkurl{place_a2b_left_novel}} & \textbf{75} {\scriptsize(65.7, 82.5)} & 46 {\scriptsize(36.6, 55.7)} & 59 {\scriptsize(49.2, 68.1)} & 46 {\scriptsize(36.6, 55.7)} \\
{\scriptsize\nolinkurl{place_a2b_right_novel}} & \textbf{65} {\scriptsize(55.3, 73.6)} & 41 {\scriptsize(31.9, 50.8)} & 59 {\scriptsize(49.2, 68.1)} & 38 {\scriptsize(29.1, 47.8)} \\
{\scriptsize\nolinkurl{card_right_of_scanner}} & 23 {\scriptsize(15.8, 32.2)} & 12 {\scriptsize(7.0, 19.8)} & 14 {\scriptsize(8.5, 22.1)} & 1 {\scriptsize(0.2, 5.4)} \\
{\scriptsize\nolinkurl{phone_right_of_laptop}} & 38 {\scriptsize(29.1, 47.8)} & 2 {\scriptsize(0.6, 7.0)} & 0 {\scriptsize(0.0, 3.7)} & 4 {\scriptsize(1.6, 9.8)} \\
{\scriptsize\nolinkurl{place_can_right_of_plasticbox}} & 58 {\scriptsize(48.2, 67.2)} & 29 {\scriptsize(21.0, 38.5)} & 9 {\scriptsize(4.8, 16.2)} & 37 {\scriptsize(28.2, 46.8)} \\
{\scriptsize\nolinkurl{place_cube_right_of_plate}} & 42 {\scriptsize(32.8, 51.8)} & 36 {\scriptsize(27.3, 45.8)} & 18 {\scriptsize(11.7, 26.7)} & 24 {\scriptsize(16.7, 33.2)} \\
{\scriptsize\nolinkurl{place_teabox_front_of_skillet}} & 0 {\scriptsize(0.0, 3.7)} & 4 {\scriptsize(1.6, 9.8)} & \textbf{23} {\scriptsize(15.8, 32.2)} & 3 {\scriptsize(1.0, 8.5)} \\
{\scriptsize\nolinkurl{put_toycar_left_basket}} & 0 {\scriptsize(0.0, 3.7)} & 2 {\scriptsize(0.6, 7.0)} & 2 {\scriptsize(0.6, 7.0)} & 1 {\scriptsize(0.2, 5.4)} \\
{\scriptsize\nolinkurl{place_between_bowls}} & 16 {\scriptsize(10.1, 24.4)} & 42 {\scriptsize(32.8, 51.8)} & 36 {\scriptsize(27.3, 45.8)} & 14 {\scriptsize(8.5, 22.1)} \\
{\scriptsize\nolinkurl{attr_bind_bottle}} & 47 {\scriptsize(37.5, 56.7)} & 46 {\scriptsize(36.6, 55.7)} & 33 {\scriptsize(24.6, 42.7)} & 40 {\scriptsize(30.9, 49.8)} \\
{\scriptsize\nolinkurl{attr_category_press_bell}} & \textbf{100} {\scriptsize(96.3, 100.0)} & 54 {\scriptsize(44.3, 63.4)} & 18 {\scriptsize(11.7, 26.7)} & \textbf{100} {\scriptsize(96.3, 100.0)} \\
{\scriptsize\nolinkurl{attr_category_can_plasticbox}} & 19 {\scriptsize(12.5, 27.8)} & 4 {\scriptsize(1.6, 9.8)} & 2 {\scriptsize(0.6, 7.0)} & 0 {\scriptsize(0.0, 3.7)} \\
{\scriptsize\nolinkurl{attr_color_beat_block}} & 87 {\scriptsize(79.0, 92.2)} & 69 {\scriptsize(59.4, 77.2)} & 28 {\scriptsize(20.1, 37.5)} & 71 {\scriptsize(61.5, 79.0)} \\
{\scriptsize\nolinkurl{attr_color_block_3way}} & 69 {\scriptsize(59.4, 77.2)} & 83 {\scriptsize(74.5, 89.1)} & 85 {\scriptsize(76.7, 90.7)} & 60 {\scriptsize(50.2, 69.1)} \\
{\scriptsize\nolinkurl{attr_size_beat_smaller}} & \textbf{86} {\scriptsize(77.9, 91.5)} & 58 {\scriptsize(48.2, 67.2)} & 59 {\scriptsize(49.2, 68.1)} & 40 {\scriptsize(30.9, 49.8)} \\
{\scriptsize\nolinkurl{attr_size_block_pick}} & 97 {\scriptsize(91.5, 99.0)} & 66 {\scriptsize(56.3, 74.5)} & 79 {\scriptsize(70.0, 85.8)} & 86 {\scriptsize(77.9, 91.5)} \\
{\scriptsize\nolinkurl{attr_bind_soap_box}} & 33 {\scriptsize(24.6, 42.7)} & 45 {\scriptsize(35.6, 54.8)} & 56 {\scriptsize(46.2, 65.3)} & 61 {\scriptsize(51.2, 70.0)} \\
{\scriptsize\nolinkurl{attr_size_rank_middle}} & 1 {\scriptsize(0.2, 5.4)} & 3 {\scriptsize(1.0, 8.5)} & 4 {\scriptsize(1.6, 9.8)} & 0 {\scriptsize(0.0, 3.7)} \\
{\scriptsize\nolinkurl{ref_block_on_plate}} & 0 {\scriptsize(0.0, 3.7)} & 0 {\scriptsize(0.0, 3.7)} & \textbf{9} {\scriptsize(4.8, 16.2)} & 0 {\scriptsize(0.0, 3.7)} \\
{\scriptsize\nolinkurl{ref_block_behind_can}} & 2 {\scriptsize(0.6, 7.0)} & 24 {\scriptsize(16.7, 33.2)} & 5 {\scriptsize(2.2, 11.2)} & 0 {\scriptsize(0.0, 3.7)} \\
{\scriptsize\nolinkurl{ref_cup_on_tray}} & 0 {\scriptsize(0.0, 3.7)} & 0 {\scriptsize(0.0, 3.7)} & \textbf{22} {\scriptsize(15.0, 31.1)} & 0 {\scriptsize(0.0, 3.7)} \\
{\scriptsize\nolinkurl{ref_leftmost_of_four}} & 43 {\scriptsize(33.7, 52.8)} & 52 {\scriptsize(42.3, 61.5)} & 23 {\scriptsize(15.8, 32.2)} & 27 {\scriptsize(19.3, 36.4)} \\
{\scriptsize\nolinkurl{ref_mug_next_to_bowl}} & 32 {\scriptsize(23.7, 41.7)} & 45 {\scriptsize(35.6, 54.8)} & 6 {\scriptsize(2.8, 12.5)} & 17 {\scriptsize(10.9, 25.5)} \\
{\scriptsize\nolinkurl{ref_not_the_red_one}} & 0 {\scriptsize(0.0, 3.7)} & 1 {\scriptsize(0.2, 5.4)} & 16 {\scriptsize(10.1, 24.4)} & 16 {\scriptsize(10.1, 24.4)} \\
{\scriptsize\nolinkurl{ref_second_from_left}} & 15 {\scriptsize(9.3, 23.3)} & 7 {\scriptsize(3.4, 13.7)} & 10 {\scriptsize(5.5, 17.4)} & 11 {\scriptsize(6.3, 18.6)} \\
{\scriptsize\nolinkurl{adjust_bottle_dustbin}} & 94 {\scriptsize(87.5, 97.2)} & 97 {\scriptsize(91.5, 99.0)} & \textbf{100} {\scriptsize(96.3, 100.0)} & 1 {\scriptsize(0.2, 5.4)} \\
{\scriptsize\nolinkurl{click_bell_press_stapler}} & 7 {\scriptsize(3.4, 13.7)} & 0 {\scriptsize(0.0, 3.7)} & 17 {\scriptsize(10.9, 25.5)} & 3 {\scriptsize(1.0, 8.5)} \\
{\scriptsize\nolinkurl{serve_teabox_then_ring}} & 0 {\scriptsize(0.0, 3.7)} & \textbf{1} {\scriptsize(0.2, 5.4)} & 0 {\scriptsize(0.0, 3.7)} & 0 {\scriptsize(0.0, 3.7)} \\
{\scriptsize\nolinkurl{stapler_pad_press}} & 97 {\scriptsize(91.5, 99.0)} & 59 {\scriptsize(49.2, 68.1)} & 99 {\scriptsize(94.6, 99.8)} & 94 {\scriptsize(87.5, 97.2)} \\
{\scriptsize\nolinkurl{lift_pot_to_pad}} & 0 {\scriptsize(0.0, 3.7)} & 11 {\scriptsize(6.3, 18.6)} & 0 {\scriptsize(0.0, 3.7)} & 0 {\scriptsize(0.0, 3.7)} \\
{\scriptsize\nolinkurl{pack_toycar_then_ring}} & 1 {\scriptsize(0.2, 5.4)} & 24 {\scriptsize(16.7, 33.2)} & 0 {\scriptsize(0.0, 3.7)} & 0 {\scriptsize(0.0, 3.7)} \\
{\scriptsize\nolinkurl{move_cup_pad_mix}} & \textbf{83} {\scriptsize(74.5, 89.1)} & 78 {\scriptsize(68.9, 85.0)} & 8 {\scriptsize(4.1, 15.0)} & 24 {\scriptsize(16.7, 33.2)} \\
{\scriptsize\nolinkurl{ref_block_on_plate_mix}} & 0 {\scriptsize(0.0, 3.7)} & 0 {\scriptsize(0.0, 3.7)} & \textbf{16} {\scriptsize(10.1, 24.4)} & 0 {\scriptsize(0.0, 3.7)} \\
{\scriptsize\nolinkurl{attr_teabox_basket_then_ring}} & 0 {\scriptsize(0.0, 3.7)} & 0 {\scriptsize(0.0, 3.7)} & \textbf{2} {\scriptsize(0.6, 7.0)} & 0 {\scriptsize(0.0, 3.7)} \\
{\scriptsize\nolinkurl{card_right_of_scanner_mix}} & 11 {\scriptsize(6.3, 18.6)} & 8 {\scriptsize(4.1, 15.0)} & 7 {\scriptsize(3.4, 13.7)} & 1 {\scriptsize(0.2, 5.4)} \\
{\scriptsize\nolinkurl{place_teabox_skillet_mix}} & 0 {\scriptsize(0.0, 3.7)} & 0 {\scriptsize(0.0, 3.7)} & 3 {\scriptsize(1.0, 8.5)} & 0 {\scriptsize(0.0, 3.7)} \\
{\scriptsize\nolinkurl{put_toycar_left_basket_mix}} & 2 {\scriptsize(0.6, 7.0)} & 4 {\scriptsize(1.6, 9.8)} & 2 {\scriptsize(0.6, 7.0)} & 0 {\scriptsize(0.0, 3.7)} \\
{\scriptsize\nolinkurl{roller_to_plate_mix}} & 5 {\scriptsize(2.2, 11.2)} & \textbf{50} {\scriptsize(40.4, 59.6)} & 0 {\scriptsize(0.0, 3.7)} & 0 {\scriptsize(0.0, 3.7)} \\
\end{longtable}\endgroup

\begingroup
\footnotesize\setlength{\tabcolsep}{2pt}
\begin{longtable}{lccccc}
\caption{Per-task policy success (\%), part 2 of 3, using the 100-trial protocol and 95\% Wilson intervals of Table~\ref{tab:all-scenes}. Bold marks taskwise maxima across all twelve policies, including ties. {Printed with one line per task in the revision; values, in \%, are unchanged.}}\label{tab:all-scenes-b}\\
\toprule
Scene & $\pi_0$$\uparrow$ & AHA-WAM$\uparrow$ & X-VLA$\uparrow$ & WLA$\uparrow$ & G0.5$\uparrow$ \\
\midrule\endfirsthead
\multicolumn{6}{l}{Table \thetable{} continued: task success (95\% Wilson interval).}\\
\toprule
Scene & $\pi_0$$\uparrow$ & AHA-WAM$\uparrow$ & X-VLA$\uparrow$ & WLA$\uparrow$ & G0.5$\uparrow$ \\
\midrule\endhead
\midrule\multicolumn{6}{r}{Continued on next page.}\\\endfoot
\bottomrule\endlastfoot
{\scriptsize\nolinkurl{grab_dumbbell}} & 52 {\scriptsize(42.3, 61.5)} & 80 {\scriptsize(71.1, 86.7)} & 81 {\scriptsize(72.2, 87.5)} & 47 {\scriptsize(37.5, 56.7)} & 90 {\scriptsize(82.6, 94.5)} \\
{\scriptsize\nolinkurl{move_can_coaster}} & 0 {\scriptsize(0.0, 3.7)} & 33 {\scriptsize(24.6, 42.7)} & 18 {\scriptsize(11.7, 26.7)} & 6 {\scriptsize(2.8, 12.5)} & 45 {\scriptsize(35.6, 54.8)} \\
{\scriptsize\nolinkurl{move_cup_pad}} & 40 {\scriptsize(30.9, 49.8)} & 76 {\scriptsize(66.8, 83.3)} & 72 {\scriptsize(62.5, 79.9)} & 77 {\scriptsize(67.8, 84.2)} & 66 {\scriptsize(56.3, 74.5)} \\
{\scriptsize\nolinkurl{place_teabox_scale}} & 6 {\scriptsize(2.8, 12.5)} & 52 {\scriptsize(42.3, 61.5)} & 0 {\scriptsize(0.0, 3.7)} & 41 {\scriptsize(31.9, 50.8)} & 67 {\scriptsize(57.3, 75.4)} \\
{\scriptsize\nolinkurl{grasp_mallet_novel}} & 94 {\scriptsize(87.5, 97.2)} & 93 {\scriptsize(86.3, 96.6)} & 93 {\scriptsize(86.3, 96.6)} & 95 {\scriptsize(88.8, 97.8)} & 79 {\scriptsize(70.0, 85.8)} \\
{\scriptsize\nolinkurl{lift_trophy}} & 44 {\scriptsize(34.7, 53.8)} & 29 {\scriptsize(21.0, 38.5)} & 25 {\scriptsize(17.5, 34.3)} & 76 {\scriptsize(66.8, 83.3)} & 30 {\scriptsize(21.9, 39.6)} \\
{\scriptsize\nolinkurl{place_seal_coaster}} & 1 {\scriptsize(0.2, 5.4)} & 49 {\scriptsize(39.4, 58.7)} & 11 {\scriptsize(6.3, 18.6)} & 16 {\scriptsize(10.1, 24.4)} & 47 {\scriptsize(37.5, 56.7)} \\
{\scriptsize\nolinkurl{place_teabox_skillet}} & 0 {\scriptsize(0.0, 3.7)} & 0 {\scriptsize(0.0, 3.7)} & 2 {\scriptsize(0.6, 7.0)} & 0 {\scriptsize(0.0, 3.7)} & 26 {\scriptsize(18.4, 35.4)} \\
{\scriptsize\nolinkurl{place_teabox_stand}} & 0 {\scriptsize(0.0, 3.7)} & 47 {\scriptsize(37.5, 56.7)} & 2 {\scriptsize(0.6, 7.0)} & 22 {\scriptsize(15.0, 31.1)} & 67 {\scriptsize(57.3, 75.4)} \\
{\scriptsize\nolinkurl{place_toycar_plasticbox}} & 8 {\scriptsize(4.1, 15.0)} & 42 {\scriptsize(32.8, 51.8)} & 19 {\scriptsize(12.5, 27.8)} & 31 {\scriptsize(22.8, 40.6)} & 60 {\scriptsize(50.2, 69.1)} \\
{\scriptsize\nolinkurl{place_a2b_left_novel}} & 5 {\scriptsize(2.2, 11.2)} & 61 {\scriptsize(51.2, 70.0)} & 19 {\scriptsize(12.5, 27.8)} & 46 {\scriptsize(36.6, 55.7)} & 67 {\scriptsize(57.3, 75.4)} \\
{\scriptsize\nolinkurl{place_a2b_right_novel}} & 5 {\scriptsize(2.2, 11.2)} & 52 {\scriptsize(42.3, 61.5)} & 20 {\scriptsize(13.3, 28.9)} & 47 {\scriptsize(37.5, 56.7)} & 60 {\scriptsize(50.2, 69.1)} \\
{\scriptsize\nolinkurl{card_right_of_scanner}} & 2 {\scriptsize(0.6, 7.0)} & 1 {\scriptsize(0.2, 5.4)} & 10 {\scriptsize(5.5, 17.4)} & 5 {\scriptsize(2.2, 11.2)} & \textbf{56} {\scriptsize(46.2, 65.3)} \\
{\scriptsize\nolinkurl{phone_right_of_laptop}} & 21 {\scriptsize(14.2, 30.0)} & 16 {\scriptsize(10.1, 24.4)} & 8 {\scriptsize(4.1, 15.0)} & 0 {\scriptsize(0.0, 3.7)} & \textbf{47} {\scriptsize(37.5, 56.7)} \\
{\scriptsize\nolinkurl{place_can_right_of_plasticbox}} & 3 {\scriptsize(1.0, 8.5)} & 28 {\scriptsize(20.1, 37.5)} & 16 {\scriptsize(10.1, 24.4)} & 32 {\scriptsize(23.7, 41.7)} & 78 {\scriptsize(68.9, 85.0)} \\
{\scriptsize\nolinkurl{place_cube_right_of_plate}} & 0 {\scriptsize(0.0, 3.7)} & 25 {\scriptsize(17.5, 34.3)} & 11 {\scriptsize(6.3, 18.6)} & 14 {\scriptsize(8.5, 22.1)} & 40 {\scriptsize(30.9, 49.8)} \\
{\scriptsize\nolinkurl{place_teabox_front_of_skillet}} & 3 {\scriptsize(1.0, 8.5)} & 3 {\scriptsize(1.0, 8.5)} & 11 {\scriptsize(6.3, 18.6)} & 4 {\scriptsize(1.6, 9.8)} & 2 {\scriptsize(0.6, 7.0)} \\
{\scriptsize\nolinkurl{put_toycar_left_basket}} & 0 {\scriptsize(0.0, 3.7)} & 1 {\scriptsize(0.2, 5.4)} & 0 {\scriptsize(0.0, 3.7)} & 1 {\scriptsize(0.2, 5.4)} & 28 {\scriptsize(20.1, 37.5)} \\
{\scriptsize\nolinkurl{place_between_bowls}} & 0 {\scriptsize(0.0, 3.7)} & 17 {\scriptsize(10.9, 25.5)} & 11 {\scriptsize(6.3, 18.6)} & 39 {\scriptsize(30.0, 48.8)} & 19 {\scriptsize(12.5, 27.8)} \\
{\scriptsize\nolinkurl{attr_bind_bottle}} & 26 {\scriptsize(18.4, 35.4)} & 44 {\scriptsize(34.7, 53.8)} & 39 {\scriptsize(30.0, 48.8)} & 28 {\scriptsize(20.1, 37.5)} & \textbf{74} {\scriptsize(64.6, 81.6)} \\
{\scriptsize\nolinkurl{attr_category_press_bell}} & 52 {\scriptsize(42.3, 61.5)} & \textbf{100} {\scriptsize(96.3, 100.0)} & 94 {\scriptsize(87.5, 97.2)} & 67 {\scriptsize(57.3, 75.4)} & 77 {\scriptsize(67.8, 84.2)} \\
{\scriptsize\nolinkurl{attr_category_can_plasticbox}} & 0 {\scriptsize(0.0, 3.7)} & 23 {\scriptsize(15.8, 32.2)} & 0 {\scriptsize(0.0, 3.7)} & 2 {\scriptsize(0.6, 7.0)} & \textbf{44} {\scriptsize(34.7, 53.8)} \\
{\scriptsize\nolinkurl{attr_color_beat_block}} & 6 {\scriptsize(2.8, 12.5)} & 85 {\scriptsize(76.7, 90.7)} & 5 {\scriptsize(2.2, 11.2)} & 22 {\scriptsize(15.0, 31.1)} & \textbf{91} {\scriptsize(83.8, 95.2)} \\
{\scriptsize\nolinkurl{attr_color_block_3way}} & 50 {\scriptsize(40.4, 59.6)} & 74 {\scriptsize(64.6, 81.6)} & 58 {\scriptsize(48.2, 67.2)} & 59 {\scriptsize(49.2, 68.1)} & 75 {\scriptsize(65.7, 82.5)} \\
{\scriptsize\nolinkurl{attr_size_beat_smaller}} & 28 {\scriptsize(20.1, 37.5)} & 71 {\scriptsize(61.5, 79.0)} & 11 {\scriptsize(6.3, 18.6)} & 68 {\scriptsize(58.3, 76.3)} & 84 {\scriptsize(75.6, 89.9)} \\
{\scriptsize\nolinkurl{attr_size_block_pick}} & 5 {\scriptsize(2.2, 11.2)} & 95 {\scriptsize(88.8, 97.8)} & 55 {\scriptsize(45.2, 64.4)} & 90 {\scriptsize(82.6, 94.5)} & 99 {\scriptsize(94.6, 99.8)} \\
{\scriptsize\nolinkurl{attr_bind_soap_box}} & 13 {\scriptsize(7.8, 21.0)} & 54 {\scriptsize(44.3, 63.4)} & 34 {\scriptsize(25.5, 43.7)} & 37 {\scriptsize(28.2, 46.8)} & 63 {\scriptsize(53.2, 71.8)} \\
{\scriptsize\nolinkurl{attr_size_rank_middle}} & 3 {\scriptsize(1.0, 8.5)} & 6 {\scriptsize(2.8, 12.5)} & 9 {\scriptsize(4.8, 16.2)} & 0 {\scriptsize(0.0, 3.7)} & 0 {\scriptsize(0.0, 3.7)} \\
{\scriptsize\nolinkurl{ref_block_on_plate}} & 0 {\scriptsize(0.0, 3.7)} & 0 {\scriptsize(0.0, 3.7)} & 0 {\scriptsize(0.0, 3.7)} & 0 {\scriptsize(0.0, 3.7)} & 0 {\scriptsize(0.0, 3.7)} \\
{\scriptsize\nolinkurl{ref_block_behind_can}} & \textbf{28} {\scriptsize(20.1, 37.5)} & 0 {\scriptsize(0.0, 3.7)} & 0 {\scriptsize(0.0, 3.7)} & 3 {\scriptsize(1.0, 8.5)} & 21 {\scriptsize(14.2, 30.0)} \\
{\scriptsize\nolinkurl{ref_cup_on_tray}} & 0 {\scriptsize(0.0, 3.7)} & 3 {\scriptsize(1.0, 8.5)} & 0 {\scriptsize(0.0, 3.7)} & 0 {\scriptsize(0.0, 3.7)} & 0 {\scriptsize(0.0, 3.7)} \\
{\scriptsize\nolinkurl{ref_leftmost_of_four}} & 13 {\scriptsize(7.8, 21.0)} & 28 {\scriptsize(20.1, 37.5)} & 4 {\scriptsize(1.6, 9.8)} & 32 {\scriptsize(23.7, 41.7)} & \textbf{70} {\scriptsize(60.4, 78.1)} \\
{\scriptsize\nolinkurl{ref_mug_next_to_bowl}} & 12 {\scriptsize(7.0, 19.8)} & 27 {\scriptsize(19.3, 36.4)} & 40 {\scriptsize(30.9, 49.8)} & 45 {\scriptsize(35.6, 54.8)} & 5 {\scriptsize(2.2, 11.2)} \\
{\scriptsize\nolinkurl{ref_not_the_red_one}} & 7 {\scriptsize(3.4, 13.7)} & 14 {\scriptsize(8.5, 22.1)} & 2 {\scriptsize(0.6, 7.0)} & 15 {\scriptsize(9.3, 23.3)} & 7 {\scriptsize(3.4, 13.7)} \\
{\scriptsize\nolinkurl{ref_second_from_left}} & 3 {\scriptsize(1.0, 8.5)} & 17 {\scriptsize(10.9, 25.5)} & 16 {\scriptsize(10.1, 24.4)} & 2 {\scriptsize(0.6, 7.0)} & 2 {\scriptsize(0.6, 7.0)} \\
{\scriptsize\nolinkurl{adjust_bottle_dustbin}} & 94 {\scriptsize(87.5, 97.2)} & 6 {\scriptsize(2.8, 12.5)} & 99 {\scriptsize(94.6, 99.8)} & \textbf{100} {\scriptsize(96.3, 100.0)} & 84 {\scriptsize(75.6, 89.9)} \\
{\scriptsize\nolinkurl{click_bell_press_stapler}} & 0 {\scriptsize(0.0, 3.7)} & 7 {\scriptsize(3.4, 13.7)} & \textbf{95} {\scriptsize(88.8, 97.8)} & 0 {\scriptsize(0.0, 3.7)} & 12 {\scriptsize(7.0, 19.8)} \\
{\scriptsize\nolinkurl{serve_teabox_then_ring}} & 0 {\scriptsize(0.0, 3.7)} & 0 {\scriptsize(0.0, 3.7)} & 0 {\scriptsize(0.0, 3.7)} & 0 {\scriptsize(0.0, 3.7)} & \textbf{1} {\scriptsize(0.2, 5.4)} \\
{\scriptsize\nolinkurl{stapler_pad_press}} & 15 {\scriptsize(9.3, 23.3)} & \textbf{100} {\scriptsize(96.3, 100.0)} & 73 {\scriptsize(63.6, 80.7)} & 83 {\scriptsize(74.5, 89.1)} & 95 {\scriptsize(88.8, 97.8)} \\
{\scriptsize\nolinkurl{lift_pot_to_pad}} & 0 {\scriptsize(0.0, 3.7)} & 0 {\scriptsize(0.0, 3.7)} & 0 {\scriptsize(0.0, 3.7)} & 0 {\scriptsize(0.0, 3.7)} & 0 {\scriptsize(0.0, 3.7)} \\
{\scriptsize\nolinkurl{pack_toycar_then_ring}} & 0 {\scriptsize(0.0, 3.7)} & 6 {\scriptsize(2.8, 12.5)} & 8 {\scriptsize(4.1, 15.0)} & 0 {\scriptsize(0.0, 3.7)} & 17 {\scriptsize(10.9, 25.5)} \\
{\scriptsize\nolinkurl{move_cup_pad_mix}} & 1 {\scriptsize(0.2, 5.4)} & 35 {\scriptsize(26.4, 44.7)} & 0 {\scriptsize(0.0, 3.7)} & 30 {\scriptsize(21.9, 39.6)} & 39 {\scriptsize(30.0, 48.8)} \\
{\scriptsize\nolinkurl{ref_block_on_plate_mix}} & 0 {\scriptsize(0.0, 3.7)} & 0 {\scriptsize(0.0, 3.7)} & 0 {\scriptsize(0.0, 3.7)} & 0 {\scriptsize(0.0, 3.7)} & 0 {\scriptsize(0.0, 3.7)} \\
{\scriptsize\nolinkurl{attr_teabox_basket_then_ring}} & 0 {\scriptsize(0.0, 3.7)} & 0 {\scriptsize(0.0, 3.7)} & 0 {\scriptsize(0.0, 3.7)} & 0 {\scriptsize(0.0, 3.7)} & 0 {\scriptsize(0.0, 3.7)} \\
{\scriptsize\nolinkurl{card_right_of_scanner_mix}} & 0 {\scriptsize(0.0, 3.7)} & 0 {\scriptsize(0.0, 3.7)} & 5 {\scriptsize(2.2, 11.2)} & 0 {\scriptsize(0.0, 3.7)} & \textbf{56} {\scriptsize(46.2, 65.3)} \\
{\scriptsize\nolinkurl{place_teabox_skillet_mix}} & 0 {\scriptsize(0.0, 3.7)} & 0 {\scriptsize(0.0, 3.7)} & 0 {\scriptsize(0.0, 3.7)} & 0 {\scriptsize(0.0, 3.7)} & \textbf{10} {\scriptsize(5.5, 17.4)} \\
{\scriptsize\nolinkurl{put_toycar_left_basket_mix}} & 0 {\scriptsize(0.0, 3.7)} & 0 {\scriptsize(0.0, 3.7)} & 0 {\scriptsize(0.0, 3.7)} & 0 {\scriptsize(0.0, 3.7)} & 22 {\scriptsize(15.0, 31.1)} \\
{\scriptsize\nolinkurl{roller_to_plate_mix}} & 0 {\scriptsize(0.0, 3.7)} & 1 {\scriptsize(0.2, 5.4)} & 0 {\scriptsize(0.0, 3.7)} & 0 {\scriptsize(0.0, 3.7)} & 36 {\scriptsize(27.3, 45.8)} \\
\end{longtable}\endgroup

\begingroup\footnotesize\setlength{\tabcolsep}{4pt}
\begin{longtable}{lccc}
\caption{Per-task policy success (\%), part 3 of 3: X-WAM, Motus, and LingBot-VA. Each cell gives success and a 95\% Wilson interval over 100 trials. Bold marks taskwise maxima across all twelve policies, including ties. {Printed with one line per task in the revision; values, in \%, are unchanged.}}\label{tab:remaining-policy-scenes}\\
\toprule
Scene & X-WAM$\uparrow$ & Motus$\uparrow$ & LingBot-VA$\uparrow$ \\
\midrule\endfirsthead
\multicolumn{4}{l}{Table \thetable{} continued: task success (95\% Wilson interval).}\\
\toprule
Scene & X-WAM$\uparrow$ & Motus$\uparrow$ & LingBot-VA$\uparrow$ \\
\midrule\endhead
\midrule\multicolumn{4}{r}{Continued on next page.}\\\endfoot
\bottomrule\endlastfoot
{\scriptsize\nolinkurl{grab_dumbbell}} & \textbf{97} {\scriptsize(91.5, 99.0)} & 62 {\scriptsize(52.2, 70.9)} & 86 {\scriptsize(77.9, 91.5)} \\
{\scriptsize\nolinkurl{move_can_coaster}} & \textbf{72} {\scriptsize(62.5, 79.9)} & 14 {\scriptsize(8.5, 22.1)} & 14 {\scriptsize(8.5, 22.1)} \\
{\scriptsize\nolinkurl{move_cup_pad}} & 92 {\scriptsize(85.0, 95.9)} & \textbf{99} {\scriptsize(94.6, 99.8)} & 93 {\scriptsize(86.3, 96.6)} \\
{\scriptsize\nolinkurl{place_teabox_scale}} & \textbf{83} {\scriptsize(74.5, 89.1)} & 28 {\scriptsize(20.1, 37.5)} & 49 {\scriptsize(39.4, 58.7)} \\
{\scriptsize\nolinkurl{grasp_mallet_novel}} & \textbf{98} {\scriptsize(93.0, 99.4)} & \textbf{98} {\scriptsize(93.0, 99.4)} & 97 {\scriptsize(91.5, 99.0)} \\
{\scriptsize\nolinkurl{lift_trophy}} & \textbf{87} {\scriptsize(79.0, 92.2)} & 52 {\scriptsize(42.3, 61.5)} & 69 {\scriptsize(59.4, 77.2)} \\
{\scriptsize\nolinkurl{place_seal_coaster}} & \textbf{60} {\scriptsize(50.2, 69.1)} & 59 {\scriptsize(49.2, 68.1)} & 8 {\scriptsize(4.1, 15.0)} \\
{\scriptsize\nolinkurl{place_teabox_skillet}} & \textbf{29} {\scriptsize(21.0, 38.5)} & 0 {\scriptsize(0.0, 3.7)} & 0 {\scriptsize(0.0, 3.7)} \\
{\scriptsize\nolinkurl{place_teabox_stand}} & \textbf{79} {\scriptsize(70.0, 85.8)} & 51 {\scriptsize(41.3, 60.6)} & 75 {\scriptsize(65.7, 82.5)} \\
{\scriptsize\nolinkurl{place_toycar_plasticbox}} & \textbf{86} {\scriptsize(77.9, 91.5)} & 48 {\scriptsize(38.5, 57.7)} & 67 {\scriptsize(57.3, 75.4)} \\
{\scriptsize\nolinkurl{place_a2b_left_novel}} & 55 {\scriptsize(45.2, 64.4)} & 60 {\scriptsize(50.2, 69.1)} & 64 {\scriptsize(54.2, 72.7)} \\
{\scriptsize\nolinkurl{place_a2b_right_novel}} & 56 {\scriptsize(46.2, 65.3)} & 64 {\scriptsize(54.2, 72.7)} & 49 {\scriptsize(39.4, 58.7)} \\
{\scriptsize\nolinkurl{card_right_of_scanner}} & 5 {\scriptsize(2.2, 11.2)} & 1 {\scriptsize(0.2, 5.4)} & 7 {\scriptsize(3.4, 13.7)} \\
{\scriptsize\nolinkurl{phone_right_of_laptop}} & 14 {\scriptsize(8.5, 22.1)} & 33 {\scriptsize(24.6, 42.7)} & 2 {\scriptsize(0.6, 7.0)} \\
{\scriptsize\nolinkurl{place_can_right_of_plasticbox}} & 3 {\scriptsize(1.0, 8.5)} & \textbf{87} {\scriptsize(79.0, 92.2)} & 29 {\scriptsize(21.0, 38.5)} \\
{\scriptsize\nolinkurl{place_cube_right_of_plate}} & 20 {\scriptsize(13.3, 28.9)} & \textbf{62} {\scriptsize(52.2, 70.9)} & 48 {\scriptsize(38.5, 57.7)} \\
{\scriptsize\nolinkurl{place_teabox_front_of_skillet}} & 4 {\scriptsize(1.6, 9.8)} & 1 {\scriptsize(0.2, 5.4)} & 4 {\scriptsize(1.6, 9.8)} \\
{\scriptsize\nolinkurl{put_toycar_left_basket}} & 21 {\scriptsize(14.2, 30.0)} & \textbf{31} {\scriptsize(22.8, 40.6)} & 6 {\scriptsize(2.8, 12.5)} \\
{\scriptsize\nolinkurl{place_between_bowls}} & 34 {\scriptsize(25.5, 43.7)} & 17 {\scriptsize(10.9, 25.5)} & \textbf{53} {\scriptsize(43.3, 62.5)} \\
{\scriptsize\nolinkurl{attr_bind_bottle}} & 62 {\scriptsize(52.2, 70.9)} & 69 {\scriptsize(59.4, 77.2)} & 66 {\scriptsize(56.3, 74.5)} \\
{\scriptsize\nolinkurl{attr_category_press_bell}} & 86 {\scriptsize(77.9, 91.5)} & \textbf{100} {\scriptsize(96.3, 100.0)} & 78 {\scriptsize(68.9, 85.0)} \\
{\scriptsize\nolinkurl{attr_category_can_plasticbox}} & 41 {\scriptsize(31.9, 50.8)} & 32 {\scriptsize(23.7, 41.7)} & 3 {\scriptsize(1.0, 8.5)} \\
{\scriptsize\nolinkurl{attr_color_beat_block}} & 69 {\scriptsize(59.4, 77.2)} & 55 {\scriptsize(45.2, 64.4)} & \textbf{91} {\scriptsize(83.8, 95.2)} \\
{\scriptsize\nolinkurl{attr_color_block_3way}} & 62 {\scriptsize(52.2, 70.9)} & 68 {\scriptsize(58.3, 76.3)} & \textbf{99} {\scriptsize(94.6, 99.8)} \\
{\scriptsize\nolinkurl{attr_size_beat_smaller}} & 49 {\scriptsize(39.4, 58.7)} & 27 {\scriptsize(19.3, 36.4)} & 35 {\scriptsize(26.4, 44.7)} \\
{\scriptsize\nolinkurl{attr_size_block_pick}} & \textbf{100} {\scriptsize(96.3, 100.0)} & 83 {\scriptsize(74.5, 89.1)} & 69 {\scriptsize(59.4, 77.2)} \\
{\scriptsize\nolinkurl{attr_bind_soap_box}} & 71 {\scriptsize(61.5, 79.0)} & 72 {\scriptsize(62.5, 79.9)} & \textbf{79} {\scriptsize(70.0, 85.8)} \\
{\scriptsize\nolinkurl{attr_size_rank_middle}} & 1 {\scriptsize(0.2, 5.4)} & 6 {\scriptsize(2.8, 12.5)} & \textbf{28} {\scriptsize(20.1, 37.5)} \\
{\scriptsize\nolinkurl{ref_block_on_plate}} & 0 {\scriptsize(0.0, 3.7)} & 0 {\scriptsize(0.0, 3.7)} & 6 {\scriptsize(2.8, 12.5)} \\
{\scriptsize\nolinkurl{ref_block_behind_can}} & 18 {\scriptsize(11.7, 26.7)} & 11 {\scriptsize(6.3, 18.6)} & 7 {\scriptsize(3.4, 13.7)} \\
{\scriptsize\nolinkurl{ref_cup_on_tray}} & 0 {\scriptsize(0.0, 3.7)} & 1 {\scriptsize(0.2, 5.4)} & 7 {\scriptsize(3.4, 13.7)} \\
{\scriptsize\nolinkurl{ref_leftmost_of_four}} & 50 {\scriptsize(40.4, 59.6)} & 37 {\scriptsize(28.2, 46.8)} & 25 {\scriptsize(17.5, 34.3)} \\
{\scriptsize\nolinkurl{ref_mug_next_to_bowl}} & 46 {\scriptsize(36.6, 55.7)} & 18 {\scriptsize(11.7, 26.7)} & \textbf{70} {\scriptsize(60.4, 78.1)} \\
{\scriptsize\nolinkurl{ref_not_the_red_one}} & 18 {\scriptsize(11.7, 26.7)} & \textbf{25} {\scriptsize(17.5, 34.3)} & 0 {\scriptsize(0.0, 3.7)} \\
{\scriptsize\nolinkurl{ref_second_from_left}} & 10 {\scriptsize(5.5, 17.4)} & 2 {\scriptsize(0.6, 7.0)} & \textbf{62} {\scriptsize(52.2, 70.9)} \\
{\scriptsize\nolinkurl{adjust_bottle_dustbin}} & 72 {\scriptsize(62.5, 79.9)} & 85 {\scriptsize(76.7, 90.7)} & 79 {\scriptsize(70.0, 85.8)} \\
{\scriptsize\nolinkurl{click_bell_press_stapler}} & 1 {\scriptsize(0.2, 5.4)} & 59 {\scriptsize(49.2, 68.1)} & 4 {\scriptsize(1.6, 9.8)} \\
{\scriptsize\nolinkurl{serve_teabox_then_ring}} & \textbf{1} {\scriptsize(0.2, 5.4)} & 0 {\scriptsize(0.0, 3.7)} & 0 {\scriptsize(0.0, 3.7)} \\
{\scriptsize\nolinkurl{stapler_pad_press}} & 95 {\scriptsize(88.8, 97.8)} & 92 {\scriptsize(85.0, 95.9)} & \textbf{100} {\scriptsize(96.3, 100.0)} \\
{\scriptsize\nolinkurl{lift_pot_to_pad}} & 2 {\scriptsize(0.6, 7.0)} & 9 {\scriptsize(4.8, 16.2)} & \textbf{34} {\scriptsize(25.5, 43.7)} \\
{\scriptsize\nolinkurl{pack_toycar_then_ring}} & \textbf{35} {\scriptsize(26.4, 44.7)} & 3 {\scriptsize(1.0, 8.5)} & 21 {\scriptsize(14.2, 30.0)} \\
{\scriptsize\nolinkurl{move_cup_pad_mix}} & 9 {\scriptsize(4.8, 16.2)} & 20 {\scriptsize(13.3, 28.9)} & 73 {\scriptsize(63.6, 80.7)} \\
{\scriptsize\nolinkurl{ref_block_on_plate_mix}} & 0 {\scriptsize(0.0, 3.7)} & 0 {\scriptsize(0.0, 3.7)} & 11 {\scriptsize(6.3, 18.6)} \\
{\scriptsize\nolinkurl{attr_teabox_basket_then_ring}} & 1 {\scriptsize(0.2, 5.4)} & 1 {\scriptsize(0.2, 5.4)} & 0 {\scriptsize(0.0, 3.7)} \\
{\scriptsize\nolinkurl{card_right_of_scanner_mix}} & 2 {\scriptsize(0.6, 7.0)} & 0 {\scriptsize(0.0, 3.7)} & 6 {\scriptsize(2.8, 12.5)} \\
{\scriptsize\nolinkurl{place_teabox_skillet_mix}} & 0 {\scriptsize(0.0, 3.7)} & 0 {\scriptsize(0.0, 3.7)} & 0 {\scriptsize(0.0, 3.7)} \\
{\scriptsize\nolinkurl{put_toycar_left_basket_mix}} & 18 {\scriptsize(11.7, 26.7)} & \textbf{28} {\scriptsize(20.1, 37.5)} & 5 {\scriptsize(2.2, 11.2)} \\
{\scriptsize\nolinkurl{roller_to_plate_mix}} & 20 {\scriptsize(13.3, 28.9)} & 6 {\scriptsize(2.8, 12.5)} & 38 {\scriptsize(29.1, 47.8)} \\
\end{longtable}\endgroup

\subsection{Agent Task Results}
\label{app:gpt6-preview}

GPT-6 Astra succeeds in 185/240 episodes, with family means of 84.00, 77.78, 71.11, 97.14, 56.67, and 71.43\%, yielding Overall 76.35\%.

Gemini 3.8 Flash succeeds in 58/240 episodes, with family means of 18.00, 24.44, 22.22, 45.71, 10.00, and 25.71\%, yielding Overall 24.35\%. Table~\ref{tab:gpt6-scenes} gives {all agents\textquotesingle\ per-task results; Table~\ref{tab:matched5-complete} reports their aggregate success and timing}.

\begingroup\footnotesize\setlength{\tabcolsep}{3pt}
\begin{longtable}{@{}lrrrrr@{}}
\caption{Agent task success (\%) on five frozen seeds per scene. Bold marks the highest score per scene, including ties. Request settings are given in Table~\ref{tab:agent-config}.}\label{tab:gpt6-scenes}\\
\toprule
Scene & \shortstack{GPT-6\\Astra} & {\shortstack{GPT-6.2\\Astra}} & {\shortstack{Opus\\5.5}} & {\shortstack{Gemini\\4 Argon}} & \shortstack{Gemini\\3.8 Flash} \\
\midrule\endfirsthead
\multicolumn{6}{l}{Table \thetable{} continued: agent task success (\%).}\\
\toprule
Scene & \shortstack{GPT-6\\Astra} & {\shortstack{GPT-6.2\\Astra}} & {\shortstack{Opus\\5.5}} & {\shortstack{Gemini\\4 Argon}} & \shortstack{Gemini\\3.8 Flash} \\
\midrule\endhead
\midrule\multicolumn{6}{r}{Continued on next page.}\\\endfoot
\bottomrule\endlastfoot
{\scriptsize\nolinkurl{grab_dumbbell}} & \textbf{100} & {\textbf{100}} & {80} & {40} & 20 \\
{\scriptsize\nolinkurl{move_can_coaster}} & \textbf{60} & {20} & {20} & {20} & 20 \\
{\scriptsize\nolinkurl{move_cup_pad}} & \textbf{80} & {20} & {60} & {40} & 0 \\
{\scriptsize\nolinkurl{place_teabox_scale}} & \textbf{80} & {\textbf{80}} & {\textbf{80}} & {40} & 0 \\
{\scriptsize\nolinkurl{grasp_mallet_novel}} & \textbf{100} & {80} & {40} & {80} & 20 \\
{\scriptsize\nolinkurl{lift_trophy}} & \textbf{100} & {\textbf{100}} & {20} & {20} & 0 \\
{\scriptsize\nolinkurl{place_seal_coaster}} & 80 & {\textbf{100}} & {60} & {20} & 20 \\
{\scriptsize\nolinkurl{place_teabox_skillet}} & 80 & {\textbf{100}} & {80} & {20} & 60 \\
{\scriptsize\nolinkurl{place_teabox_stand}} & 80 & {\textbf{100}} & {60} & {60} & 0 \\
{\scriptsize\nolinkurl{place_toycar_plasticbox}} & \textbf{80} & {\textbf{80}} & {40} & {40} & 40 \\
{\scriptsize\nolinkurl{place_a2b_left_novel}} & 80 & {\textbf{100}} & {60} & {0} & 60 \\
{\scriptsize\nolinkurl{place_a2b_right_novel}} & \textbf{100} & {\textbf{100}} & {40} & {20} & 40 \\
{\scriptsize\nolinkurl{card_right_of_scanner}} & 40 & {\textbf{60}} & {20} & {0} & 20 \\
{\scriptsize\nolinkurl{phone_right_of_laptop}} & 80 & {\textbf{100}} & {40} & {40} & 0 \\
{\scriptsize\nolinkurl{place_can_right_of_plasticbox}} & \textbf{40} & {\textbf{40}} & {20} & {20} & 0 \\
{\scriptsize\nolinkurl{place_cube_right_of_plate}} & \textbf{100} & {\textbf{100}} & {60} & {40} & 80 \\
{\scriptsize\nolinkurl{place_teabox_front_of_skillet}} & \textbf{100} & {\textbf{100}} & {80} & {0} & 20 \\
{\scriptsize\nolinkurl{put_toycar_left_basket}} & \textbf{60} & {20} & {40} & {0} & 0 \\
{\scriptsize\nolinkurl{place_between_bowls}} & \textbf{100} & {\textbf{100}} & {\textbf{100}} & {0} & 0 \\
{\scriptsize\nolinkurl{attr_bind_bottle}} & 20 & {80} & {\textbf{100}} & {40} & 0 \\
{\scriptsize\nolinkurl{attr_category_press_bell}} & \textbf{100} & {\textbf{100}} & {\textbf{100}} & {\textbf{100}} & \textbf{100} \\
{\scriptsize\nolinkurl{attr_category_can_plasticbox}} & \textbf{80} & {60} & {60} & {0} & 0 \\
{\scriptsize\nolinkurl{attr_color_beat_block}} & 20 & {\textbf{40}} & {\textbf{40}} & {0} & 0 \\
{\scriptsize\nolinkurl{attr_color_block_3way}} & \textbf{100} & {\textbf{100}} & {60} & {40} & 20 \\
{\scriptsize\nolinkurl{attr_size_beat_smaller}} & \textbf{40} & {\textbf{40}} & {20} & {0} & 0 \\
{\scriptsize\nolinkurl{attr_size_block_pick}} & \textbf{100} & {\textbf{100}} & {\textbf{100}} & {40} & 40 \\
{\scriptsize\nolinkurl{attr_bind_soap_box}} & \textbf{80} & {\textbf{80}} & {40} & {20} & 0 \\
{\scriptsize\nolinkurl{attr_size_rank_middle}} & \textbf{100} & {\textbf{100}} & {80} & {80} & 40 \\
{\scriptsize\nolinkurl{ref_block_on_plate}} & \textbf{100} & {80} & {80} & {20} & 40 \\
{\scriptsize\nolinkurl{ref_block_behind_can}} & \textbf{100} & {80} & {\textbf{100}} & {60} & 0 \\
{\scriptsize\nolinkurl{ref_cup_on_tray}} & \textbf{100} & {80} & {\textbf{100}} & {60} & 20 \\
{\scriptsize\nolinkurl{ref_leftmost_of_four}} & \textbf{80} & {\textbf{80}} & {\textbf{80}} & {60} & \textbf{80} \\
{\scriptsize\nolinkurl{ref_mug_next_to_bowl}} & \textbf{100} & {\textbf{100}} & {60} & {20} & 20 \\
{\scriptsize\nolinkurl{ref_not_the_red_one}} & \textbf{100} & {\textbf{100}} & {\textbf{100}} & {40} & 60 \\
{\scriptsize\nolinkurl{ref_second_from_left}} & \textbf{100} & {80} & {\textbf{100}} & {20} & \textbf{100} \\
{\scriptsize\nolinkurl{adjust_bottle_dustbin}} & 20 & {\textbf{100}} & {60} & {0} & 0 \\
{\scriptsize\nolinkurl{click_bell_press_stapler}} & 80 & {80} & {\textbf{100}} & {40} & 40 \\
{\scriptsize\nolinkurl{serve_teabox_then_ring}} & \textbf{60} & {40} & {0} & {20} & 20 \\
{\scriptsize\nolinkurl{stapler_pad_press}} & 80 & {\textbf{100}} & {60} & {80} & 0 \\
{\scriptsize\nolinkurl{lift_pot_to_pad}} & 0 & {0} & {\textbf{40}} & {0} & 0 \\
{\scriptsize\nolinkurl{pack_toycar_then_ring}} & \textbf{100} & {80} & {60} & {40} & 0 \\
{\scriptsize\nolinkurl{move_cup_pad_mix}} & \textbf{80} & {40} & {60} & {20} & 0 \\
{\scriptsize\nolinkurl{ref_block_on_plate_mix}} & \textbf{100} & {\textbf{100}} & {\textbf{100}} & {60} & 80 \\
{\scriptsize\nolinkurl{attr_teabox_basket_then_ring}} & 20 & {20} & {\textbf{40}} & {0} & 0 \\
{\scriptsize\nolinkurl{card_right_of_scanner_mix}} & 60 & {\textbf{100}} & {40} & {0} & 40 \\
{\scriptsize\nolinkurl{place_teabox_skillet_mix}} & \textbf{100} & {80} & {80} & {20} & 60 \\
{\scriptsize\nolinkurl{put_toycar_left_basket_mix}} & 40 & {20} & {\textbf{60}} & {0} & 0 \\
{\scriptsize\nolinkurl{roller_to_plate_mix}} & \textbf{100} & {\textbf{100}} & {60} & {40} & 0 \\
\end{longtable}
\noindent\footnotesize With five trials, success rates 0, 20, 40, 60, 80, and 100\% correspond to 95\% Wilson intervals (0.0, 43.4), (3.6, 62.4), (11.8, 76.9), (23.1, 88.2), (37.6, 96.4), and (56.6, 100.0)\%, respectively.
\endgroup

\FloatBarrier
{\section{Target Reassignment and Its Controls}
\label{app:diagnostic-scope}

This appendix documents the diagnostic chain of Table~\ref{tab:diagnostic-chain}: what each control separates (Appendix~\ref{app:what-separates}), target reassignment on the held-out scenes (Appendix~\ref{app:reassignment}), the controls on RoboTwin training and source tasks (Appendix~\ref{app:reassignment-controls}), and what the results imply about cause and remedy (Appendix~\ref{app:cause-hypotheses}).

\subsection{Design: What Each Control Separates}
\label{app:what-separates}

A held-out scene differs from training in its request and in its visual configuration, and each formal scene fixes one requested rule. A failure under the formal instruction can therefore come from a missing prerequisite skill, from the changed scene, or from a choice that does not depend on the instruction; a success can come from a preference that coincides with the request. Four measurements separate these cases. Commitment records separate moving a wrong object from moving nothing (Section~\ref{sec:selection-metrics}). Target reassignment on frozen initial states separates a fixed choice from instruction following (Appendix~\ref{app:reassignment}). The same reassignment on unmodified RoboTwin training tasks removes the scene change. Source-task success in the same evaluator identifies missing prerequisite skills.

The selection scenes pose a question that the training tasks do not. In the three RoboTwin block tasks from which they derive (\nolinkurl{stack_blocks_three}, \nolinkurl{blocks_ranking_rgb}, \nolinkurl{blocks_ranking_size}), each task has one fixed goal and its instruction templates paraphrase that goal, so no instruction has to distinguish one block from another. On \nolinkurl{blocks_ranking_size}, ten checkpoints displace the smallest block first in 193 of 200 runs under the training instructions, in 182 of 200 runs that ask for the largest block, and in 164 of 200 runs with an empty instruction (Table~\ref{tab:training-scene}). The instruction still matters for completing the task: 161 of 200 runs succeed under the training instructions and 71 of 200 with an empty instruction.

\paragraph{Why the scenes come without demonstrations.}
The two extremes confound in opposite ways: a held-out scene lets a policy fail for reasons other than language, and a training scene lets it succeed without language. \Benchmark{} therefore draws objects and skills from the RoboTwin training tasks and makes the scene composition and the request new, and the controls of this appendix separate the remaining confound. Demonstrations paired with a scene would let a policy answer from visual memory: it recognizes the scene and reproduces the demonstrated routine, and the instruction need only act as a label that an arbitrary code could replace. Both parts appear on the training tasks. Under an empty instruction, the verified checkpoints displace the red block first in 110 of 120 runs on the two color tasks (Appendix~\ref{app:reassignment-controls}): the scene alone carries the choice. X-VLA, conditioned on task-name strings, completes two training tasks in 28/30 and 16/30 episodes with those strings and in 9/30 and 0/30 with natural-language instructions (Appendix~\ref{app:checkpoints}): a label suffices. Success on a scene with paired demonstrations therefore does not show that the content of the instruction is used. A held-out request is not the answer of any memorized scene. Under ``Pick up the block that is not red.'', ten checkpoints displace the requested block first in 11 of 100 runs, and the positive control follows both instructions of that scene on 5 of 5 states (Tables~\ref{tab:reassignment-instruction} and~\ref{tab:reassignment-scene}). Withholding demonstrations removes recall as a route to success. It does not remove a preference inherited from the source tasks, which target reassignment tests.

\subsection{Target Reassignment on Held-Out Scenes}
\label{app:reassignment}

\paragraph{Protocol.}
Target reassignment uses ten pick-type selection scenes: three attribute-selection scenes (color, size, size rank) and the seven referring-expression scenes. For a scene with $k$ candidates, each initial state $s$ is replayed under $k$ instructions, each designating a different candidate (for example ``Pick up the red/green/blue block.''), and under an empty instruction. Object poses, robot state, and seed are identical across the $k+1$ runs, and the evaluator's target is the designated candidate. Let $c_{s,j}=1$ when the first candidate displaced by at least 5\,cm under instruction $j$ is candidate $j$. All-$k$ correctness is
\begin{equation}
 P_{\text{all-}k}=\frac{1}{S}\sum_{s}\prod_{j=1}^{k}c_{s,j}.
 \label{eq:reassignment}
\end{equation}
Each state is \emph{follow} (every instruction followed), \emph{fixed} (the same object first under every instruction), \emph{no movement} (nothing displaced under any instruction), or \emph{mixed}. An instruction-blind chooser, which picks uniformly among the candidates whatever the instruction, gives \emph{follow} with probability $k^{-k}$ and \emph{fixed} with probability $k^{1-k}$. Every checkpoint is run on the first ten initial states of each scene; proportions carry 95\% Wilson intervals. The intervals treat state--checkpoint pairs as independent although the checkpoints share the initial states, so the pooled interval is descriptive; per-scene counts are reported with it. A \emph{follow} share below the chooser's indicates systematic choices: a fixed preference follows every instruction on no state.

\paragraph{Reading rules.}
The rules were fixed before data collection. (i) Formal success with a selection that stays at the default preference under reassignment is reported as a fixed preference. (ii) Selection that follows the reassigned target is reported as instruction-dependent. (iii) The same reassignment on unmodified RoboTwin training tasks tests whether the scene change explains a fixed choice. (iv) A checkpoint enters training-task conclusions only if the published clean-setting success on the task does not exceed the 95\% Wilson upper bound of ours. (v) A checkpoint--scene pair with a linked source task below 30\% success is not read as failing to follow.

\paragraph{Results.}
Table~\ref{tab:reassignment} pools the scenes per checkpoint, Table~\ref{tab:reassignment-scene} gives the classes per checkpoint and scene, and Table~\ref{tab:reassignment-instruction} the selection under each instruction; Figure~\ref{fig:probe-overview} shows both pooled over the ten checkpoints that enter pooled statements (all but X-VLA and $\pi_0$; Appendix~\ref{app:checkpoints}). Every instruction is followed on 39 of 1,000 initial states (3.9\%) and the same object is displaced first under every instruction on 679 (68\%); the instruction-blind chooser gives 16\% and 33\%. \emph{Follow} is 20/100 in the three-color scene, where LingBot-VA follows all three instructions on 10/10 states and $\pi_{0.5}$ on 6/10, and at most 5/100 in each other scene. Selection tracks a preference: the smaller block is displaced first in 89/100 runs that request it and the larger in 8/100; the block next to the plate in 99/100 and the block on the plate in 2/100. In \nolinkurl{ref_block_behind_can}, 34 of 100 states show no movement.

The positive control is the GPT-6 agent (Appendix~\ref{app:agents}), run on every instructed condition of five initial states per scene, the number of episodes per scene of the agent evaluation (Section~\ref{sec:setup}). It follows every instruction on 44 of 50 states (88\%; 95\% Wilson interval 76--94\%) and displaces the designated object first in 118 of 125 runs (Table~\ref{tab:reassignment-scene}), so the reassigned requests are solvable in these scenes, and the control gives an upper reference for the policies. Figure~\ref{fig:keyframes-w1} shows one initial state under each instruction, on a color training task and on the held-out three-color scene.}

\begin{figure}[!htbp]\centering
\includegraphics[width=\textwidth]{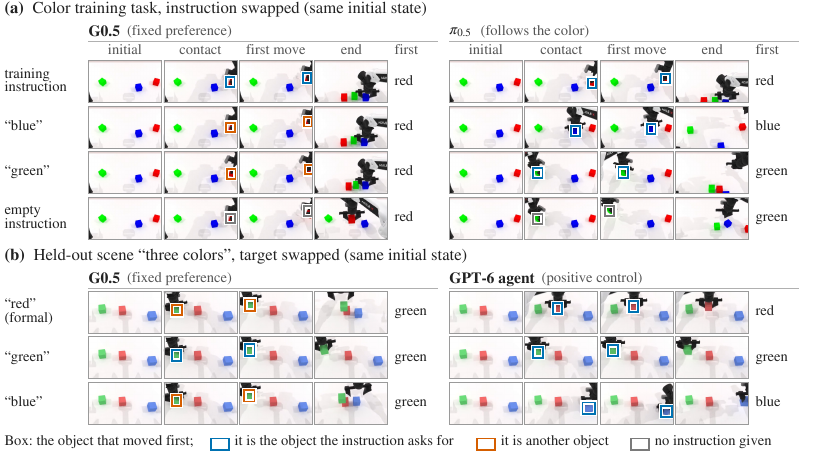}
\caption{{One initial state under different instructions. Each row is one episode; the box marks the object that moved first (blue: the requested object; vermillion: another object; gray: no instruction). (a) On a color training task, G0.5 moves the red block, which the demonstrations handle first, under every instruction, while $\pi_{0.5}$ moves the requested block. (b) On the held-out three-color scene, G0.5 moves the same block whatever is requested, while the GPT-6 agent follows each instruction.}}
\label{fig:keyframes-w1}
\end{figure}

\begin{table}[!htbp]
\centering\scriptsize\color{black}
\setlength{\tabcolsep}{3pt}
\caption{{Target reassignment by checkpoint and scene, ten initial states per cell. Each cell gives \emph{follow}/\emph{fixed}/\emph{no movement}: states on which every instruction is followed, on which the same object is displaced first under every instruction, and on which nothing is displaced; the remainder of the ten is \emph{mixed}. The agent rows are the positive control: five states per cell and $k$ instructions per state (125 runs), with the instructed runs in which the designated object is displaced first and in which the task succeeds. Scene codes: C3~=~\nolinkurl{attr_color_block_3way}; SZ~=~\nolinkurl{attr_size_block_pick}; RK~=~\nolinkurl{attr_size_rank_middle}; PL~=~\nolinkurl{ref_block_on_plate}; CN~=~\nolinkurl{ref_block_behind_can}; TR~=~\nolinkurl{ref_cup_on_tray}; L4~=~\nolinkurl{ref_leftmost_of_four}; MG~=~\nolinkurl{ref_mug_next_to_bowl}; NR~=~\nolinkurl{ref_not_the_red_one}; 2L~=~\nolinkurl{ref_second_from_left}. The last rows give the instruction-blind expectation per state, $k^{-k}$ and $k^{1-k}$. $^{\S}$, $^{\P}$: as in Table~\ref{tab:reassignment}.}}
\label{tab:reassignment-scene}
\begin{tabular}{@{}lcccccccccc@{}}
\toprule
Checkpoint & C3 & SZ & RK & PL & CN & TR & L4 & MG & NR & 2L \\
$k$ & 3 & 2 & 3 & 2 & 2 & 2 & 4 & 2 & 2 & 3 \\
\midrule
DM0.5 & 0/7/0 & 0/10/0 & 0/9/0 & 0/10/0 & 0/0/4 & 0/7/1 & 0/3/0 & 0/5/0 & 0/10/0 & 0/4/0 \\
$\pi_{0.5}$ & 6/0/0 & 1/7/0 & 0/7/0 & 0/10/0 & 0/2/1 & 0/9/1 & 0/4/0 & 1/6/0 & 0/8/0 & 0/3/0 \\
StarVLA & 3/1/0 & 0/7/2 & 0/9/0 & 1/7/0 & 0/1/6 & 4/6/0 & 0/0/0 & 0/8/0 & 0/10/0 & 1/3/0 \\
FastWAM & 0/6/0 & 0/9/0 & 0/8/1 & 0/9/1 & 0/2/5 & 0/6/3 & 0/7/0 & 0/10/0 & 0/6/0 & 0/9/0 \\
$\pi_0$$^{\P}$ & 0/7/0 & 0/1/6 & 0/2/4 & 0/7/1 & 0/3/3 & 0/4/2 & 0/3/0 & 0/9/0 & 0/5/1 & 0/3/0 \\
AHA-WAM & 0/6/0 & 0/9/0 & 0/8/0 & 0/10/0 & 0/3/6 & 0/10/0 & 0/4/0 & 1/8/0 & 0/7/0 & 0/6/0 \\
X-VLA$^{\S}$ & 0/6/0 & 0/10/0 & 0/5/2 & 0/8/2 & 0/4/4 & 0/8/1 & 0/4/0 & 0/8/0 & 0/9/0 & 0/7/0 \\
WLA & 0/9/0 & 0/10/0 & 0/10/0 & 0/10/0 & 0/5/4 & 0/9/1 & 0/9/0 & 0/9/0 & 0/10/0 & 0/9/0 \\
G0.5 & 1/2/0 & 0/10/0 & 0/10/0 & 0/10/0 & 0/2/3 & 0/8/2 & 0/2/0 & 0/8/1 & 0/10/0 & 0/2/0 \\
X-WAM & 0/9/0 & 0/10/0 & 0/10/0 & 0/10/0 & 1/5/0 & 0/9/1 & 0/8/0 & 0/8/0 & 0/10/0 & 0/8/0 \\
Motus & 0/5/0 & 0/9/0 & 0/9/0 & 0/10/0 & 0/1/3 & 0/9/1 & 0/7/0 & 1/7/0 & 1/8/0 & 0/7/0 \\
LingBot-VA & 10/0/0 & 2/7/0 & 0/6/0 & 1/9/0 & 0/5/2 & 0/9/0 & 1/2/0 & 2/7/0 & 0/2/0 & 1/3/0 \\
\midrule
GPT-6 agent & 5/0/0 & 4/0/0 & 4/0/0 & 5/0/0 & 4/0/0 & 4/1/0 & 3/0/0 & 5/0/0 & 5/0/0 & 5/0/0 \\
\quad designated object first & 15/15 & 9/10 & 14/15 & 10/10 & 9/10 & 9/10 & 17/20 & 10/10 & 10/10 & 15/15 \\
\quad task success & 14/15 & 8/10 & 14/15 & 10/10 & 9/10 & 9/10 & 16/20 & 9/10 & 10/10 & 13/15 \\
\midrule
Blind: follow (\%) & 3.7 & 25.0 & 3.7 & 25.0 & 25.0 & 25.0 & 0.4 & 25.0 & 25.0 & 3.7 \\
Blind: fixed (\%) & 11.1 & 50.0 & 11.1 & 50.0 & 50.0 & 50.0 & 1.6 & 50.0 & 50.0 & 11.1 \\
\bottomrule
\end{tabular}
\end{table}

\begin{table}[!htbp]
\centering\scriptsize\color{black}
\setlength{\tabcolsep}{1.35pt}
\caption{{Selection under each instruction: number of the ten runs in which the first displaced object is the designated one. The first instruction of each scene is the formal one. \emph{empty}: empty-instruction runs (of ten) in which the formal target is displaced first. \emph{Pooled}: the ten checkpoints without $^{\S}$ or $^{\P}$, of 100 runs. Scene codes as in Table~\ref{tab:reassignment-scene}.}}
\label{tab:reassignment-instruction}
\begin{tabular}{@{}lrrrr@{\hspace{5pt}}rrr@{\hspace{5pt}}rrrr@{\hspace{5pt}}rrr@{\hspace{5pt}}rrr@{\hspace{5pt}}rrr@{\hspace{5pt}}rrrrr@{\hspace{5pt}}rrr@{\hspace{5pt}}rrr@{\hspace{5pt}}rrrr@{}}
\toprule
 & \multicolumn{4}{c}{C3} & \multicolumn{3}{c}{SZ} & \multicolumn{4}{c}{RK} & \multicolumn{3}{c}{PL} & \multicolumn{3}{c}{CN} & \multicolumn{3}{c}{TR} & \multicolumn{5}{c}{L4} & \multicolumn{3}{c}{MG} & \multicolumn{3}{c}{NR} & \multicolumn{4}{c}{2L} \\
\cmidrule(lr){2-5}\cmidrule(lr){6-8}\cmidrule(lr){9-12}\cmidrule(lr){13-15}\cmidrule(lr){16-18}\cmidrule(lr){19-21}\cmidrule(lr){22-26}\cmidrule(lr){27-29}\cmidrule(lr){30-32}\cmidrule(lr){33-36}
Checkpoint & \rotatebox{90}{red} & \rotatebox{90}{green} & \rotatebox{90}{blue} & \rotatebox{90}{\emph{empty}} & \rotatebox{90}{smaller} & \rotatebox{90}{larger} & \rotatebox{90}{\emph{empty}} & \rotatebox{90}{medium} & \rotatebox{90}{smallest} & \rotatebox{90}{largest} & \rotatebox{90}{\emph{empty}} & \rotatebox{90}{on the plate} & \rotatebox{90}{next to the plate} & \rotatebox{90}{\emph{empty}} & \rotatebox{90}{behind the can} & \rotatebox{90}{farther from the can} & \rotatebox{90}{\emph{empty}} & \rotatebox{90}{on the tray} & \rotatebox{90}{next to the tray} & \rotatebox{90}{\emph{empty}} & \rotatebox{90}{leftmost} & \rotatebox{90}{second from left} & \rotatebox{90}{third from left} & \rotatebox{90}{rightmost} & \rotatebox{90}{\emph{empty}} & \rotatebox{90}{next to the bowl} & \rotatebox{90}{farther from the bowl} & \rotatebox{90}{\emph{empty}} & \rotatebox{90}{not red} & \rotatebox{90}{not blue} & \rotatebox{90}{\emph{empty}} & \rotatebox{90}{second from left} & \rotatebox{90}{leftmost} & \rotatebox{90}{rightmost} & \rotatebox{90}{\emph{empty}} \\
\midrule
DM0.5 & 5 & 6 & 1 & 4 & 10 & 0 & 10 & 1 & 9 & 0 & 0 & 0 & 10 & 0 & 0 & 4 & 0 & 1 & 7 & 0 & 6 & 1 & 1 & 6 & 5 & 4 & 1 & 4 & 0 & 10 & 0 & 0 & 8 & 5 & 1 \\
$\pi_{0.5}$ & 8 & 10 & 8 & 4 & 9 & 1 & 1 & 0 & 9 & 0 & 0 & 0 & 10 & 0 & 3 & 1 & 1 & 0 & 9 & 0 & 6 & 1 & 1 & 4 & 7 & 5 & 3 & 6 & 0 & 8 & 0 & 1 & 9 & 5 & 1 \\
StarVLA & 9 & 9 & 4 & 6 & 7 & 0 & 10 & 1 & 9 & 0 & 0 & 1 & 10 & 1 & 1 & 3 & 0 & 5 & 9 & 3 & 7 & 3 & 2 & 7 & 5 & 1 & 7 & 3 & 2 & 8 & 1 & 1 & 9 & 8 & 0 \\
FastWAM & 6 & 2 & 1 & 6 & 8 & 1 & 8 & 0 & 9 & 0 & 0 & 0 & 9 & 1 & 3 & 1 & 6 & 1 & 6 & 0 & 7 & 3 & 1 & 3 & 5 & 3 & 7 & 4 & 2 & 4 & 7 & 1 & 8 & 2 & 1 \\
$\pi_0$$^{\P}$ & 3 & 5 & 0 & 6 & 2 & 1 & 1 & 0 & 3 & 0 & 1 & 0 & 9 & 0 & 6 & 0 & 6 & 0 & 5 & 1 & 1 & 1 & 1 & 3 & 1 & 2 & 7 & 1 & 0 & 7 & 0 & 0 & 3 & 4 & 1 \\
AHA-WAM & 6 & 0 & 6 & 2 & 9 & 1 & 0 & 0 & 10 & 0 & 2 & 0 & 10 & 1 & 1 & 3 & 1 & 2 & 8 & 6 & 3 & 2 & 3 & 8 & 0 & 6 & 4 & 6 & 2 & 5 & 7 & 1 & 6 & 5 & 2 \\
X-VLA$^{\S}$ & 8 & 3 & 0 & 9 & 10 & 0 & 6 & 1 & 6 & 0 & 3 & 0 & 8 & 0 & 0 & 5 & 0 & 0 & 9 & 0 & 2 & 1 & 2 & 4 & 3 & 5 & 3 & 8 & 1 & 9 & 0 & 0 & 4 & 6 & 1 \\
WLA & 4 & 6 & 0 & 5 & 10 & 0 & 10 & 0 & 10 & 0 & 0 & 0 & 10 & 0 & 0 & 5 & 0 & 0 & 9 & 0 & 5 & 0 & 1 & 4 & 4 & 5 & 4 & 5 & 1 & 9 & 1 & 1 & 6 & 3 & 1 \\
G0.5 & 6 & 9 & 4 & 5 & 10 & 0 & 10 & 0 & 10 & 0 & 0 & 0 & 10 & 0 & 1 & 1 & 0 & 0 & 8 & 0 & 6 & 2 & 0 & 8 & 5 & 1 & 7 & 0 & 1 & 9 & 1 & 0 & 10 & 7 & 0 \\
X-WAM & 4 & 5 & 2 & 4 & 10 & 0 & 10 & 0 & 10 & 0 & 0 & 0 & 10 & 0 & 5 & 4 & 5 & 0 & 9 & 0 & 7 & 1 & 0 & 1 & 7 & 4 & 4 & 3 & 2 & 8 & 3 & 0 & 8 & 4 & 0 \\
Motus & 6 & 5 & 3 & 3 & 9 & 1 & 7 & 0 & 10 & 0 & 0 & 0 & 10 & 0 & 1 & 3 & 1 & 1 & 8 & 1 & 3 & 0 & 0 & 9 & 1 & 3 & 6 & 2 & 1 & 9 & 2 & 1 & 6 & 6 & 1 \\
LingBot-VA & 10 & 10 & 10 & 7 & 7 & 4 & 7 & 3 & 7 & 0 & 6 & 1 & 10 & 0 & 1 & 5 & 1 & 2 & 7 & 2 & 3 & 3 & 7 & 3 & 1 & 7 & 4 & 2 & 0 & 2 & 7 & 8 & 6 & 4 & 3 \\
\midrule
Pooled & 64 & 62 & 39 & 46 & 89 & 8 & 73 & 5 & 93 & 0 & 8 & 2 & 99 & 3 & 16 & 30 & 15 & 12 & 80 & 12 & 53 & 16 & 16 & 53 & 40 & 39 & 47 & 35 & 11 & 72 & 29 & 14 & 76 & 49 & 10 \\
\bottomrule
\end{tabular}
\end{table}

{\subsection{Controls: Training Tasks and Source Tasks}
\label{app:reassignment-controls}

\paragraph{Training tasks: the scene change.}
The same reassignment on three unmodified RoboTwin training tasks removes the scene change (Table~\ref{tab:training-scene}; Figure~\ref{fig:probe-training}). \nolinkurl{blocks_ranking_rgb} and \nolinkurl{stack_blocks_three} contain a red, a green, and a blue block; \nolinkurl{blocks_ranking_size} contains three blocks that differ only in size. By rule (iv), six checkpoints are verified on the color tasks and nine on the size task; $\pi_{0.5}$ completes the size task in 3/20 episodes under the training instructions, as published for it, and falls under rule (v). Conclusions use the six and the remaining eight. Published values for $\pi_{0.5}$ are those of the group that released the checkpoint.

\emph{Color.} Five of the six verified checkpoints never displace the blue block first under ``Pick up the blue block.'' (0/100 runs; red first in 99), and four of them still complete the original ranking task in most of those runs. $\pi_{0.5}$ displaces the blue block first in 15/20 runs and the green block in 19/20; G0.5 displaces green first in 17/20 runs and blue in none. Under the empty instruction the verified checkpoints displace the red block first in 110/120 runs.

\emph{Size.} The scripted expert of this task handles the smallest block first. The prediction that policies displace the smallest block first under every instruction was logged before the first outcome of this task; the held-out size scenes had already shown a preference for the smaller block. Asked for the largest block, the eight checkpoints displace it first in 1 of 160 runs and the smallest in 145; asked for the medium block, they displace it first in 18 and the smallest in 139. In 112 of the 160 ``largest'' runs the original ranking task is still completed. LingBot-VA and Motus are the least fixed (medium block first in 7/20 and 5/20 runs when requested).

In all three tasks, the object displaced first under every instruction is the one the scripted demonstrations handle first: the red block in two tasks and the smallest block in the third (Appendix~\ref{app:cause-hypotheses}).}

\begin{table}[!htbp]
\centering\scriptsize\color{black}
\setlength{\tabcolsep}{2.6pt}
\caption{{Target reassignment on three unmodified RoboTwin training tasks. \emph{Color} (R~=~\nolinkurl{blocks_ranking_rgb}, S~=~\nolinkurl{stack_blocks_three}; ten runs per cell): runs under ``Pick up the blue/green block.'' in which that block is displaced first, runs under the blue instruction in which the original task is still completed (\emph{Done}), and empty-instruction runs with the red block first. \emph{Size} (\nolinkurl{blocks_ranking_size}; 20 runs per cell): runs in which the largest (L), medium (M), or smallest (S) block is displaced first under each instruction, and runs under ``largest'' in which the original task is still completed. \emph{Train}: success under the training instructions, with the published clean-setting value (\%) in parentheses. $^{\circ}$No published value. $^{\ast}$Below the published value on one task by rule (iv). $^{\diamond}$Below 30\% under the training instructions (rule (v)). $^{\S}$, $^{\P}$: as in Table~\ref{tab:reassignment}.}}
\label{tab:training-scene}\label{tab:training-scene-size}
\begin{tabular}{@{}lllrrrrrrrr@{\hspace{8pt}}lrrrrrrrr@{}}
\toprule
 & \multicolumn{10}{c}{Color tasks} & \multicolumn{9}{c}{Size task} \\
\cmidrule(lr){2-11}\cmidrule(l){12-20}
 & \multicolumn{2}{c}{Train} & \multicolumn{2}{c}{Blue} & \multicolumn{2}{c}{Green} & \multicolumn{2}{c}{Done} & \multicolumn{2}{c}{Empty} & Train & \multicolumn{2}{c}{``largest''} & \multicolumn{2}{c}{``medium''} & ``small.'' & Empty & Train & Done \\
\cmidrule(lr){2-3}\cmidrule(lr){4-5}\cmidrule(lr){6-7}\cmidrule(lr){8-9}\cmidrule(lr){10-11}\cmidrule(lr){13-14}\cmidrule(lr){15-16}
Checkpoint & R & S & R & S & R & S & R & S & R & S & & L & S & M & S & S & S & S & \\
\midrule
$\pi_{0.5}$ & 17/30 (43) & 11/30 (15) & 6 & 9 & 9 & 10 & 0 & 0 & 5 & 10 & 3/20 (8)$^{\diamond}$ & 0 & 17 & 3 & 14 & 17 & 10 & 19 & 1 \\
StarVLA & 10/10 (99) & 8/10 (94) & 0 & 0 & 1 & 3 & 0 & 0 & 10 & 10 & 17/20 (79) & 0 & 20 & 0 & 20 & 20 & 20 & 20 & 16 \\
FastWAM & 10/10 (100) & 10/10 (95) & 0 & 0 & 0 & 0 & 10 & 0 & 10 & 10 & 19/20 (94) & 0 & 18 & 2 & 18 & 19 & 18 & 19 & 15 \\
AHA-WAM & 10/10 (100) & 10/10 (99) & 0 & 0 & 0 & 0 & 8 & 0 & 7 & 8 & 19/20 (98) & 0 & 19 & 3 & 17 & 19 & 11 & 20 & 18 \\
WLA & 10/10 (98) & 10/10 (95) & 0 & 0 & 0 & 0 & 8 & 9 & 10 & 10 & 17/20 (93) & 0 & 20 & 0 & 20 & 20 & 20 & 20 & 16 \\
G0.5 & 10/10 (100) & 9/10 (98) & 0 & 0 & 8 & 9 & 7 & 3 & 10 & 10 & 18/20 (96) & 0 & 20 & 0 & 20 & 20 & 20 & 20 & 17 \\
\midrule
DM0.5 & 10/10$^{\circ}$ & 9/10$^{\circ}$ & 1 & 0 & 1 & 0 & 9 & 0 & 9 & 10 & 16/20$^{\circ}$ & 0 & 20 & 0 & 20 & 20 & 20 & 20 & 20 \\
X-WAM & 29/30 (99)$^{\ast}$ & 27/30 (97)$^{\ast}$ & 0 & 0 & 0 & 0 & 10 & 0 & 10 & 10 & 19/20 (76) & 0 & 19 & 1 & 19 & 19 & 19 & 19 & 15 \\
Motus & 30/30 (99)$^{\ast}$ & 23/30 (91)$^{\ast}$ & 0 & 0 & 0 & 0 & 8 & 0 & 10 & 10 & 14/20 (75) & 0 & 15 & 5 & 15 & 14 & 13 & 16 & 8 \\
LingBot-VA & 28/30 (99)$^{\ast}$ & 30/30 (99)$^{\ast}$ & 5 & 1 & 7 & 6 & 5 & 8 & 10 & 10 & 19/20 (94) & 1 & 14 & 7 & 10 & 15 & 13 & 20 & 7 \\
\midrule
X-VLA$^{\S}$ & 9/30 (79) & 0/30 (22) & 0 & 0 & 0 & 0 & 7 & 0 & 10 & 10 & -- & -- & -- & -- & -- & -- & -- & -- & -- \\
$\pi_0$$^{\P}$ & 0/30 & 0/30 & 0 & 0 & 0 & 0 & 0 & 0 & 7 & 9 & -- & -- & -- & -- & -- & -- & -- & -- & -- \\
\bottomrule
\end{tabular}
\end{table}

{\paragraph{Source tasks: prerequisite skills.}
Table~\ref{tab:source-tasks} gives success under training instructions, in the same evaluator, on the ten RoboTwin source tasks linked to the reassignment scenes. Nine of the ten pooled checkpoints complete at least nine of the ten tasks in 70\% or more of episodes; the task below 70\% is \nolinkurl{hanging_mug} for seven of them. On 30 of the 34 source tasks linked to \Benchmark{} scenes (Figure~\ref{fig:task-dependencies}), run for nine checkpoints, eight reach 70\% on at least 25 tasks. Rule (v) removes six of the 100 checkpoint--scene pairs ($\pi_{0.5}$ in the three-color, size, size-rank, and mug scenes; StarVLA and LingBot-VA in the mug scene). On the remaining 94 pairs every instruction is followed on 29 of 940 states (3.1\%) and the same object is displaced under every instruction on 68.5\%, against 16.0\% and 33.3\% for the instruction-blind chooser.}

\begin{table}[H]
\centering\scriptsize\color{black}
\setlength{\tabcolsep}{2.5pt}
\caption{{Per-checkpoint results. \emph{Formal failures}: decomposition over the eleven pick-type selection scenes (1,100 formal episodes per checkpoint), columns as in Table~\ref{tab:failure-decomposition} (D~=~distractor only, B~=~both, T~=~target only, N~=~neither; \% of failures). \emph{Source tasks}: successes/episodes under training instructions in the same evaluator on the ten RoboTwin tasks linked to the reassignment scenes (RGB~=~\nolinkurl{blocks_ranking_rgb}; ST3~=~\nolinkurl{stack_blocks_three}; SIZ~=~\nolinkurl{blocks_ranking_size}; HND~=~\nolinkurl{handover_block}; CNT~=~\nolinkurl{place_container_plate}; A2L, A2R~=~\nolinkurl{place_a2b_left}, \nolinkurl{place_a2b_right}; BRG~=~\nolinkurl{place_burger_fries}; MUG~=~\nolinkurl{hanging_mug}; BWL~=~\nolinkurl{stack_bowls_two}); bold marks cells below the 30\% threshold of rule (v). $^{\S}$Natural-language instructions, out of distribution for this checkpoint; $^{\P}$as in Table~\ref{tab:reassignment}.}}
\label{tab:failure-decomposition-policy}\label{tab:source-tasks}
\begin{tabular}{@{}lrrrrrr@{\hspace{8pt}}rrrrrrrrrr@{}}
\toprule
 & \multicolumn{6}{c}{Formal failures} & \multicolumn{10}{c}{Source tasks} \\
\cmidrule(lr){2-7}\cmidrule(l){8-17}
Checkpoint & SR & $n$ & D & B & T & N & RGB & ST3 & SIZ & HND & CNT & A2L & A2R & BRG & MUG & BWL \\
\midrule
DM0.5 & 27.8 & 794 & 41.4 & 46.1 & 0.9 & 11.6 & 10/10 & 9/10 & 9/10 & 9/10 & 10/10 & 9/10 & 10/10 & 10/10 & 6/10 & 10/10 \\
$\pi_{0.5}$ & 29.7 & 773 & 47.6 & 38.2 & 7.1 & 7.1 & 17/30 & 11/30 & \textbf{2/10} & \textbf{1/10} & 7/10 & 5/10 & 6/10 & 6/10 & \textbf{1/10} & 9/10 \\
StarVLA & 26.5 & 808 & 24.0 & 59.8 & 4.0 & 12.3 & 10/10 & 8/10 & 9/10 & 9/10 & 10/10 & 10/10 & 9/10 & 10/10 & \textbf{1/10} & 10/10 \\
FastWAM & 23.4 & 843 & 36.4 & 43.8 & 5.8 & 14.0 & 10/10 & 10/10 & 9/10 & 8/10 & 10/10 & 10/10 & 10/10 & 10/10 & 5/10 & 10/10 \\
$\pi_0$$^{\P}$ & 13.4 & 953 & 44.3 & 14.1 & 1.7 & 40.0 & \textbf{0/30} & \textbf{0/30} & \textbf{0/10} & \textbf{0/10} & \textbf{0/10} & \textbf{0/10} & \textbf{1/10} & \textbf{0/10} & \textbf{0/10} & \textbf{1/10} \\
AHA-WAM & 28.0 & 792 & 32.4 & 55.6 & 3.0 & 9.0 & 10/10 & 10/10 & 9/10 & 9/10 & 9/10 & 9/10 & 9/10 & 10/10 & 10/10 & 8/10 \\
X-VLA$^{\S}$ & 20.3 & 877 & 43.0 & 37.1 & 7.4 & 12.5 & 9/30 & \textbf{0/30} & 3/10 & 5/10 & 9/10 & \textbf{2/10} & \textbf{2/10} & 9/10 & 5/10 & 7/10 \\
WLA & 24.9 & 826 & 36.3 & 55.3 & 0.8 & 7.5 & 10/10 & 10/10 & 9/10 & 10/10 & 10/10 & 8/10 & 9/10 & 10/10 & 7/10 & 8/10 \\
G0.5 & 32.1 & 747 & 53.5 & 37.9 & 0.5 & 8.0 & 10/10 & 9/10 & 9/10 & 10/10 & 10/10 & 10/10 & 10/10 & 10/10 & 3/10 & 10/10 \\
X-WAM & 33.4 & 733 & 41.7 & 50.8 & 1.9 & 5.6 & 29/30 & 27/30 & 9/10 & 9/10 & 10/10 & 10/10 & 10/10 & 10/10 & 3/10 & 10/10 \\
Motus & 29.1 & 780 & 36.7 & 53.3 & 1.4 & 8.6 & 30/30 & 23/30 & 8/10 & 10/10 & 10/10 & 8/10 & 7/10 & 10/10 & 3/10 & 10/10 \\
LingBot-VA & 39.9 & 661 & 20.9 & 73.2 & 0.9 & 5.0 & 28/30 & 30/30 & 10/10 & 10/10 & 10/10 & 9/10 & 10/10 & 10/10 & \textbf{2/10} & 9/10 \\
\midrule
All twelve & 27.4 & 9587 & 38.4 & 46.1 & 3.0 & 12.4 &  &  &  &  &  &  &  &  &  &  \\
\bottomrule
\end{tabular}
\end{table}

{\subsection{From Diagnosis to Cause and Remedy}
\label{app:cause-hypotheses}

Each measurement points to a different place to intervene. Wrong-object commitment that persists under reassignment points to how training data and conditioning bind the instruction to the choice; target-only and no-movement failures point to grasping and control in the changed scene; a forecast that already depicts the wrong target locates the error before execution.

The demonstrations are one candidate cause. In the RoboTwin block tasks that include a red block (\nolinkurl{stack_blocks_three}, \nolinkurl{blocks_ranking_rgb}, and \nolinkurl{stack_blocks_two}), the released scripted expert always manipulates the red block first~\citep{chen2025robotwin2}, and in \nolinkurl{blocks_ranking_size} it handles the smallest block first. On the three training tasks probed, the object a checkpoint displaces first under every instruction is that object (Appendix~\ref{app:reassignment-controls}); in the paired study, Motus first contacts the red block in 48 of 80 trials, at nearly the same rate whether red or blue is requested (Table~\ref{tab:language-color-layout}). That the smallest block is also the easiest to grasp is an alternative these data leave open. $\pi_{0.5}$, trained on the same demonstrations, follows color instructions, so the demonstrations do not force a fixed choice. These results suggest three directions for a remedy. On the data side, the block tasks each have one fixed goal, so the instruction never decides which object is handled; demonstrations in which it does, with first targets balanced across the objects, address the cause directly. On the model side, failed episodes mostly displace a distractor and the forecast already depicts it, so a remedy should act on target selection, not on grasping or control. Since the demonstrations do not force the failure, remedies that add no data are also worth testing, such as inference-time guidance of the kind proposed for LIBERO-CF~\citep{fang2026liberocf}. Each can be evaluated with the measurements above: the fraction of initial states on which every instruction is followed, against the instruction-blind chooser.

\paragraph{Two remedies without new data, tested on one checkpoint.}
We tested the last two directions on DM0.5, which follows every reassigned instruction on 0 of 10 initial states in each scene used here, with the frozen states, instructions, and scoring of Appendix~\ref{app:reassignment}. \emph{Guidance.} Each denoising step of the flow-matching action head is evaluated with the instruction and with an empty instruction from the same images, state, and noise, and the velocity is $v_{\emptyset}+w\,(v_{\text{instr}}-v_{\emptyset})$. On the three-color scene, every instruction is followed on 0 of 10 states for $w=2$, $3$, and $5$ (success in 11, 16, and 14 of 30 episodes, against 12 for $w=1$): amplifying the instruction does not supply the selection. \emph{Handover.} The GPT-6 agent reads the instruction, selects the object it refers to, and brings the open gripper of one arm to it without grasping; DM0.5 then takes over with the original instruction. Every instruction is followed on 7 of 10 states of the three-color scene, 6 of 10 of the negation scene, and 3 of 10 of the second-from-left scene, and episodes succeed in 27/30, 16/20, and 23/30 (DM0.5 alone: 12/30, 10/20, and 13/30). The agent brought the gripper to the instructed object in all 80 episodes; the remaining errors follow the handover. This is an illustration on one checkpoint and three scenes, not a tuned method; it indicates that supplying target selection, with execution left to the policy, removes much of the failure.}

\FloatBarrier
{\section{Relation to LIBERO-Based Probes}
\label{app:libero-relation}

LIBERO-PRO, LIBERO-Plus, and LIBERO-CF probe instruction following in LIBERO (Table~\ref{tab:libero-comparison}). LIBERO-PRO reports nearly unchanged trajectories under corrupted instructions and attributes them to memorized action sequences and layouts~\citep{zhou2025liberopro}. LIBERO-Plus adds blank-instruction and goal-replacement tests, attributes the behavior to fixed vision--action mappings, and improves robustness by fine-tuning on a perturbed training set~\citep{fei2026liberoplus}. LIBERO-CF assigns alternative feasible instructions, scores contact with the instructed object, shows that policies default to the training task of the scene, supports this by removing the training-task objects, and proposes inference-time guidance that it also tests on a real robot~\citep{fang2026liberocf}.

Table~\ref{tab:libero-comparison} places the four studies side by side. All four change the instructed target within a scene and test an empty or meaningless instruction. LIBERO-PRO and LIBERO-Plus add visual and layout perturbations and report task success; LIBERO-Plus and LIBERO-CF also propose mitigations, by fine-tuning on a perturbed training set and by inference-time guidance, and LIBERO-CF reports real-robot results. Our remedy test is illustrative (Appendix~\ref{app:cause-hypotheses}). \Benchmark{} is complementary on the diagnostic side. Like LIBERO-CF, it records which object the policy acts on. It adds a joint score, whether every instruction is followed on one frozen initial state, read against a chance reference and a measured positive control. Its training demonstrations place the blocks over a 56\,cm\,$\times$\,13\,cm region (the block tasks), against 2--5\,cm within a LIBERO task, and its platform is bimanual.

Our results extend this account in four ways. (a) Within a training task that involves several objects, the object displaced first under every instruction is the one the scripted demonstrations handle first: the red block in two tasks and the smallest block in a third (Appendix~\ref{app:cause-hypotheses}). (b) In a scene with one fixed instruction, the success rate measures whether that preferred object is the target: 77.0\% in the scene that requests the smaller block and 1.2\% in the scene that requests the block on the plate (Section~\ref{sec:language-results}). (c) One set of checkpoints passes through a chain of controls: reassignment on frozen held-out states with an instruction-blind chooser and a positive control, the same reassignment on unmodified training tasks, and source-task rates in the same evaluator. (d) The probes cover attribute, relational, ordinal, and negated references, for vision--language--action and world-action checkpoints trained on the same demonstrations. On the two-arm platform, a successful episode can include a non-target grasped by the other gripper (Figure~\ref{fig:keyframes-w3}(a); 43 of 745 successes include a non-target grasp, Appendix~\ref{app:displacement-rescoring}).}

\begin{table}[!htbp]
\centering\small\color{black}
\setlength{\tabcolsep}{4pt}
\caption{{This work and three LIBERO-based studies of instruction following. \checkmark: present; $\times$: not part of the study or not found in its implementation.}}
\label{tab:libero-comparison}
\begin{tabularx}{\textwidth}{@{}>{\raggedright\arraybackslash}Xcccc@{}}
\toprule
 & LIBERO-PRO & LIBERO-Plus & LIBERO-CF & \Benchmark{} \\
\midrule
\multicolumn{5}{@{}l}{\emph{Instruction probes}} \\
Target changed to another object of the scene & \checkmark & \checkmark & \checkmark & \checkmark \\
Empty or meaningless instruction & \checkmark & \checkmark & \checkmark & \checkmark \\
\midrule
\multicolumn{5}{@{}l}{\emph{Robustness and deployment}} \\
Visual and layout perturbations & \checkmark & \checkmark & $\times$ & $\times$ \\
Real-robot results & $\times$ & $\times$ & \checkmark & $\times$ \\
Mitigation method & $\times$ & \checkmark & \checkmark & illustrative \\
\midrule
\multicolumn{5}{@{}l}{\emph{Diagnostic measurements}} \\
Object acted on is recorded & $\times$ & $\times$ & \checkmark & \checkmark \\
Every instruction followed on one state, with chance reference & $\times$ & $\times$ & $\times$ & \checkmark \\
Measured positive control & $\times$ & $\times$ & $\times$ & \checkmark \\
\midrule
\multicolumn{5}{@{}l}{\emph{Setting}} \\
Object position range in training demonstrations & 2--5\,cm & 2--5\,cm & 2--5\,cm & 56$\times$13\,cm \\
Bimanual platform & $\times$ & $\times$ & $\times$ & \checkmark \\
\bottomrule
\end{tabularx}
\end{table}

{\paragraph{The same probe in LIBERO-Object.}
We ran target reassignment on LIBERO-Object with the official LIBERO checkpoints of $\pi_{0.5}$ and StarVLA (Qwen3-VL with an OFT head): the object name in the instruction is replaced by that of another graspable object in the scene (Table~\ref{tab:libero-probe}). In the training data of this suite the target occupies one of two slots. When the instruction names the object in the other target slot, both checkpoints displace it first in 99--100\% of runs and place it in the basket. When it names an object in a slot that never holds a target, $\pi_{0.5}$ displaces it first in 49 of 400 runs and StarVLA in none; in a post hoc count, 295 and 391 of these 400 runs go to the target slot that the instructed name occupies in its own training task. The pattern differs from the fixed choice in RoboTwin: almost every state is \emph{mixed}, and the instruction acts between the positions that held targets in training. Four limits apply: position and language are confounded in this suite; the non-target slots have no grasp demonstrations; the checkpoints share model families with ours but are different weights; and the setting is single-arm with about 50 demonstrations per task.}

\begin{table}[!htbp]
\centering\scriptsize\color{black}
\setlength{\tabcolsep}{3.5pt}
\caption{{Target reassignment on LIBERO-Object (ten tasks, ten initial states each; readout as in Appendix~\ref{app:reassignment}). \emph{Original}: task success under the original instruction (published value). \emph{Instructed object first}: runs in which the object named in the instruction is the first displaced by 5\,cm, by where that object sits: its own target slot, the other slot that holds targets in training, or one of the slots that never hold a target (four instructions pooled). \emph{Home slot}, post hoc: runs of the last group in which the first displaced object sits in the target slot that the instructed name occupies in its own training task. \emph{Empty}: empty-instruction runs with a target-slot object first. \emph{Classes}: follow/fixed/mixed states with $k=3$.}}
\label{tab:libero-probe}
\begin{tabular}{@{}lrrrrrrr@{}}
\toprule
 & & \multicolumn{3}{c}{Instructed object first} & & & \\
\cmidrule(lr){3-5}
Checkpoint & Original & Own slot & Other target slot & Non-target slots & Home slot (post hoc) & Empty & Classes \\
\midrule
$\pi_{0.5}$ & 99/100 (98.2\%) & 99/100 & 99/100 & 49/400 & 295/400 & 91/100 & 3/0/97 \\
StarVLA & 100/100 (99.6\%) & 100/100 & 100/100 & 0/400 & 391/400 & 92/100 & 0/0/100 \\
\bottomrule
\end{tabular}
\end{table}

{\paragraph{Position and identity in the training data.}
The confound follows from how little object positions vary in LIBERO. Across the 50 initial states released with a task, the manipulated object moves by at most 1.9\,cm in LIBERO-Object, by about 3\,cm in LIBERO-Spatial and LIBERO-Goal (at most 8.6 and 4\,cm), and by about 5\,cm in LIBERO-10; the sampling regions of the task definitions are 5\,cm\,$\times$\,5\,cm in LIBERO-Object and mostly 2\,cm\,$\times$\,2\,cm in LIBERO-Spatial and LIBERO-Goal. The target occupies one of two positions across the ten LIBERO-Object tasks, and the ten LIBERO-Goal tasks share one layout. An object name can therefore be learned as a position. Consistent with this, the leaderboard of the LIBERO-PRO repository lists, under position perturbation, success rates of 0.00 on all four suites for OpenVLA and $\pi_0$ and of 0.08--0.38 for $\pi_{0.5}$~\citep{zhou2025liberopro}. LIBERO-PRO and LIBERO-Plus perturb the test scenes of policies trained on these demonstrations, so a probe run there either keeps the trained layout, where a memorized position suffices, or perturbs it, where a failure of position and a failure of instruction following are mixed. The RoboTwin block tasks differ: \nolinkurl{stack_blocks_three}, \nolinkurl{blocks_ranking_rgb}, and \nolinkurl{blocks_ranking_size} sample each block within a 56\,cm\,$\times$\,13\,cm region with a random yaw~\citep{chen2025robotwin2}, so a memorized position cannot replace the instruction on these training tasks. The \Benchmark{} reassignment scenes use fixed slots: two scenes permute the candidates over the slots, six mirror the layout left to right in half of the initial states, and the two ordinal scenes do neither, since position defines their target.}

\end{document}